\documentclass{article}
\usepackage{algorithm}
\usepackage{algorithmic}

\PassOptionsToPackage{numbers, compress}{natbib}
\usepackage[final]{neurips_2025}

\usepackage[utf8]{inputenc} % allow utf-8 input
\usepackage[T1]{fontenc}    % use 8-bit T1 fonts
\usepackage{url}            % simple URL typesetting
\usepackage{booktabs}       % professional-quality tables
\usepackage{amsfonts}       % blackboard math symbols
\usepackage{nicefrac}       % compact symbols for 1/2, etc.
\usepackage{microtype}      % microtypography
\usepackage{xcolor}         % colors
\usepackage{subfigure}

\usepackage{graphicx}
\usepackage[font=small]{caption}

\definecolor{cb_orange}{RGB}{213,94,0}
\definecolor{cb_green}{RGB}{34,136,51}
\definecolor{cbgreen}{RGB}{34,136,51}
\definecolor{sky_blue}{RGB}{204, 238, 255}
\definecolor{cb_purple}{RGB}{170, 51, 119}
\definecolor{cb_red}{RGB}{204, 51, 17}
\definecolor{cb_blue}{RGB}{0, 119, 187}
\definecolor{mydarkblue}{rgb}{0,0.08,0.45}
\definecolor{forestgreen}{RGB}{34,139,34}
\definecolor{periwinkle}{rgb}{0.8, 0.8, 1.0}
\definecolor{royalazure}{rgb}{0.0, 0.22, 0.66}
\definecolor{royalblue}{rgb}{0.0, 0.14, 0.4}
\usepackage[colorlinks=true,citecolor=mydarkblue,linkcolor=mydarkblue,urlcolor=mydarkblue]{hyperref}
\usepackage{tcolorbox}
\tcbuselibrary{breakable}

\usepackage{amsmath}
\usepackage{amssymb}
\usepackage{mathtools}
\usepackage{amsthm}

\usepackage[capitalize,noabbrev]{cleveref}

\usepackage{wrapfig}
\usepackage{enumitem}

\theoremstyle{plain}

\theoremstyle{definition}

\theoremstyle{remark}

\newcommand{\argmax}{\operatorname{argmax}}

\definecolor{cbgreen}{RGB}{34,136,51}
\definecolor{cbblue}{RGB}{0, 119, 187}
\definecolor{cbred}{RGB}{204, 51, 17}
\definecolor{richlilac}{rgb}{0.71, 0.4, 0.82}

\title{Inference-Time Text-to-Video Alignment with Diffusion Latent Beam Search}

\author{%
  Yuta~Oshima\textsuperscript{1}\quad
  Masahiro~Suzuki\textsuperscript{1}\quad
  Yutaka~Matsuo\textsuperscript{1}\quad
  Hiroki~Furuta\textsuperscript{2}\thanks{Work done as an advisory role only.} \\ 
  \textsuperscript{1}The University of Tokyo \quad
  \textsuperscript{2}Google DeepMind \\
  \texttt{yuta.oshima@weblab.t.u-tokyo.ac.jp} %\\
}

\begin{document}

\maketitle
\setcounter{footnote}{0}

\begin{abstract}
  % The abstract paragraph should be indented \nicefrac{1}{2}~inch (3~picas) on
  % both the left- and right-hand margins. Use 10~point type, with a vertical
  % spacing (leading) of 11~points.  The word \textbf{Abstract} must be centered,
  % bold, and in point size 12. Two line spaces precede the abstract. The abstract
  % must be limited to one paragraph.
The remarkable progress in text-to-video diffusion models enables the generation of photorealistic videos, although the content of these generated videos often includes unnatural movement or deformation, reverse playback, and motionless scenes.
Recently, an alignment problem has attracted huge attention, where we steer the output of diffusion models based on some measure of the content's goodness.
Because there is a large room for improvement of perceptual quality along the frame direction, we should address which metrics we should optimize and how we can optimize them in the video generation.
In this paper, we propose \textit{diffusion latent beam search} with \textit{lookahead estimator}, which can select a better diffusion latent to maximize a given alignment reward at inference time.
We then point out that improving perceptual video quality with respect to alignment to prompts requires \textit{reward calibration} by weighting existing metrics.
This is because when humans or vision language models evaluate outputs, many previous metrics to quantify the naturalness of video do not always correlate with the evaluation.
We demonstrate that our method improves the perceptual quality evaluated on the calibrated reward, VLMs, and human assessment, without model parameter update, and outputs the best generation compared to greedy search and best-of-N sampling under much more efficient computational cost. 
The experiments highlight that our method is beneficial to many capable generative models, and provide a practical guideline: we should prioritize the inference-time compute allocation into enabling the lookahead estimator and increasing the search budget, rather than expanding the denoising steps. 
\footnote{Website: \url{https://sites.google.com/view/t2v-dlbs}}
\footnote{Code: \url{https://github.com/shim0114/T2V-Diffusion-Search}}
% \footnote{Code: \url{https://anonymous.4open.science/r/T2V-Diffusion-Search-537B}}
\end{abstract}

\section{Introduction}
The remarkable progress in text-to-video diffusion models enables photorealistic, high-resolution video generation~\citep{openai2024sora,gdm2024veo,chen2024videocrafter2,blattmann2023stablevd}.
Many future applications are anticipated, such as creating novel games~\citep{bruce2024genie}, movies~\citep{zhu2023moviefactory}, or simulators to control real-world robots~\citep{videoworldsimulators2024}. 
However, the detailed contents of the generated video often include unnatural movement or deformation, reverse playback, and motionless scenes, which should not happen in the real world.
For instance, simulating factual physics in the generated video is still challenging~\citep{liu2024physgen,bansal2024videophy}.
Recently, it has attracted a lot of attention to steering the output of diffusion models based on reward evaluation, quantifying the goodness of the content, which is studied as an alignment problem~\citep{lee2023aligning,huang2024scg}.
There is a large room for improvement of perceptual quality along the frame direction in the video, and to align models with our preferences, we should address which metrics to optimize and how to optimize them.

\begin{figure}[t]
\centering
\subfigure{\includegraphics[width=0.53\textwidth]{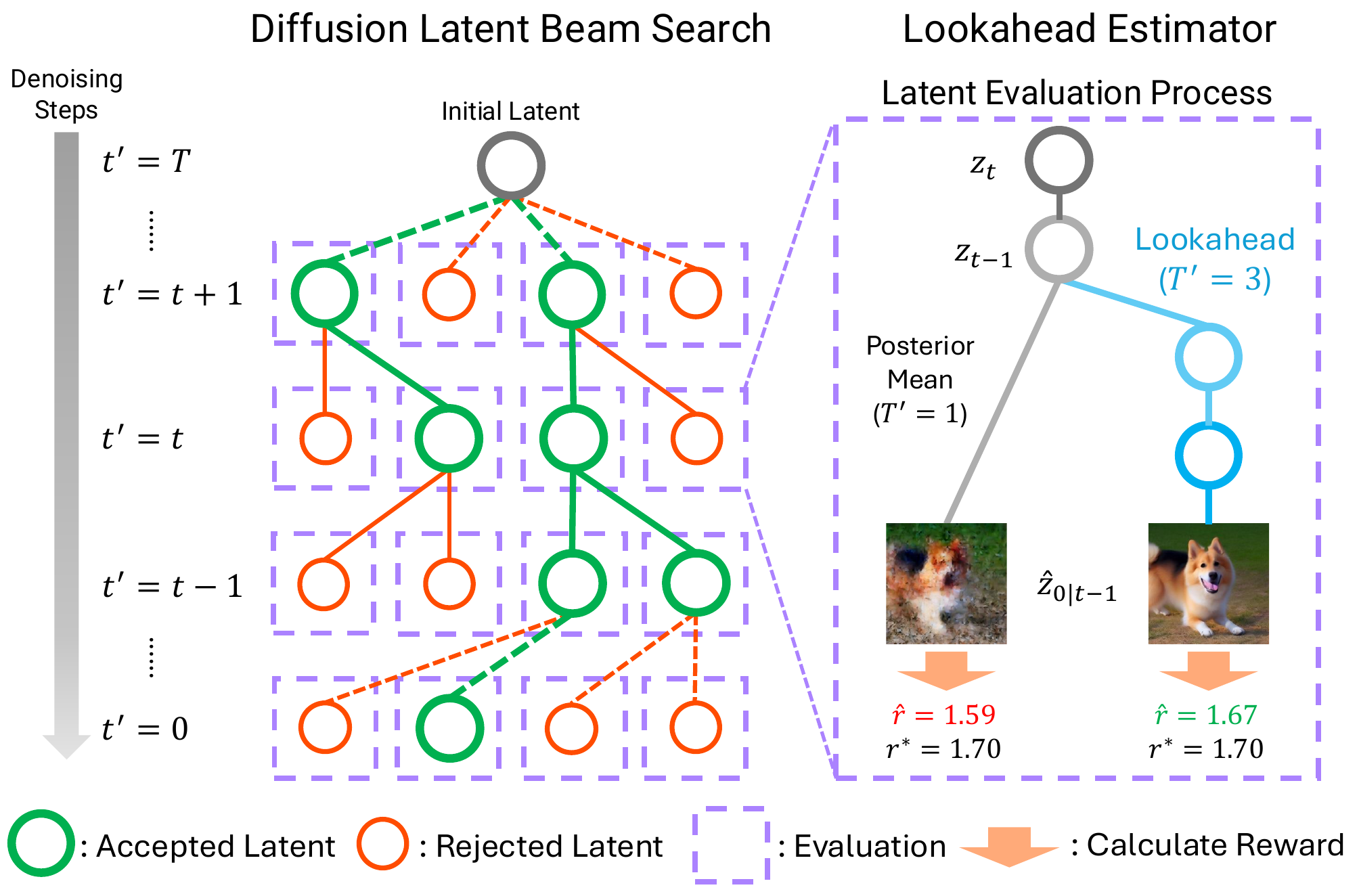}}
\subfigure{\includegraphics[width=0.45\textwidth]{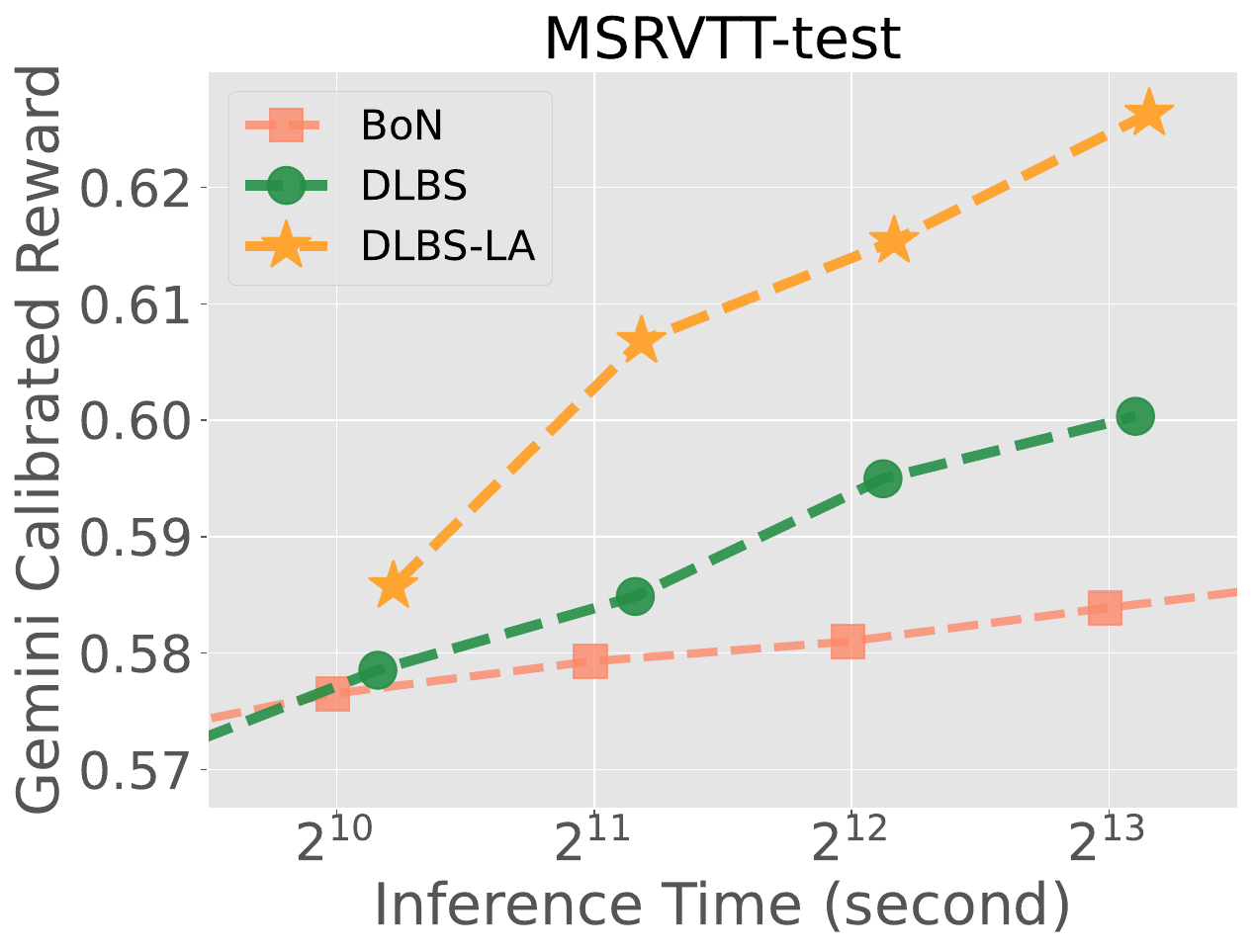}}
% \subfigure{\includegraphics[width=0.46\textwidth]{figures/reward_cost_fig1_v2.pdf}}
% \subfigure{\includegraphics[width=0.46\textwidth]{figures/reward_cost_fig1.pdf}}
% \subfigure{\includegraphics[width=0.54\textwidth]{figures/reward_cost.pdf}}
\vskip -0.125in
\caption{(\textbf{Left}) Diffusion latent beam search (DLBS) seeks a better diffusion path over the reverse process; sampling $K$ latents per beam and possessing $B$ beams for the next step, which mitigates the effect from inaccurate $\argmax$. Lookahead (LA) estimator notably reduces the noise at latent reward evaluation by interpolating the rest of the time steps from the current latent with deterministic DDIM.
(\textbf{Right}) DLBS achieves much better computational-efficiency than best-of-N (BoN), as achieving higher performance gains under the same execution time. LA estimator (DLBS-LA) could remarkably boost efficiency only with marginal overhead on top of DLBS.
% (\textbf{Right}) Trade-off between the number of function evaluations (NFE) or execution time and alignment performance. 
% DLBS achieves more efficient performance gains compared to Best-of-N (BoN) and Greedy Search (GS). 
% Furthermore, employing lookahead estimator (DLBS-LA) further accentuates these benefits, demonstrating that slight additional computational overhead realizes the improved alignment performance much more efficiently than with BoN or GS.
}
\vskip -0.2in
\label{fig:gd_vs_dlbs}
\end{figure}

In this paper, we propose \textit{Diffusion Latent Beam Search} (DLBS) with \textit{lookahead estimator}, an inference-time search over the reverse process~(\autoref{fig:gd_vs_dlbs}; \textbf{Left}), which can select a better diffusion latent to maximize a given alignment reward. 
A lookahead estimator reduces the noise in the reward estimate, and a beam search robustly explores the latent paths, avoiding inaccurate $\argmax$ operations.

We then point out that the improvement of perceptual video quality, considering the alignment to prompts, requires \textit{reward calibration} of existing metrics~\citep{huang2023vbench}.
When evaluating outputs using capable vision language models~\citep{openai2023gpt4, geminiteam2023gemini} or human raters, many previous metrics for quantifying video naturalness do not always correlate with them. 
Optimal reward design for measuring perceptual quality highly depends on the degree of dynamics described in evaluation prompts.
We design a weighted linear combination of multiple metrics, which is calibrated to perceptual quality and improves the correlation with VLM/human preference.

\begin{figure}[t]
  \centering
  \includegraphics[width=\linewidth]{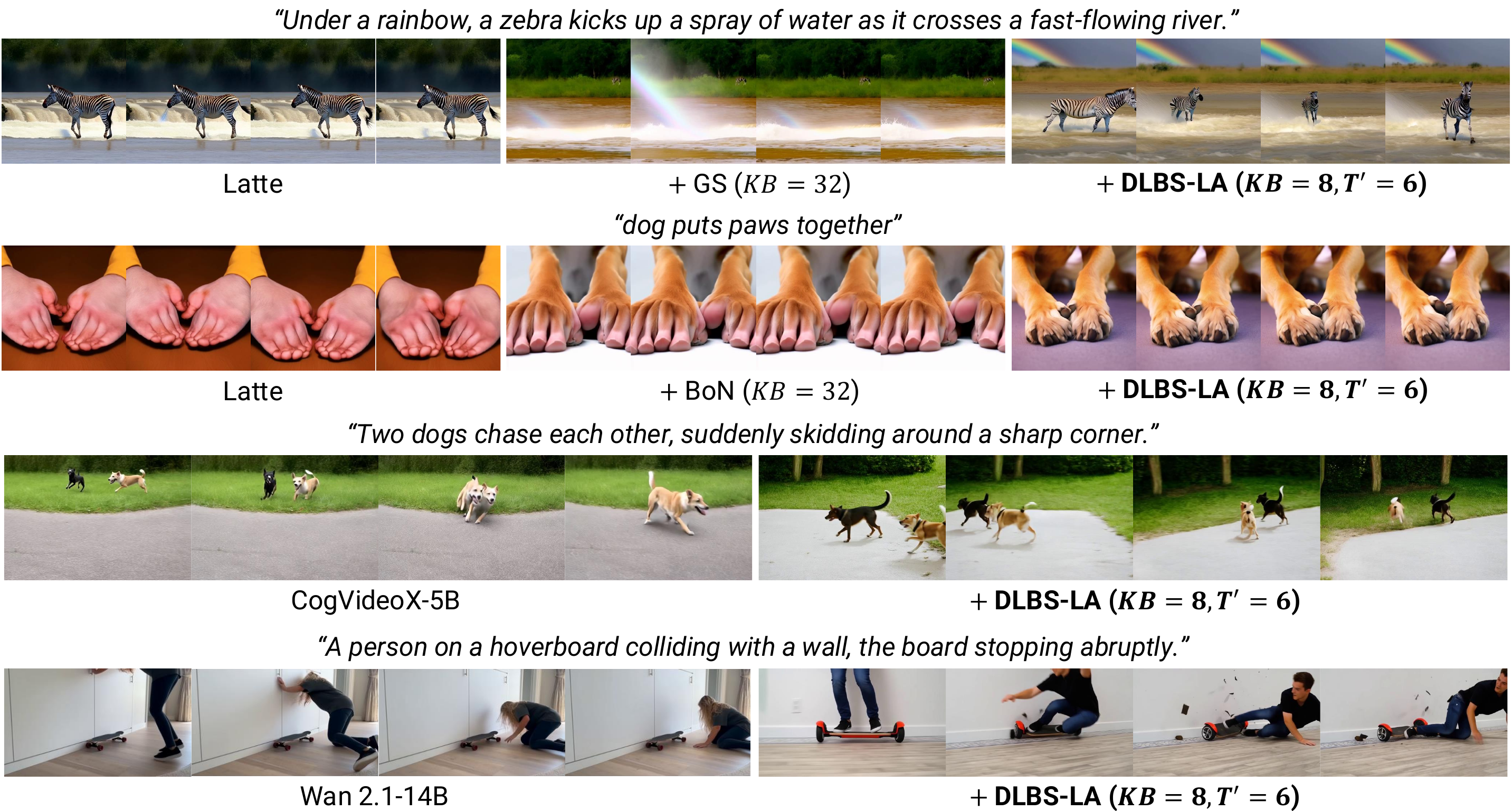}
  % \vskip -0.05in
  \caption{
  % Qualitative evaluation among different search methods. 
  % We test prompts from DEVIL-high, DEVIL-medium, DEVIL-static, and MSRVTT-test (from top to bottom).
  % DLBS-LA generates more dynamic, and prompt-aligned videos than GS or BoN.
  % We test prompts from DEVIL-high (above) and DEVIL-static (below).
  Comparison of text-to-video results between DLBS-LA, base models, and other sampling methods on SoTA models (Latte~\citep{ma2024latte}, CogVideoX~\citep{yang2024cogvideox}, and Wan 2.1~\citep{wan2025}). 
  DLBS-LA produces more dynamic, natural, and prompt-aligned videos than all baselines.
  }
  \vskip -0.175in
  \label{fig:search_qualitative}
\end{figure}

We demonstrate that DLBS can induce high-quality outputs based on the calibrated reward, AI, and human feedback~(\autoref{fig:search_qualitative}), without model parameter update, and realize the best generation under much more efficient computational cost compared to greedy search~\citep{kim2024free2guide,huang2024scg} and best-of-N sampling~\citep{zhang2024confronting,ma2025inferencetime}.
The experiments also highlight that our method is beneficial to many SoTA models (e.g., Latte~\citep{ma2024latte}, CogVideoX~\citep{yang2024cogvideox}, and Wan 2.1~\citep{wan2025}), and provide a practical guideline that we should prioritize the inference-time compute allocation into enabling the lookahead estimator and increasing the search budget, rather than expanding the denoising steps.
% We also provide practical efficiency guidelines on which axes, among search budget, lookahead steps for reward estimate, and denoising steps, in the denoising process we should allocate the inference-time computation.

\section{Preliminaries}
\textbf{Latent Diffusion Models}~~
Latent diffusion models~\citep{rombach2022ldm,ma2024latte} are a special class of diffusion probabilistic models~\citep{sohl-dickstein15,ho2020ddpm}, and popular choices for high-resolution text-to-video generation~\citep{he2022lvdm,blattmann2023videoldm,blattmann2023stablevd}, which considers the diffusion process in embedding space. Let $\textbf{x}_0$ be a video and encode it as $\textbf{z}_0= \text{Enc}(\textbf{x}_0)$ using VAE~\citep{kingma2013vae}.
Continuous-time forward diffusion process can be modeled as a solution to a stochastic differential equation (SDE)~\citep{song2021scorebased}: $d\textbf{z} = \textbf{f}(\textbf{z}, t)dt + g(t)d\textbf{w},$
% \begin{equation}
%     d\textbf{z} = \textbf{f}(\textbf{z}, t)dt + g(t)d\textbf{w},
% \end{equation}
where $\textbf{z}_0 \sim p_{0}(\textbf{z})$ is the latent as initial condition while $p_t(\textbf{z})$ is the marginal distribution of $\textbf{z}_t$, $\textbf{f}: \mathbb{R}^{d} \times \mathbb{R} \rightarrow \mathbb{R}^{d}$ is the drift coefficient, $g: \mathbb{R} \rightarrow \mathbb{R}$ is the diffusion coefficient, and $\textbf{w} \in \mathbb{R}^{d}$ is $d$-dimensional standard Wiener process.
$\textbf{f}(\cdot, \cdot)$ and $g(\cdot)$ are designed appropriately for the marginal distribution to reach $p_{T}(\textbf{z}) \approx \mathcal{N}(0, \textbf{I})$ as $t \rightarrow T$~\citep{karras2022elucidating}.
Reverse diffusion process generates samples $\textbf{z}_0$ through the following reverse-time SDE: $d\textbf{z} = [\textbf{f}(\textbf{z}, t) - g(t)^2\nabla_{\textbf{z}}\log p_t(\textbf{z})]dt 
 + g(t)d\bar{\textbf{w}},$
% \begin{equation}
%     d\textbf{z} = [\textbf{f}(\textbf{z}, t) - g(t)^2\nabla_{\textbf{z}}\log p_t(\textbf{z})]dt 
%  + g(t)d\bar{\textbf{w}},
%  \label{eq:reverse_sde}
% \end{equation}
where $dt$ here is an infinitesimal negative time step from $T$ to $0$ and $\bar{\textbf{w}} \in \mathbb{R}^{d}$ is a standard reverse-time Wiener process.
We start this with $\textbf{z}_T \sim \mathcal{N}(0, \textbf{I})$.
This SDE induces the marginal distribution on the data $p^{\text{pre}}(\textbf{z})$ (i.e., pre-trained diffusion models).
While we omit the notation for simplicity, we consider the text-to-video generation problem, where the diffusion process is conditioned on text prompts $\textbf{c}$.

\textbf{Alignment for Text-to-Video Diffusion Models}
In this paper, we define the alignment problem in text-to-video generation as increasing the probability of generating perceptually good video for humans, such as $\max \mathbb{E} [p(\mathcal{O}=1 | \textbf{x}_0, \textbf{c} )]$ where $\mathcal{O} \in \{0,1\}$ represents if the generated video $\textbf{x}_{0}$ conditioned on \textbf{c} is perceptually higher quality or not.
The common assumption is such a probability depends on a proxy scalar reward function $r(\textbf{x}_{0}, \textbf{c})$ such as $p(\mathcal{O}=1 | \textbf{x}_0, \textbf{c} ) \propto \exp(\beta^{-1} r(\textbf{x}_{0}, \textbf{c}))$ with $\beta \in \mathbb{R}$, and then the problem comes down to reward maximization.
The proxy reward function may input the generated video $\textbf{x}_0$ and a prompt $\textbf{c}$.

% \begin{figure*}[ht]
% \end{figure*}

% \subsection{Alignment as Stochastic Optimal Control}
\textbf{Alignment as Stochastic Optimal Control}~~
% \label{sec:soc}
Previous works formulate such a reward maximization problem from the view of stochastic optimal control~\citep{uehara2024finetuning,huang2024scg,domingoenrich2024adjoint}, where we aim to find an additional drift term $\textbf{u}(\cdot, \cdot)$ for the following reverse SDE:
$d\textbf{z} = [\textbf{f}(\textbf{z}, t) - g(t)^2\nabla_{\textbf{z} }\log p_t(\textbf{z}) + \textbf{u}(\textbf{z}, t)]dt  + g(t)d\bar{\textbf{w}}$.
%\begin{equation}
%    d\textbf{z} = [\textbf{f}(\textbf{z}, t) - g(t)^2\nabla_{\textbf{z} }\log p_t(\textbf{z}) + \textbf{u}(\textbf{z}, t)]dt  + g(t)d\bar{\textbf{w}}.
% \label{eq:sc_reverse_sde}
% \end{equation}
For convenience, we adopt the change-of-variables as $\boldsymbol{\nu}_{t} := \mathbf{z}_{T-t}$ and $\bar{\textbf{f}}(\boldsymbol{\nu}, t) := \textbf{f}(\boldsymbol{\nu}, t)- g(t)^2\nabla_{\boldsymbol{\nu}}\log p_t(\boldsymbol{\nu})$ because stochastic control is often based on the standard flow of time ($t: 0 \rightarrow T$), and then the original SDE is re-written as: $d\boldsymbol{\nu} = [\bar{\textbf{f}}(\boldsymbol{\nu}, t) + \textbf{u}(\boldsymbol{\nu}, t)]dt + g(t)d\textbf{w}$,
% and then \autoref{eq:sc_reverse_sde} is written as:
%\begin{equation}
%    d\boldsymbol{\nu} = [\bar{\textbf{f}}(\boldsymbol{\nu}, t) + \textbf{u}(\boldsymbol{\nu}, t)]dt + g(t)d\textbf{w},
% \label{eq:sc_reverse_sde_simple}
%\end{equation}
where $dt$ here is an infinitesimal time step and $d\textbf{w}$ is a standard Wiener process.

Because the alignment problem comes down to reward maximization, the objective in stochastic control literature is
\begin{equation}
    \textbf{u}^{*} = \underset{\textbf{u}}{\argmax}~\mathbb{E}\left[r'(\boldsymbol{\nu}_{T}) - \frac{\lambda}{2} \int_{t=0}^{T} \frac{\|\textbf{u}(\boldsymbol{\nu}_{t}, t) \|^2}{g(t)^2}dt  \right]
\end{equation}
where $r'(\cdot) := r(\text{Dec}(\cdot))$ evaluates the latent in the video space and $\lambda > 0$.
$\mathbb{E}[\cdot]$ is taken over sampling process above.
% The expectation is taken over sampling process above.
% The expectation is taken over samples from \autoref{eq:sc_reverse_sde_simple}.
In stochastic control, the optimal value function is known to be defined as,
\begin{equation}
\begin{split}
    v^{*}_{t}(\boldsymbol{\nu}) &= \mathbb{E}_{p^{*}}\left[r'(\boldsymbol{\nu}_{T}) - \frac{\lambda}{2} \int_{s=t}^{T} \frac{\|\textbf{u}(\boldsymbol{\nu}_{s}, s) \|^2}{g(s)^2}ds | \boldsymbol{\nu}_{t} = \boldsymbol{\nu}  \right], % \\
    % \textbf{u}^{*}(\boldsymbol{\nu}, t) &= g(t)^2 \nabla_{\boldsymbol{\nu}} \frac{v^{*}_{t}(\boldsymbol{\nu})}{\lambda},
\end{split}
\end{equation}
where $p^{*}_{t}(\boldsymbol{\nu}) = \frac{1}{Z}\exp(\frac{v^{*}_{t}(\boldsymbol{\nu})}{\lambda})p^{\text{pre}}_{t}(\boldsymbol{\nu})$, and obtain the optimal drift $\textbf{u}^{*}(\boldsymbol{\nu}, t) = g(t)^2 \nabla_{\boldsymbol{\nu}} \frac{v^{*}_{t}(\boldsymbol{\nu})}{\lambda}$~\citep{pavon1989stochastic}.
This optimal value function is the solution of stochastic Hamilton-Jacobi-Bellman equation~\citep{evans2022pde} according to this Feynman-Kac formula~\citep{oksendal2003sde,moral2004feynman}:
\begin{equation}
    \exp\left(\frac{v^{*}_{t}(\boldsymbol{\nu})}{\lambda}\right) = \mathbb{E}_{p^{\text{pre}}}\left[\exp\left(\frac{r'(\boldsymbol{\nu}_{T})}{\lambda}\right) | \boldsymbol{\nu}_{t}=\boldsymbol{\nu} \right]
\end{equation}
and then we obtain a tractable form of the optimal drift term as:
\begin{equation}
    \textbf{u}^{*}(\boldsymbol{\nu}_{t}, t) = g(t)^2 \nabla_{\boldsymbol{\nu}} \log \mathbb{E}_{p^{\text{pre}}}\left[\exp\left(\frac{r'(\boldsymbol{\nu}_{T})}{\lambda}\right) | \boldsymbol{\nu}_{t}=\boldsymbol{\nu} \right].
    \label{eq:tractable_optimal_drift}
\end{equation}
The intuition here is that the optimal drift pulls the current latent $\boldsymbol{\nu}$, while following the pre-trained reverse SDE, into the region achieving a higher reward at time $T$.

\section{Diffusion Latent Beam Search}
\label{sec:method}
We first provide a unified view of existing inference-time alignment methods through several practical approximations of optimal drift $\mathbf{u}^{*}(\boldsymbol{\nu}_{t}, t)$~(Section~\ref{sec:approx}). To mitigate errors from approximations, we propose a novel search algorithm, \textit{diffusion latent beam search} with \textit{lookahead estimator}~(Section~\ref{sec:dlbm}).

\subsection{A Unified View on Practical Approximations}
\label{sec:approx}
While \autoref{eq:tractable_optimal_drift} has a relatively tractable form, it is still computationally expensive, since the expectation requires complete diffusion sampling to evaluate the latent at each time step and face numerical instability.
Previous alignment methods rely on multiple-step practical approximations.

\textbf{Step. 1: Jensen's Inequality}~~
First, when assuming $\frac{r'(\boldsymbol{\nu}_{T})}{\lambda}$ is almost deterministic (this might hold when $t\rightarrow T$), Jensen's inequality yields the following approximation by exchanging $\log$ and $\mathbb{E}[\cdot]$, which can be considered as a certain form of classifier guidance~\citep{dhariwal2021diffusion}:
\begin{equation}
    \textbf{u}^{*}(\boldsymbol{\nu}_{t}, t) \approx \frac{g(t)^2}{\lambda} \nabla_{\boldsymbol{\nu}}\mathbb{E}_{p^{\text{pre}}}\left[r'(\boldsymbol{\nu}_{T}) | \boldsymbol{\nu}_{t}=\boldsymbol{\nu} \right].
    \label{eq:jensen_optimal_drift}
\end{equation}

\newcommand{\alglinelabel}{%
\addtocounter{ALC@line}{-1}
\refstepcounter{ALC@line}
\label%
}
\begin{wrapfigure}{r}{0.59\linewidth}
\scalebox{0.715}{
\begin{minipage}{1.35\linewidth}
\vskip -0.35in
% \vskip -0.2in
\begin{algorithm}[H]
% \begin{algorithm}[tb]
\caption{Diffusion Latent Beam Search (DLBS) with Stochastic DDIM}
\renewcommand{\algorithmicrequire}{\textbf{Input:}}
\label{alg:dlbs}
\begin{algorithmic}[1]
\REQUIRE latent diffusion model $\epsilon_{\theta}$, reward function $r'$, noise scheduling parameter $\{\alpha_t\}_{t=0}^{T}, \{\sigma_t\}_{t=0}^{T}$, number of beams $B$, number of candidates $K$
\STATE $\textbf{z}^{1}_{T}, \cdots, \textbf{z}^{B}_{T} \sim \mathcal{N}(\mathbf{0}, \mathbf{I})$ ~ {\color{richlilac} $\triangleright$ Initial $B$ beams} \alglinelabel{op:init}
\FOR{$t=T$ {\bfseries to} $1$}
    \FOR{$j=1$ {\bfseries to} $B$}
        \STATE {\color{richlilac} $\triangleright$ Compute the posterior mean of $\mathbf{z}_{t-1}^j$}
        \STATE $\hat{\textbf{z}}_{0|t}^j = \frac{1}{\sqrt{\alpha_{t}}}(\textbf{z}_{t}^j - \sqrt{1-\alpha_{t}} \epsilon_{\theta}(\textbf{z}_{t}^j))$
        \STATE $\textbf{z}^j_{t-1} = \sqrt{\alpha_{t-1}}\hat{\textbf{z}}_{0|t}^j + \sqrt{1-\alpha_{t-1} - \sigma^2_t} \epsilon_{\theta}(\textbf{z}_{t}^j)$  
    \ENDFOR
    \IF{$t>1$}
        \FOR{$j=1$ {\bfseries to} $B$}
            \STATE {\color{richlilac} $\triangleright$ Sample $K$ next candidate latents}
            \STATE $\mathbf{z}_{t-1}^{ij} = \mathbf{z}^{j}_{t-1} + \sigma_t \epsilon_{t}^i$ with $\epsilon_{t}^1, ..., \epsilon_{t}^K \sim \mathcal{N}(\mathbf{0}, \mathbf{I})$ \alglinelabel{op:sampling}
            \STATE {\color{richlilac} $\triangleright$ Estimate the clean sample from noisy latent}
            \STATE $\hat{\mathbf{z}}_{0|t-1}^{ij} = \frac{1}{\sqrt{\alpha_{t-1}}}(\mathbf{z}_{t-1}^{ij} - \sqrt{1-\alpha_{t-1}}\epsilon_\theta(\mathbf{z}^{ij}_{t-1}))$ \alglinelabel{op:posterior}
        \ENDFOR
        \STATE {\color{richlilac} $\triangleright$ Search $B$ higher-reward beams from $KB$ latents}
        \STATE $\texttt{budget} := \{(\mathbf{z}_{t-1}^{11},\hat{\mathbf{z}}_{0|t-1}^{11}),\cdots,(\mathbf{z}_{t-1}^{KB},\hat{\mathbf{z}}_{0|t-1}^{KB}) \}$
        \FOR{$j'=1$ {\bfseries to} $B$}
            \STATE $\mathbf{z}_{t-1}^{j'} = \argmax_{\mathbf{z}^{ij}_{t-1}\in\texttt{budget}}~r'(\hat{\mathbf{z}}^{ij}_{0|t-1})$ \alglinelabel{op:eval}
            \STATE $\texttt{budget} = \texttt{budget} \setminus \{(\mathbf{z}_{t-1}^{j'},\hat{\mathbf{z}}^{\argmax}_{0|t-1})\} $ \alglinelabel{op:bs}
        \ENDFOR
        \STATE $j \in \{1,\cdots,B\} \leftarrow j'$ ~ {\color{richlilac} $\triangleright$ Reset selected $B$ indices}
    \ENDIF
\ENDFOR
\STATE {\bfseries return: } $\mathbf{z}_{0} = \argmax_{\mathbf{z}^{j}_0\in\{\textbf{z}^1_{0}, \cdots,\textbf{z}^B_{0}\}}~r'(\mathbf{z}^{j}_0)$
\end{algorithmic}
\end{algorithm}
\end{minipage}
}
\vskip -0.15in
\end{wrapfigure}

\textbf{Step. 2: Tweedie's Formula}~~
To avoid the computationally expensive expectation, the expected reward is further approximated as
$\mathbb{E}_{p^{\text{pre}}}[r'(\boldsymbol{\nu}_T) | \boldsymbol{\nu}_{t} = \boldsymbol{\nu}] \approx r'(\hat{\boldsymbol{\nu}}_{T|t})$
where $\hat{\boldsymbol{\nu}}_{T|t} \approx \mathbb{E}_{p^{\text{pre}}}[\boldsymbol{\nu}_T | \boldsymbol{\nu}_{t}=\boldsymbol{\nu}]$ is a one-step approximation of posterior mean~\citep{chung2023diffusion}, which can be calculated only with the current latent $\boldsymbol{\nu}_{t}$ without full diffusion path.
Therefore, the optimal drift term can be seen as solely depending on the current time step $t$:
\begin{equation}
    \textbf{u}^{*}(\boldsymbol{\nu}_{t}, t) \approx \frac{g(t)^2}{\lambda} \nabla_{\boldsymbol{\nu}} r'(\hat{\boldsymbol{\nu}}_{T|t}).
    \label{eq:tweedie_optimal_drift}
\end{equation}
Such a computationally tractable drift term has been leveraged for previous inference-time alignment methods via approximate guidance or twisted sequential Monte Carlo (SMC)~\citep{wu2024twistedsmc}.
However, as the approximated posterior mean $\hat{\boldsymbol{\nu}}_{T|t}$in intermediate steps is noisy, evaluation with the reward function for clean data $r'(\cdot)$ may not provide a reliable signal~\citep{liang2024stepaware}.
Moreover, \autoref{eq:tweedie_optimal_drift} requires the reward gradient, which is not applicable to non-differentiable rewards, such as AI feedback, and is also not suitable for modalities whose reward gradient imposes a huge computational cost in practice, such as video.

\textbf{Step. 3: Converting Reward Gradient into argmax}~~
The usage of reward gradient can be converted into $\argmax$ operator~\citep{huang2024scg,li2024derivative,bansal2023universal}.
The intuition here is that since the optimal drift in \autoref{eq:tweedie_optimal_drift} induces the diffusion latent to the direction where it maximizes the reward, we replace such a maximization with a zeroth-order search.
The SDE is approximated as:
% Therefore, the approximated SDE has the following form:
\begin{equation}
    d\boldsymbol{\nu} = \bar{\textbf{f}}(\boldsymbol{\nu}, t)dt 
 + g(t)d\textbf{w}^{*}~~\text{where}~~d\textbf{w}^{*} = \argmax_{d\textbf{w}}~r'(\hat{\boldsymbol{\nu}}_{T|t}).
 \label{eq:sc_reverse_sde_argmax}
\end{equation}
Note that the current diffusion latents $\boldsymbol{\nu}_{t}$ and posterior mean $\hat{\boldsymbol{\nu}}_{T|t}$ are sampled by following the standard Wiener process $d\textbf{w}$.
This approximation is leveraged for inference-time alignment via greedy search~\citep{huang2024scg,li2024derivative} or SMC~\citep{singhal2025general} of diffusion latents.
However, greedy search can result in sub-optimal generation affected by inaccurate reward estimate~$r'(\hat{\boldsymbol{\nu}}_{T|t})$ due to its noisy input.
Moreover, it can be challenging to obtain an accurate density ratio term required in SMC for a high-dimensional domain, such as video generation.

\subsection{Mitigating Approximation Errors via Beam Search}
\label{sec:dlbm}
Existing practical algorithms based on these three approximations, such as greedy search~\citep{huang2024scg,li2024derivative}, fall into sub-optimal generation due to the erroneous reward evaluation with a noisy estimate of the posterior mean~\citep{chung2023diffusion}, and $\argmax$ operator based on them.
To resolve the error accumulation, we propose a simple yet robust modification, \textit{diffusion latent beam search (DLBS)} with \textit{lookahead estimator}.
To clearly describe the practical implementation, we use the notation of a discrete-time diffusion process in the rest of the section (see \autoref{sec:dlbs_for_dpmsolver++} for the continuous-time diffusion process).

\textbf{Practical Implementation}~~
We summarize the detailed sampling procedure of DLBS in Algorithm~\ref{alg:dlbs}.
For the diffusion sampler, we use stochastic DDIM~\citep{song2021denoising} with
a decreasing sequence $\{\alpha_t\}_{t=1}^{T} \in (0,1]^{T}$, noise level $\eta$, and noise schedule $\sigma_t = \eta \sqrt{(1-\alpha_{t-1})/(1-\alpha_{t})}\sqrt{1-\alpha_{t-1}/\alpha_{t}}$, which is equivalent to DDPM~\citep{ho2020ddpm} when $\eta = 1.0$.
We initialize $B$ latent beams from the Gaussian distribution (Line \ref{alg:dlbs}.\ref{op:init}), sample $K$ latents per beam in the next time step (Line \ref{alg:dlbs}.\ref{op:sampling}), and then compute the one-step estimation of the posterior mean~(Line~\ref{alg:dlbs}.\ref{op:posterior}).
DLBS evaluates the estimator of posterior mean $\hat{\mathbf{z}}_{0|t-1}$ with reward function~(Line~\ref{alg:dlbs}.\ref{op:eval}) and selects Top-$B$-rewarded latent beams instead of Top-$1$ (i.e., $\argmax$) from $KB$ candidates~(Line~\ref{alg:dlbs}.\ref{op:bs}), which is iterated over entire reverse process from $t=T$ to $t=0$.
DLBS can possess latent beams more widely than greedy search under the same budget, which mitigates error propagation due to the approximated diffusion latent evaluation.

\begin{wrapfigure}{r}{0.49\linewidth}
\scalebox{0.715}{
\begin{minipage}{1.35\linewidth}
\vskip -0.3in
\begin{algorithm}[H]
% \begin{algorithm}[tb]
\caption{Lookahead (LA) with Deterministic DDIM}
\renewcommand{\algorithmicrequire}{\textbf{Input:}}
\label{alg:lookahead}
\begin{algorithmic}[1]
\REQUIRE latent diffusion model $\epsilon_{\theta}$, current diffusion latent $\textbf{z}_{t-1}$, number of lookahead steps $T' (<<T)$
\STATE {\color{richlilac} $\triangleright$ Run $T'$-step deterministic DDIM  starting from $\textbf{z}_{t-1}$}
% \STATE Interpolate time steps $\tilde{t}(s) \in \{\lfloor\frac{T'}{T'}(t-1)\rfloor, \lfloor\frac{T'-1}{T'}(t-1)\rfloor, \cdots, \lfloor\frac{s}{T'}(t-1)\rfloor, \cdots , \lfloor\frac{1}{T'}(t-1)\rfloor\}$
\STATE $\tilde{t}(s) \in \{t-1, \dots,\lfloor\frac{s}{T'}(t-1)\rfloor,\dots , \lfloor\frac{1}{T'}(t-1)\rfloor,0\}$
% \STATE $\tilde{t}(s) \in \{\lfloor\frac{T'}{T'}(t-1)\rfloor, \cdots, \lfloor\frac{s}{T'}(t-1)\rfloor, \cdots , \lfloor\frac{1}{T'}(t-1)\rfloor\}$
\STATE Select new lookahead noise schedule $\{\tilde{\alpha}_{s}\}_{s=0}^{T'}$ for \textbf{$T'$-step interpolation} of the rest of original $\{\alpha_{t'}\}_{t'=0}^{t-1}$
\STATE $\textbf{z}_{\tilde{t}(T')} := \textbf{z}_{t-1}$
% \STATE $\textbf{z}_{\lfloor\frac{T'}{T'}(t-1)\rfloor} := \textbf{z}_{t-1}$
% \STATE $\textbf{z}_{\lfloor T'/T'(t-1)\rfloor} := \textbf{z}_{t-1}$
\STATE $\tilde{\textbf{z}}_{0|\tilde{t}(T')} = \frac{\textbf{z}_{\tilde{t}(T')} - \sqrt{1-\tilde{\alpha}_{T'}} \epsilon_{\theta}(\textbf{z}_{\tilde{t}(T')})}{\sqrt{\tilde{\alpha}_{T'}}}$
\FOR{$s=T'$ {\bfseries to} $1$}
    \STATE $\textbf{z}_{\tilde{t}(s-1)} = \sqrt{\tilde{\alpha}_{s-1}}\tilde{\textbf{z}}_{0|\tilde{t}(s)} + \sqrt{1-\tilde{\alpha}_{s-1}} \epsilon_{\theta}(\textbf{z}_{\tilde{t}(s)})$
    % \STATE $\tilde{\textbf{z}}_{0|\tilde{t}(s-1)} = \frac{1}{\sqrt{\tilde{\alpha}_{s-1}}}(\textbf{z}_{\tilde{t}(s-1)} - \sqrt{1-\tilde{\alpha}_{s-1}} \epsilon_{\theta}(\textbf{z}_{\tilde{t}(s-1)}))$
    \STATE $\tilde{\textbf{z}}_{0|\tilde{t}(s-1)} = \frac{\textbf{z}_{\tilde{t}(s-1)} - \sqrt{1-\tilde{\alpha}_{s-1}} \epsilon_{\theta}(\textbf{z}_{\tilde{t}(s-1)})}{\sqrt{\tilde{\alpha}_{s-1}}}$
\ENDFOR
\STATE {\bfseries return: } $(\textbf{z}_{t-1}, \tilde{\textbf{z}}_{0|\tilde{t}(0)})$ {\color{richlilac} $\triangleright$ Latent and LA estimator}
\end{algorithmic}
\end{algorithm}
\end{minipage}
}
\vskip -0.1in
\end{wrapfigure}

\textbf{Lookahead Estimator}~~
The other source of approximation errors than the $\argmax$ operator is a one-step estimator of the posterior mean $\hat{\mathbf{z}}_{0|t-1}$ from Tweedie's formula, which is still noisy, especially in earlier time steps, and leads to inaccurate reward evaluation.
To reduce errors in reward evaluation, we propose a lookahead (LA) estimator $\tilde{\mathbf{z}}_{0|\tilde{t}(0)}$, which is estimated by running $T'$-step deterministic DDIM $(1< T'<< T)$ while equally interpolating the rest of time steps from the current latent $\mathbf{z}_{t-1}$ to $\mathbf{z}_{0}$~(Algorithm~\ref{alg:lookahead}). 
While requiring additional denoising steps, its cost is almost the same as naive DLBS because most computational costs come from when we decode $\textbf{z}_0$ (i.e., reward evaluation). 
Theoretically, enlarging the lookahead steps \(T'\) monotonically tightens the upper bound on the reward-approximation error (see \autoref{sec:lookahead_proof}). 
Empirically, a modest horizon (\(T'=2,3,6\)) delivers substantial search improvements (\autoref{fig:scaling_ablation}; \textbf{Left}), but the marginal gains saturate, so pushing \(T'\) further yields little additional benefit (see Appendix~\ref{sec:lookahead_ablation}).
% This more ``accurate'' estimator significantly improves the performance even with $T' = 2$ or $3$ (Section~\ref{sec:lookahead_result}).
% Enlarging the look-ahead steps \(T'\) tightens the upper bound on reward-approximation error (see \autoref{sec:lookahead_proof}), yet the resulting performance gains plateau when the estimator is used for search (see \autoref{sec:additional_results}).

\section{Calibrating Reward to Preference Feedback}
\label{sec:reward_calibration}
Human evaluation is one of the most valuable assessments for generative models, yet gathering human feedback at scale is prohibitively costly. 
A practical approach to reduce the time and cost is to leverage AI feedback from VLMs~\citep{wu2024multimodal}, which has been shown to modestly align with human judgment on video quality~\citep{na2024boost, wu2024boosting, furuta2024improving} (see \autoref{sec:corr_vlm_human_eval}).
In this work, we assume that the VLM evaluation works as an oracle, and we align model outputs with the preferences of VLMs, which is reasonable due to their capability and the cost to be saved. Our qualitative and quantitative evaluations also confirm that the highly rated video by VLMs is generally good for us.
% If the cost allows, it is possible to replace VLMs with humans in our framework.

However, because alignment via inference-time search requires massive reward evaluation queries, we still need to build more tractable proxy rewards that do not rely on humans or external VLM APIs.
The question here is what metrics for perceptual video quality can improve the feedback from VLMs.
Because the criteria of videos preferred by humans are multi-objective, maximizing a single metric may lead to undesirable generation due to over-optimization.
For instance, focusing exclusively on temporal consistency or frame-by-frame quality metrics can unintentionally reduce the video's motion magnitude (see \autoref{sec:detailed_analysis_reward_metrics}).
In this section, we first review the possible video quality metrics (Section~\ref{sec:summary_of_video_metric}), evaluate the Pearson correlation between these and the VLM feedback score, and then propose a reward calibration (Section~\ref{sec:reward_results}), aiming to align the existing video rewards to VLMs by considering their weighted linear combination through the brute-force search of coefficients.

\begin{figure*}[t]
  \centering
  \includegraphics[width=\linewidth]{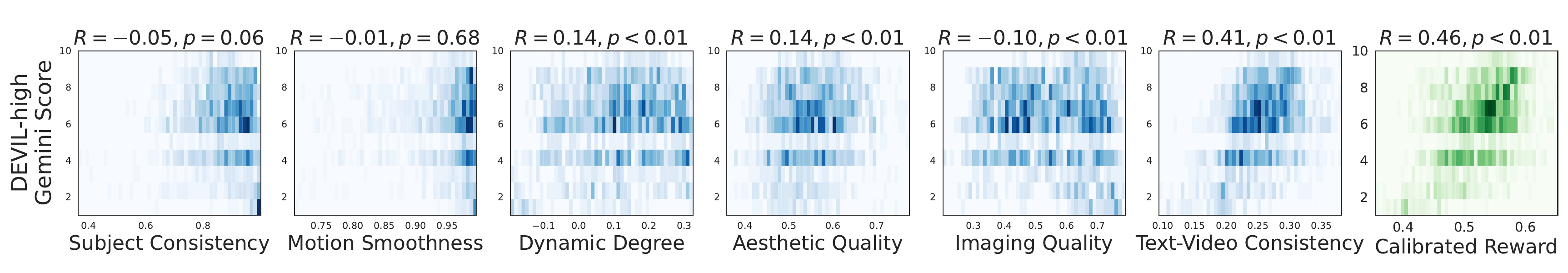}
  \vskip -0.1in
  \caption{%
  2D-histogram and correlation between reward functions for perceptual video quality~\citep{huang2023vbench} and AI feedback from Gemini~\citep{geminiteam2023gemini}.
  % We generate 64 videos per prompt from pre-trained Latte~\citep{ma2024latte}.
  A single reward (e.g., subject consistency; \textcolor{cbblue}{blue}) is often not aligned well with a preference from Gemini, which happens for all the prompt sets with different dynamics grades (see \autoref{fig:hist_gemini_full}).
  The calibrated reward, a linear combination of perceptual metrics via brute-force search (\textcolor{cbgreen}{green}), achieves the best Pearson correlation coefficient in all settings (statistically significant with $p<0.01$).
  }
  \vskip -0.15in
  \label{fig:hist_gemini}
\end{figure*}

\subsection{Metric Reward for Perceptual Video Quality}
\label{sec:summary_of_video_metric}
Following~\citet{huang2023vbench}, we select six base reward functions for perceptual video quality (see \autoref{sec:metric_reward}): 

\begin{itemize}[leftmargin=0.5cm,topsep=0pt,itemsep=0pt]
\item \textbf{Subject Consistency}~quantifies how consistently the subject appears across video frames with DINO~\citep{caron2021dino}.

\item \textbf{Motion Smoothness} leverages the motion prior in AMT~\citep{li2023amt} to evaluate whether the generated video’s motion is smooth and physically plausible. 

\item \textbf{Dynamic Degree} quantifies the overall magnitude of dynamic object movement by estimating optical flow~\citep{teed2020raft} for each pair of consecutive frames. 

\item \textbf{Aesthetic Quality} measures compositional rules, color harmony, and the overall artistic merit of each video frame with LAION aesthetic predictor~\citep{laion2022aesthentic}.

\item \textbf{Imaging Quality} assesses low-level distortions (e.g., over-exposure, noise, blur) in each frame with MUSIQ predictor~\citep{ke2021musiq}.

\item \textbf{Text-Video Consistency} captures how closely the content in a video aligns with a prompt with ViCLIP~\citep{xu2021videoclip}. 
\end{itemize}

% \begin{figure}
\begin{wrapfigure}{r}[0pt]{0.49\linewidth}
  \vskip -0.15in
  \centering
  \includegraphics[width=\linewidth]{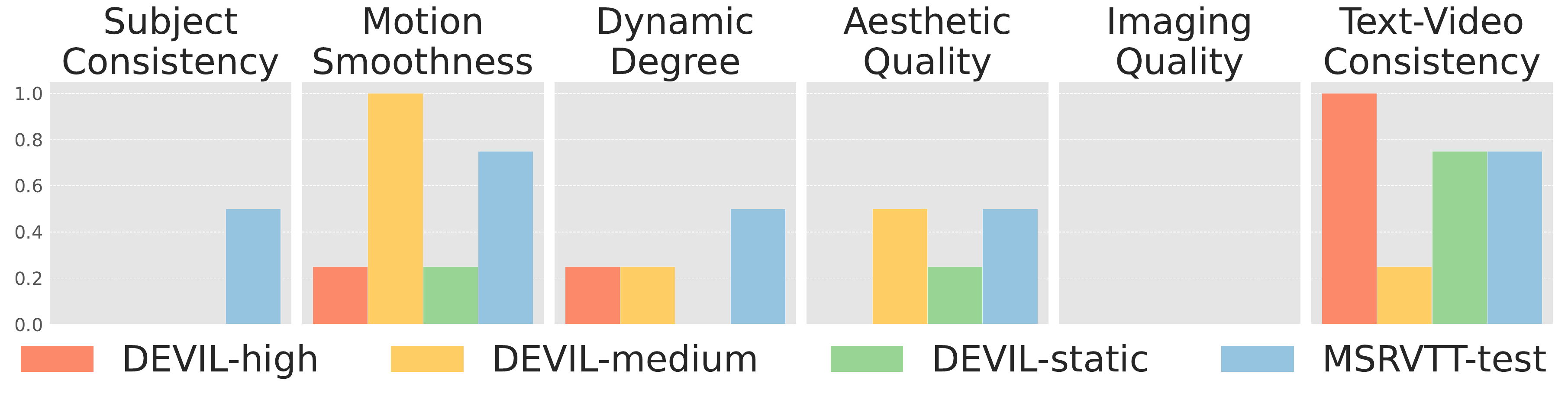}
  \vskip -0.08in
  \caption{%
  % The best coefficient of calibrated reward $w_i$ against feedback from Gemini. We observe that each set of prompts that depicit different degree of dynamics requires different mixture of perceptual video qualities. For instance, dynamic degree and temporal consistency exhibit pronounced variation-taking on greater or lesser importance among DEVIL-high, -medium, and -static.
  The coefficient of calibrated reward $w_i$ with feedback from Gemini. 
  Each set of prompts, which has a different dynamics grade, requires a distinct mixture of perceptual video qualities. % For instance, dynamic degree and motion smoothness exhibit pronounced variation among DEVIL-high, -medium, and -static.
  }
  \label{fig:weight_selection_gemini}
  \vskip -0.3in
\end{wrapfigure}
% \end{figure}

\textbf{Reward Calibration}~~To reflect the multi-dimensional aspect of preferred videos, we model the calibrated reward function $r^{*}(\cdot, \cdot)$ as a weighted linear combination of video quality metrics: $r^{*}(\mathbf{x}_0, \mathbf{c}) := \sum_{i=1}^{M} w_i r_i(\mathbf{x}_0, \mathbf{c})/\sum_{i=1}^{M} w_i$.
The coefficient $w_i$ is determined by maximizing the Pearson correlation with preference feedback. We heuristically conduct a brute-force search within a reasonable range (Section~\ref{sec:reward_results}).

% \subsection{Experimental Setup}
\textbf{Experimental Setup}~~We leverage Gemini-1.5~\citep{geminiteam2023gemini} and GPT-4o~\citep{openai2023gpt4} as automated raters for generated videos.
We provide a prompt and generated video as inputs, instructing VLMs to assign discrete scores (from 1 to 10) based on overall visual quality (e.g., clarity, resolution, brightness, and aesthetic appeal), the appropriateness of motion for either static or dynamic scenes, the smoothness and consistency of shapes and motions, and the degree of alignment with a prompt (see \autoref{sec:aif_prompt}).

We select four prompt sets from two distinct datasets (see \autoref{sec:prompts}). 
DEVIL~\citep{liao2024evaluation} classifies its prompts into five categories depending on the dynamics grade, each further divided by subject type (e.g., cat, horse, truck, nature, etc.).
We focus on three of the five dynamics grades (high, medium, and static) and select one prompt randomly from each subject-subdivision within a chosen category.
We also draw 30 random captions from the test split of MSRVTT~\citep{xu2016msr}, widely used as a video benchmark.

We generate 64 videos per prompt from pre-trained Latte~\citep{ma2024latte} using the DDIM sampler with $T=50$ and $\eta=0.0$ to examine the correlation among AI feedback and perceptual quality metrics.
We also prepare candidates for the calibrated reward by choosing the combination of weights $w_i \in \{0.0, 0.25, 0.5, 0.75, 1.0\}$ and use those later to rank them based on the correlation with AI feedback.
%We generate 64 videos per prompt from pre-trained Latte~\citep{ma2024latte} using the DDIM sampler with $T=50$ and $\eta=0.0$.
%We evaluate these videos using both single metrics and a calibrated reward. 
%The calibration procedure selects a linear combination of single metrics that achieves the highest correlation with the VLM scores. 
%We choose the weights \(w_i\) from the set $\{0.0, 0.25, 0.5, 0.75, 1.0\}$ and adopt the combination that yields the best correlation.

\subsection{Correlation and Reward Calibration}
\label{sec:reward_results}
\autoref{fig:hist_gemini} illustrates the 2D-histogram and the corresponding correlation between each metric and feedback from Gemini. 
Relying on a single metric often yields low correlation, which supports the multifaceted nature of perceptual video quality.
See Appendix~\ref{sec:reward_gpt_4} for further results, where we can see that the relative importance of each metric depends on the dynamics grade of the prompts; in highly dynamic DEVIL-high, the dynamic degree correlates more strongly with VLMs than consistency metrics.
Conversely, subject consistency and motion smoothness play more prominent roles in less-dynamic DEVIL-medium or DEVIL-static. 
Because the aesthetic score focuses on frame-by-frame visual quality, it tends to correlate strongly with VLM in low-motion scenarios. 
In high-motion scenarios, in contrast, rapid movements and frequent transitions often introduce motion blur or abrupt changes in composition, reducing the frame-level aesthetic quality and thus weakening its correlation with VLMs.

Reward calibration, a weighted linear combination of these metrics, yields the highest correlation with Gemini (\autoref{fig:hist_gemini}, \textcolor{cbgreen}{green}).
We select the best coefficients among brute-force candidates, based on the correlation with Gemini, for each set of prompts with a different dynamics grade (\autoref{fig:weight_selection_gemini}).
Prompts with a high dynamics grade, i.e., DEVIL-high, place greater weight on the dynamic degree. 
In contrast, prompts that describe slight motion, i.e., DEVIL-medium and DEVIL-static, place a smaller weight on it. 
In addition, \autoref{sec:reward_calibration_qualitative} presents results from best-of-64 sampling with a single metric or calibrated reward, where a single metric often leads to over-optimization.
This highlights the importance of reward calibration, appropriately weighting multiple criteria, as aligning with a request in the prompt.

% These trends align with previous findings that dynamic degree and consistency metrics correlate negatively~\citep{huang2023vbench} (see \autoref{sec:detailed_analysis_reward_metrics}). 

%Notably, combining these multidimensional metrics through reward calibration yields the highest correlation with Gemini, underscoring the importance of appropriately weighting multiple evaluation criteria through the reward calibration.

%\autoref{fig:weight_selection_gemini} illustrates how the coefficients for each metric shift are based on the prompt sets. 
%While dynamic degree and temporal consistency exhibit the most pronounced variation-taking on greater or lesser importance in DEVIL-high versus DEVIL-medium or DEVIL-static settings—the weights of other factors, such as text-video consistency, subject consistency, and aesthetic considerations, also adjust accordingly. 
%These findings mirror the earlier correlation analysis, confirming that video quality evaluation is inherently multidimensional and that different prompts demand distinct emphasis on each metric.  

\begin{figure}[t]
\centering
\subfigure{\includegraphics[width=0.63\textwidth]{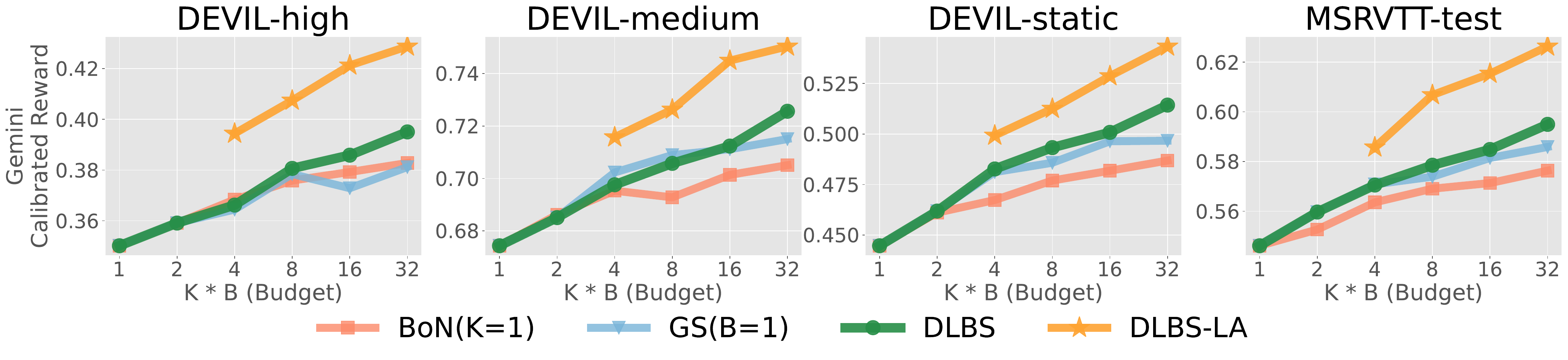}}
\subfigure{\includegraphics[width=0.36\textwidth]{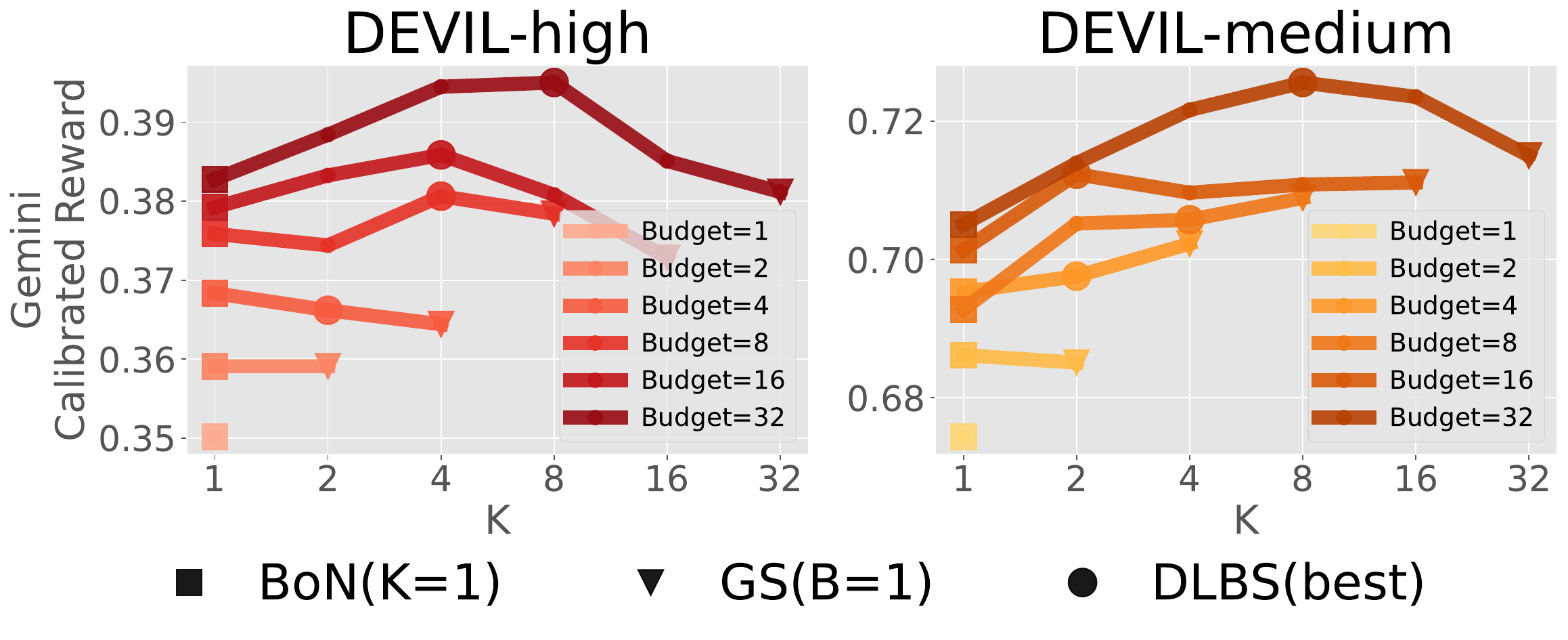}}
% \subfigure{\includegraphics[width=0.54\textwidth]{figures/reward_cost.pdf}}
\vskip -0.15in
\caption{(\textbf{Left}) 
      Comparison among diffusion latent beam search (DLBS), best-of-N (BoN), and greedy search (GS).
      We measure the performance in terms of a combinational reward calibrated to Gemini.
      % (above) and GPT-4o (below).
      DLBS improves all the calibrated rewards the best as the search budget $KB$ increases (especially $KB=16,32$), while BoN and GS, in some cases, eventually slow down or saturate the performance.
      Notably, an LA estimator with a small search budget ($KB=8, T'=6$) is comparable to or even outperforms DLBS ($KB=32$).
      (\textbf{Right}) 
      Optimal balance between the number of latent $K$ and the number of beams $B$ under the same budget.
      For instance, as we increase the budget to $KB=16,32$, we peak around $K=4,8,16$, which is about 25--50\% of the budget.
}
\vskip -0.2in
% \vskip -0.25in
\label{fig:main}
\end{figure}

% \begin{figure*}[t]
%   \centering
%   \includegraphics[width=0.875\linewidth]{figures/scaling_v2.pdf}
%   % \includegraphics[width=\linewidth]{figures/scaling.pdf}
%   \vskip -0.1in
%   \caption{%
%       Comparison among diffusion latent beam search (DLBS), best-of-N (BoN) and greedy search (GS).
%       We measure the performance in terms of a combinational reward calibrated to Gemini (above) and GPT-4o (below).
%       DLBS improves all the calibrated reward the best as the search budget $KB$ increases (especially $KB=16,32$) while BoN and GS in some cases eventually slows down or saturates the performance.
%       Notably, LA estimator with small search budget ($KB=8, T'=6$) is comparable to or even outperforming DLBS ($KB=32$).
%   }
%   \label{fig:main}
%   \vskip -0.15in
% \end{figure*}

\section{Inference-Time Text-to-Video Alignment}
\label{sec:main_results}

\textbf{Experimental Setup}~~We use the same prompts and Gemini-/GPT-calibrated rewards as in Section~\ref{sec:reward_calibration}.
We compare the following inference-time search methods with a noise level $\eta = 1.0$ for DDIM:
\begin{itemize}[leftmargin=0.5cm,topsep=0pt,itemsep=0pt]
    \item \textbf{Best-of-N Sampling (BoN):}~~We initialize $B$ latents and they follow the reverse process independently ($K=1$). At $t=0$, we evaluate the reward and select the best.
    \item \textbf{Greedy Search (GS):}~~At each denoising step, we select the best-rewarded diffusion latent ($B=1$) from $K$ candidates sampled in a reverse process.
    \item \textbf{DLBS:}~~Given the budget $KB$, we sweep possible combinations in terms of power of 2 (e.g., $K=8, B=2$), and report the best results except for the case with $K=1$ and $B=1$.
    % \item \textbf{DLBS:}~~Given the budget $KB$ ($K>1,B>1$), we sweep possible combinations in terms of power of 2 (e.g., $K=4, B=4$ or $K=8, B=2$), and report the best results.
     \item \textbf{DLBS-LA:}~~We combine DLBS with a lookahead estimator from the 6-step deterministic DDIM. 
     % Due to the resources, we have experiments with $KB=8$.
\end{itemize}
% We normalize the rewards with min/max values so that the range is around $[0, 1]$.
% Note that the range of values may be different from each other as they have different coefficients.
%
Our experiments aim to assess:
(1) scaling the search budget and computational costs for efficient resource allocation (Section~\ref{sec:dlbs_results});
(2) evaluating alignment performance with feedback from humans and VLMs (Section~\ref{sec:vlm_eval});
(3) assessing scalability to capable SoTA models (Section~\ref{sec:scaling_model_params});
(4) quantitative analysis on the diversity of generated video (Section~\ref{sec:diversity_eval}); 
(5) validating that DLBS is complementary to fine-tuning methods (Section~\ref{sec:dpo_dlbs});
(6) performing detailed ablations on LA steps $T'$, and robustness to diverse and complex prompts (Section~\ref{sec:ablation_study}).
%

% We use the same prompt sets as in Section~\ref{sec:reward_calibration}, and Gemini-/GPT-calibrated rewards in Section~\ref{sec:reward_results}. Note that the range of values may be different from each other as they have different coefficients.
% We test the performance when scaling (1) search budget $KB$ (Section~\ref{sec:dlbs_results}), (2) LA steps $T'$ for reward estimate  (Section~\ref{sec:lookahead_result}), and (3) denoising steps (Section~\ref{sec:scale_denoise}).
% Under a fixed inference budget, we conduct a controlled comparison across these three axes and provide practical guidelines for efficient resource allocation (Section \ref{sec:scale_cost}).
% We further analyze whether DLBS maximizing reward calibrated to VLMs can actually improve VLM and Human evaluation (Section~\ref{sec:vlm_eval}), how the diversity of generated samples is in the inference-time search (Section~\ref{sec:diversity_eval}), and whether DLBS is effective for other models (Section~\ref{sec:other_models}) and for a complex and large prompt set (Section~\ref{sec:moviegen_full}).

\subsection{Scaling Search Budget and Computational Cost}
\label{sec:dlbs_results}
\autoref{fig:main} (\textbf{Left}) measures the combinatorial reward calibrated to Gemini while increasing the search budget $KB \in \{1,2,4,8,16,32\}$.
DLBS improves all the calibrated rewards the best as $KB$ increases (especially $KB=16,32$), while BoN and GS eventually slow down or saturate the performance in some cases.
See Appendix~\ref{sec:gpt_scaling} for results with GPT calibrated reward and Appendix~\ref{sec:larger_scaling} for the results with $KB=64$, where we still observe the improvement.

\autoref{fig:main} (\textbf{Right}) demonstrates the scaling trend of DLBS, proportional to the search budget, under various choices of $K$.
The results show that there is an optimal balance between the number of latent $K$ and the number of beams $B$ under the same budget.
For instance, as we increase the budget to $KB=16,32$, we peak around $K=4,8,16$, which is about 25--50\% of the budget.
This implies that balancing possession and exploration of diffusion latents in DLBS helps search for the best outputs robustly. 
See Appendix \ref{sec:full_optimal_balance} for further results.

\begin{figure}[t]
\centering
\subfigure{\includegraphics[width=0.47\textwidth]{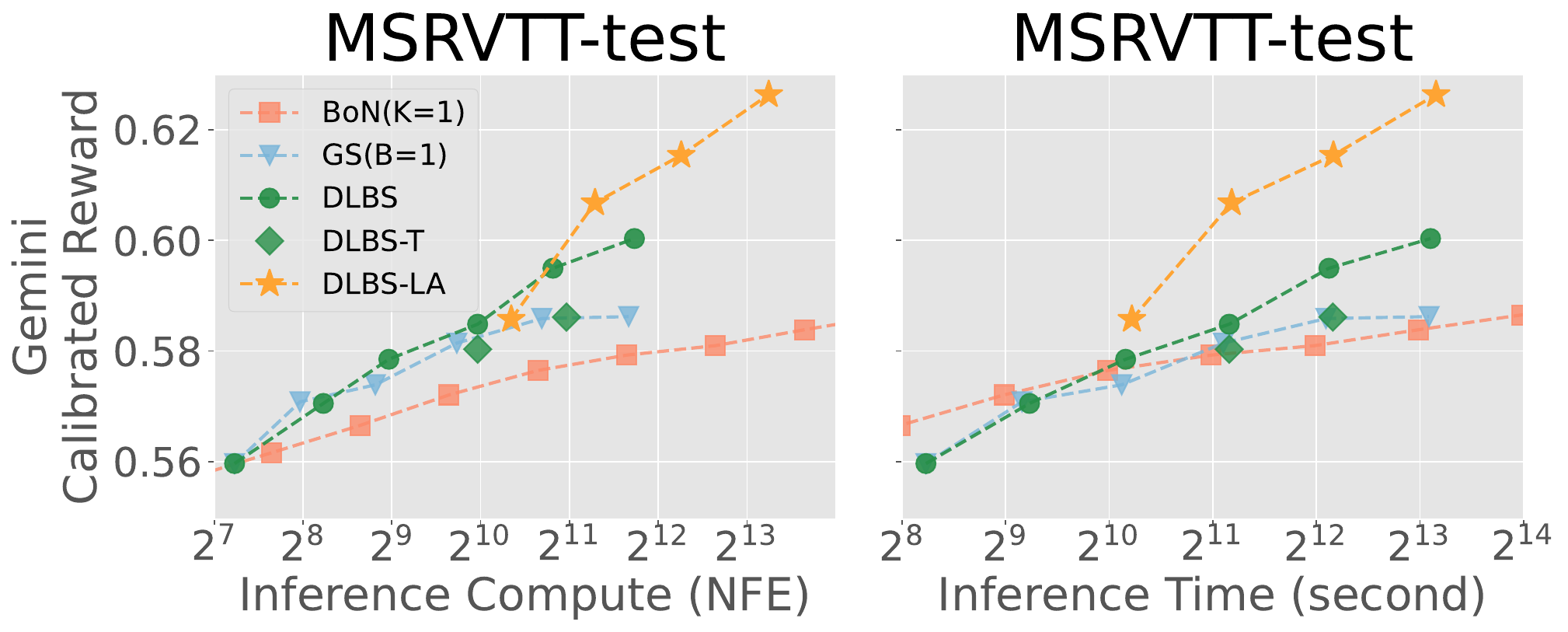}}
\subfigure{\includegraphics[width=0.258\textwidth]{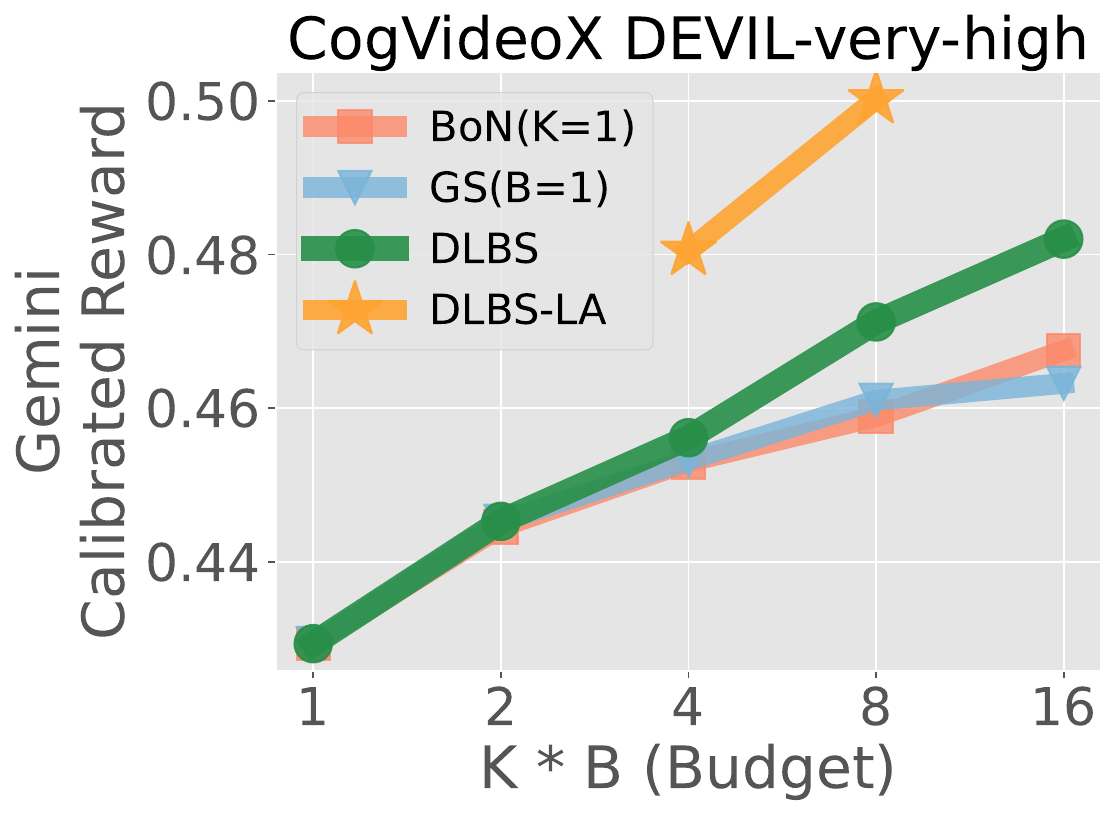}}
\subfigure{\includegraphics[width=0.225\textwidth]{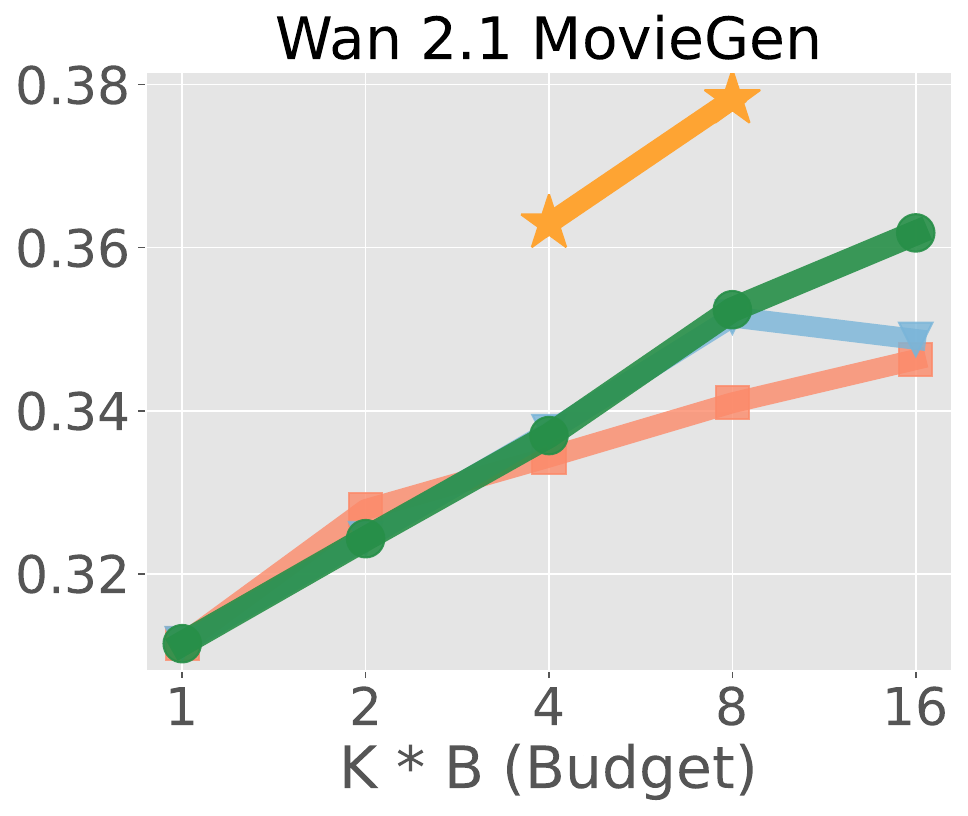}}
% \subfigure[VLM evaluation]{\includegraphics[width=0.44\textwidth]{figures/vlm_scaling.pdf}}
% \subfigure[Human evaluation]{\includegraphics[width=0.34\textwidth]{figures/human_eval.pdf}}
% \subfigure{\includegraphics[width=0.59\textwidth]{figures/diversity_scatter_5.pdf}}

\vskip -0.15in
\caption{
(\textbf{Left}) DLBS achieves alignment performance gains more efficiently than BoN and GS under the same number of function evaluations (NFE) or execution time.
Increasing the search budget provides larger improvements under an equivalent computational cost than scaling the number of diffusion steps (DLBS-T; $KB=8, T=100,200$).
Employing the LA estimator (DLBS-LA) further amplifies these gains, only with marginal overhead, yielding remarkably better efficiency than BoN or GS.
(\textbf{Right}) DLBS and DLBS-LA help the latest SoTA models, CogVideoX-5B~\citep{yang2024cogvideox} and Wan 2.1-14B~\citep{wan2025}, improve the  generated video quality.
%(\textbf{Left}) DLBS achieves alignment performance gains more efficiently than BoN and GS under the same number of function evaluations (NFE) or execution time.  Increasing the search budget provides larger performance improvements under an equivalent computational cost than scaling the number of diffusion steps (DLBS-T; $KB=8, T=100,200$). Moreover, employing the LA estimator (DLBS-LA) further amplifies these gains, with only marginal overhead yielding remarkably better alignment efficiency than BoN or GS.
%(\textbf{Right}) DLBASearch with CogVideoX-5B~\citep{yang2024cogvideox} and Wan 2.1-14B~\citep{wan2025}.
% (\textbf{Right}) 
% Alignment--diversity tradeoff ($KB = 32$), which is measured by the mean pairwise distance of ViCLIP embeddings. DLBS achieves high performance maintaining higher diversity.
}
% \vskip -0.15in
\vskip -0.05in
\label{fig:cost_and_advanced_models}
\end{figure}

We also analyze how alignment performance scales with different DLBS configurations under fixed computational budgets, using the number of function evaluations (NFE) and wall-clock time as cost measures. 
As shown in \autoref{fig:cost_and_advanced_models} (\textbf{Left}), DLBS consistently outperforms BoN and GS across both budgets, demonstrating superior efficiency in utilizing compute. 
Adding the LA estimator further amplifies this advantage, offering substantial performance gains with minimal overhead. 
In contrast, increasing the number of diffusion steps $T$ (DLBS-T; $KB=8$, $T\in\{100,200\}$) results in only marginal improvements despite the higher computational cost. 
Our results suggest a clear strategy for inference-time budget allocation: prioritize enabling the LA estimator and increasing the search budget 
$KB$ leads to substantial performance gains, while increasing the diffusion steps $T$ provides limited benefit relative to its computational cost.

\begin{figure}[t]
\vskip -0.1in
\centering
\subfigure{\includegraphics[width=0.39\textwidth]{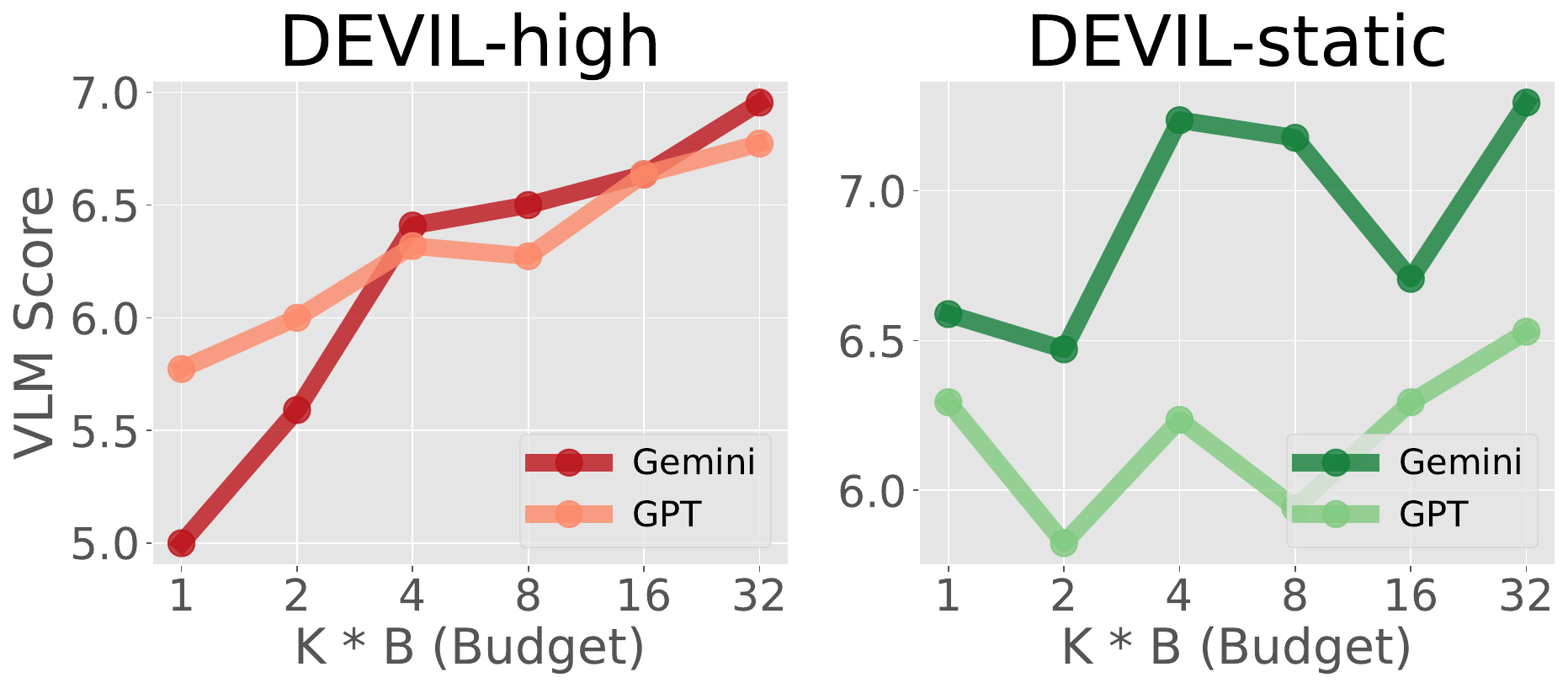}}
\subfigure{\includegraphics[width=0.60\textwidth]{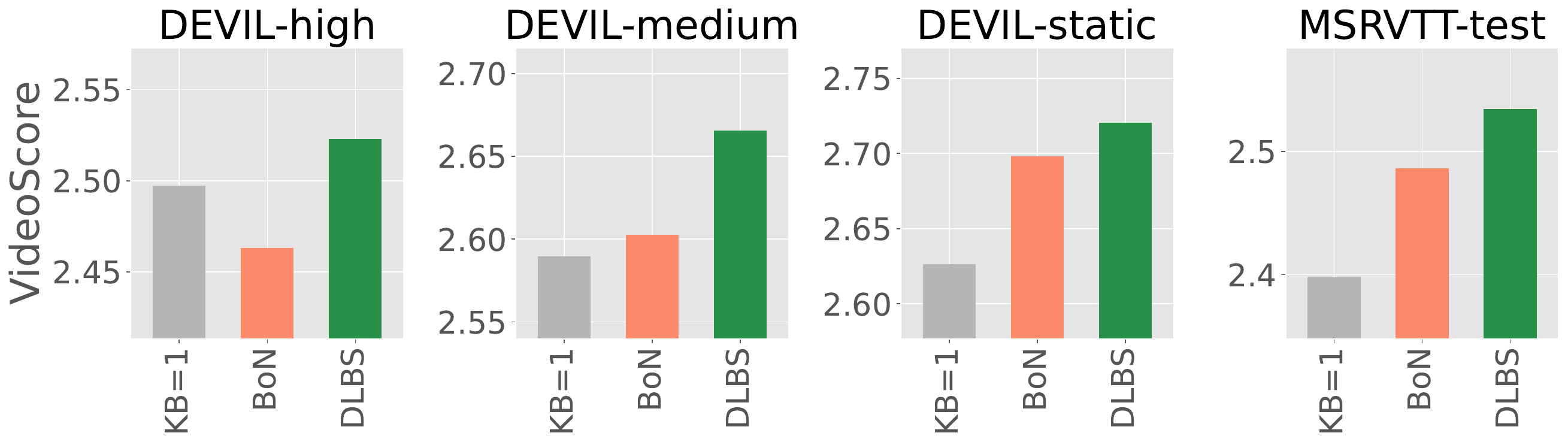}}
% \subfigure{\includegraphics[width=0.30\textwidth]{figures/human_eval_dlbs_bon.pdf}}
% \subfigure{\includegraphics[width=0.39\textwidth]{figures/vlm_scaling.pdf}}
% \subfigure{\includegraphics[width=0.285\textwidth]{figures/videoscor_average_only.pdf}}
% \subfigure{\includegraphics[width=0.305\textwidth]{figures/human_eval_mini.pdf}}

\vskip -0.15in
\caption{
(\textbf{Left}) DLBS on calibrated reward can improve the original preference feedback from VLMs. As in \autoref{fig:main}, we use each calibrated reward (\autoref{fig:weight_selection_gemini}) for DLBS and evaluate the quality with Gemini or GPT-4o.
(\textbf{Right}) DLBS on calibrated reward also improves another qualitative metric, the most, VideoScore~\citep{he2024videoscore}, which is not involved in a reward calibration.
% (\textbf{Right}) The pairwise comparisons between DLBS-LA ($KB=8$, $T’=6$, $\text{NFE}=2500$) and BoN ($KB=64$, $\text{NFE}=3200$) by three human raters, when DLBS-LA searches effectively for the best latents in a diffusion process.
% The pairwise comparisons between DLBS-LA and GS ($KB=1$) by three human raters, when DLBS-LA searches for the best latents in a diffusion process.
}
\vskip -0.2in
% \vskip -0.1in
\label{fig:evaluation_1}
\end{figure}

\subsection{Evaluation with AI and Human Feedback}
\label{sec:vlm_eval}
As discussed in Section~\ref{sec:reward_calibration}, we obtain a manageable reward function through the reward calibration, which reduces the cost for frequent evaluation queries in inference-time search. While DLBS efficiently improves the calibrated reward (\autoref{fig:main}; \textbf{Left}), a natural question is whether DLBS can improve an actual assessment by VLMs or humans by optimizing their calibrated rewards.
We first use each calibrated reward for DLBS, then evaluate the quality using discrete scores (from 1 to 10) from Gemini or GPT-4o.
% \autoref{fig:vlm_scaling} 
\autoref{fig:evaluation_1} (\textbf{Left}) demonstrates that DLBS maximizing calibrated rewards can improve the original preference feedback from VLMs, as we grow the search budget.

Next, we evaluate with VideoScore~\citep{he2024videoscore}, a metric trained on human judgments that evaluates videos across five quality criteria.
\autoref{fig:evaluation_1} (\textbf{Right}) shows that DLBS significantly improves the quantitative evaluation based on human evaluation.
Lastly, we perform pairwise comparisons between DLBS-LA ($KB=8$, $T’=6$, $\text{NFE}=2500$) and BoN ($KB=64$, $\text{NFE}=3200$) by three human evaluators (\autoref{fig:evaluation_2}; \textbf{Left}). 
The results confirm that, whatever models or prompts we choose, the quality of content generated by DLBS-LA consistently outperforms that of a baseline despite requiring fewer NFEs.
This emphasizes that our proposed method, integrating reward calibration and beam search in a latent space, effectively enhances perceptual video quality.
See  Appendix~\ref{sec:vlm_eval_appendix} for the details.
% Lastly, we perform pairwise comparisons between DLBS-LA and GS ($KB=1$) by three human evaluators (\autoref{fig:evaluation}; \textbf{Right}). 
% The results confirm that, whatever models or prompts we choose, the quality of content generated by DLBS-LA consistently outperforms that of a baseline.
% This emphasizes that our proposed method, integrating reward calibration and beam search in a latent space, effectively enhances perceptual video quality.
% See  Appendix~\ref{sec:vlm_eval_appendix} for the details.

% We also conducted evaluations using videoscore~\citep{he2024videoscore} (\autoref{fig:evaluation}; \textbf{Middle}) — a metric trained on human judgments and computable at 8 fps — as well as pairwise comparison by three human evaluators (\autoref{fig:evaluation}; \textbf{Right}).
% % (\autoref{fig:human_evaluation}). 
% These experiments confirmed that the content produced via our search strategy outperformed the content produced without search, both in score and in win rate. Furthermore, the results indicate that the higher the prompt dynamics (e.g. static $\rightarrow$ medium $\rightarrow$ high), the greater the win rate benefit from search, suggesting that search becomes increasingly meaningful.
% This highlights that our recipe for designing inference-time search, combining reward calibration and beam search, can improve video quality.

% \input{figure_and_table/evaluation}
% \input{figure_and_table/advanced_models}

\subsection{Scaling Model Parameters and Capabilities}
\label{sec:scaling_model_params}
\begin{figure}[t]
% \vskip -0.1in
\centering
% \subfigure{\includegraphics[width=0.39\textwidth]{figures/vlm_scaling.pdf}}
% \subfigure{\includegraphics[width=0.29\textwidth]{figures/videoscor_average_only.pdf}}
\subfigure{\includegraphics[width=0.33\textwidth]{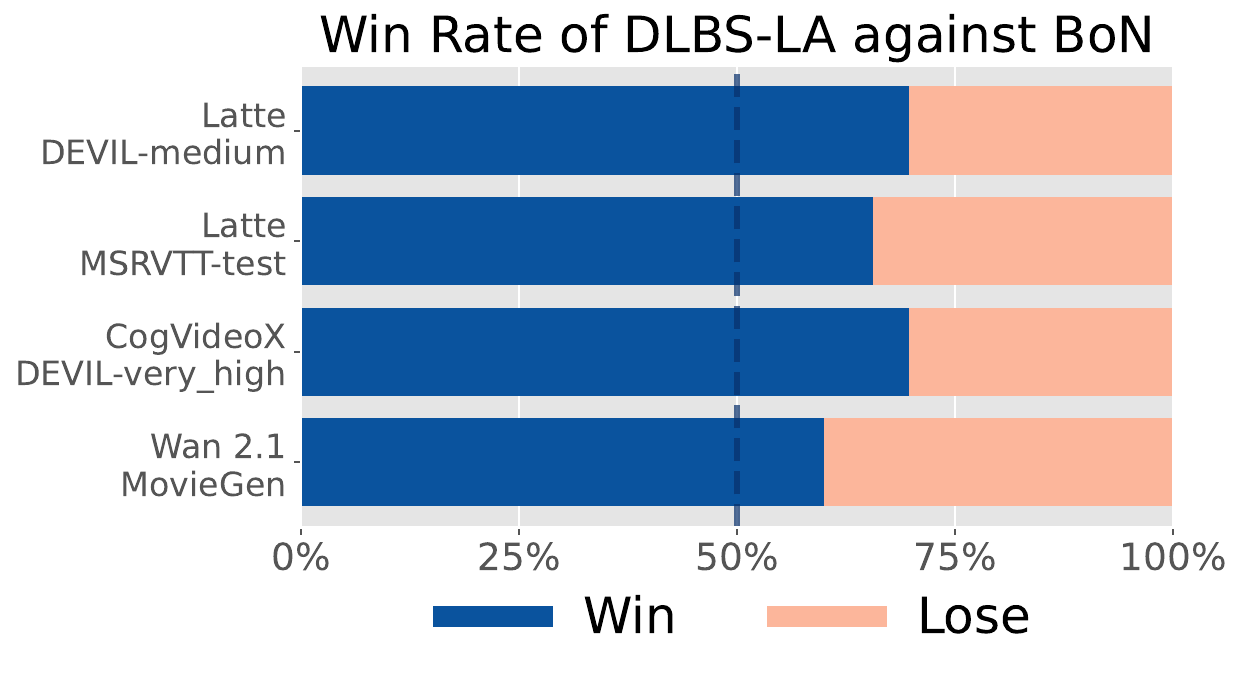}}
\subfigure{\includegraphics[width=0.66\textwidth]{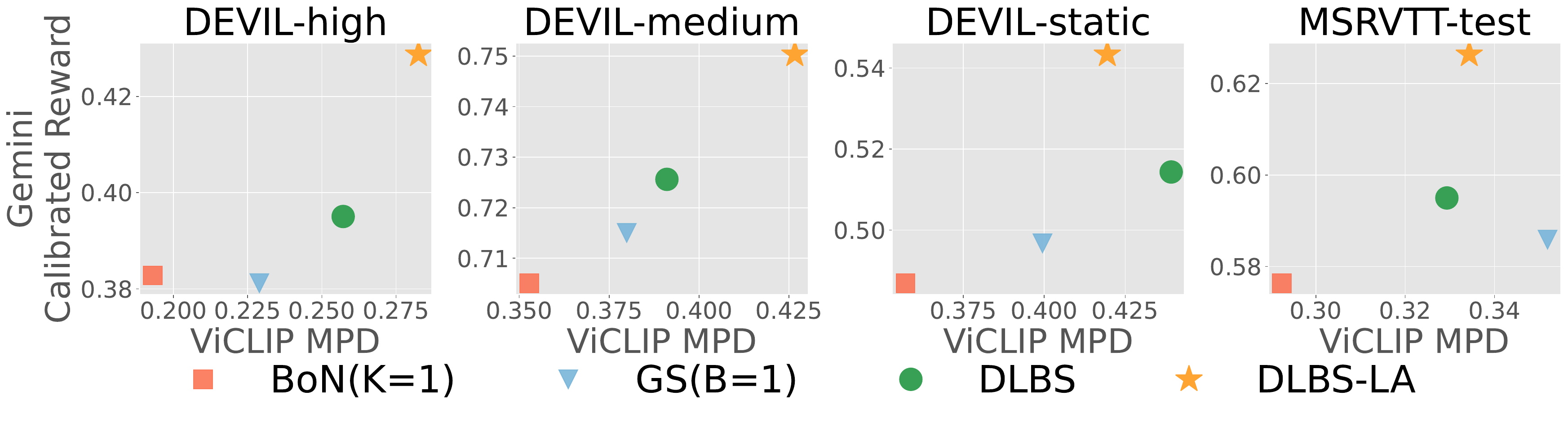}}
% \subfigure{\includegraphics[width=0.39\textwidth]{figures/vlm_scaling.pdf}}
% \subfigure{\includegraphics[width=0.285\textwidth]{figures/videoscor_average_only.pdf}}
% \subfigure{\includegraphics[width=0.305\textwidth]{figures/human_eval_mini.pdf}}

\vskip -0.15in
\caption{
(\textbf{Left}) The pairwise comparisons of DLBS-LA ($KB=8$, $T’=6$, $\text{NFE}=2500$) and BoN ($KB=64$, $\text{NFE}=3200$) by three human raters, when DLBS-LA searches effectively for the best latents in a diffusion process. 
(\textbf{Right}) Alignment--diversity tradeoff ($KB = 32$).
The mean pairwise distance (MPD) of ViCLIP embeddings is used as a measure of diversity. 
DLBS and DLBS-LA ($T' = 6$) achieve high performance while maintaining higher diversity.
% The pairwise comparisons between DLBS-LA and GS ($KB=1$) by three human raters, when DLBS-LA searches for the best latents in a diffusion process.
}
% \vskip -0.2in
\vskip -0.1in
\label{fig:evaluation_2}
\end{figure}

We scale up the base diffusion model from Latte to the latest SoTA models, such as CogVideoX-5B~\citep{yang2024cogvideox} and Wan 2.1-14B~\citep{wan2025}, and evaluate if DLBS can improve the generation quality from those larger models. 
Note that we use SDE-DPMSolver++~\citep{lu2022dpmsolverpp} for Wan 2.1 experiments.
Since base models are more capable here, we adopt more challenging prompts from DEVIL-very-high (22 prompts) and Movie Gen Video Bench~\citep{meta2024moviegen} (20 prompts). See Appendix~\ref{sec:prompts} and Appendix~\ref{sec:reward_challenging} for the details.
\autoref{fig:cost_and_advanced_models} (\textbf{Right}) shows that our methods achieve significant improvements in calibrated reward for both models. 
As shown in \autoref{fig:evaluation_2} (\textbf{Left}), human evaluation also supports that our proposed method could generally work well with any text-to-video models, even with more capable models in the future.
See Appendix~\ref{sec:long_scalability} for results with a maximum frame length, and Appendix~\ref{sec:large_models_full} for further results with CogVideoX and Wan 2.1.

\subsection{Alignment-Diversity Tradeoff}
\label{sec:diversity_eval}

Alignment for diffusion models can steer desirable outputs, but it is said that the diversity of generated samples or the performance of original models often degrade~\citep{lee2023aligning,wu2024twistedsmc}.
While inference-time search does not change or degrade the model itself, we here compare the diversity of samples among BoN, GS, DLBS, and DLBS-LA. 
We measure the sample diversity as the mean pairwise distance of ViCLIP~\citep{xu2021videoclip} embeddings (see \autoref{sec:detailed_diversity}).
\autoref{fig:evaluation_2} (\textbf{Right}) reveals that DLBS and DLBS-LA achieve high performance with higher diversity than BoN or GS. 
This exhibits a benefit from the wider possession and exploration of diffusion latents in DLBS and DLBS-LA.

\subsection{DLBS is Compatible with Finetuning}
\label{sec:dpo_dlbs}

\begin{figure}[t]
  \centering
  % 左側: 表
  \begin{minipage}[t]{0.47\linewidth}
    \vspace{0pt}
    \centering
    \captionof{table}{Performance of DLBS with DPO finetuned VideoCrafter2 on DEVIL-high and MSRVTT-test datasets. 
    While DPO alone yields marginal improvements, combining it with DLBS leads to notable gains, demonstrating the compatibility of inference-time search with fine-tuning approaches.}
    \setlength{\tabcolsep}{1.75pt}
    \begin{tabular}{lcc}
      \toprule
      \textbf{Method} & \textbf{DEVIL-high} & \textbf{MSRVTT-test} \\
      \midrule
      VideoCrafter2        & 0.337 & 0.555 \\
      + DPO                & 0.335 & 0.556 \\
      + DPO \& DLBS        & \textbf{0.359} & \textbf{0.576} \\
      \bottomrule
    \end{tabular}
    \label{tab:dpo_dlbs}
  \end{minipage}
  \hfill
  % 右側: 図
  \begin{minipage}[t]{0.5\linewidth}
    \vspace{0pt}
    \centering
    \includegraphics[width=0.49\textwidth]{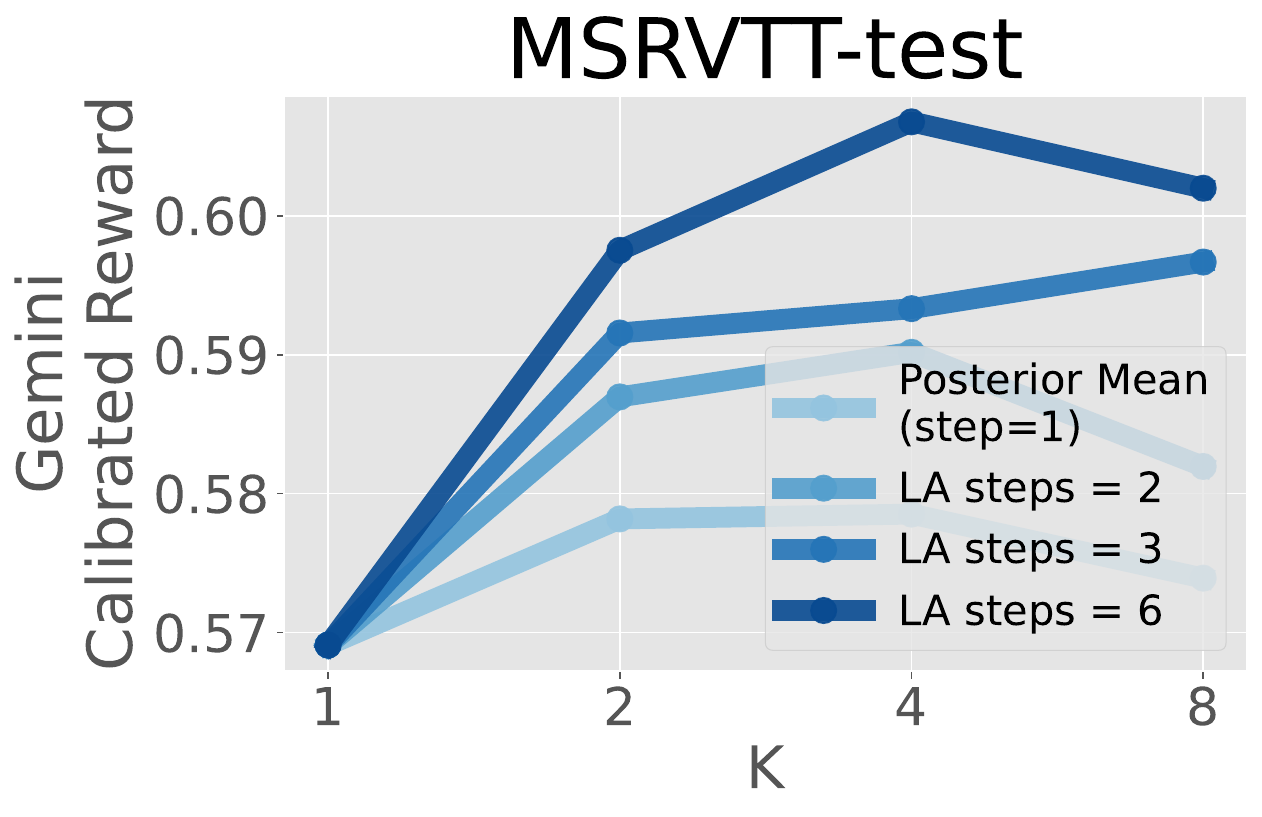}
    \includegraphics[width=0.47\textwidth]{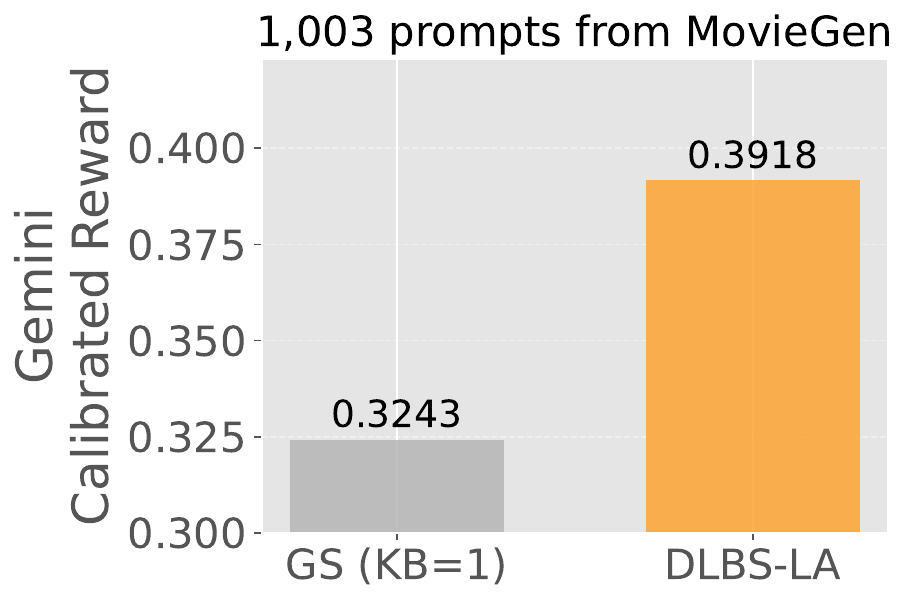}
    \vskip -0.05in
    \captionof{figure}{(\textbf{Left}) Comparison of different LA steps $T'$ on MSRVTT-test ($KB=8$). The performance improves as the number of LA steps increases. 
    (\textbf{Right}) Perceptual quality comparison on a large and complex prompt set (1,003 prompts from Movie Gen Video Bench~\citep{meta2024moviegen}). DLBS-LA can generalize to diverse prompts.}
    \label{fig:scaling_ablation}
  \end{minipage}
  \vskip -0.15in
\end{figure}

In image generation, \citet{ma2025inferencetime} has shown that allocating additional computation at inference time can be more effective than relying solely on post-training approaches. 
We find a similar trend in video generation. To examine this, we apply a representative fine-tuning method, VideoDPO~\citep{liu2024videodpo}, to VideoCrafter2~\citep{chen2024videocrafter2}. 
As shown in Table~\ref{tab:dpo_dlbs}, VideoDPO alone brings negligible improvement over the baseline. However, when combined with DLBS, the performance increases substantially on both DEVIL-high and MSRVTT-test.
These results indicate that DLBS is complementary to post-training methods, enabling further performance gains even after fine-tuning.
See Appendix~\ref{sec:gradient_based_search} and Appendix~\ref{sec:gradient_free_search} for results comparing DLBS with other alignment methods.

\subsection{Ablation Study}
\label{sec:ablation_study}
% \subsection{Scaling Lookahead Steps for Reward Estimate}
% \label{sec:lookahead_result}
% \input{figure_and_table/lookahead_fig}
\textbf{Lookahead Steps for Reward Estimate}~~We scale up LA steps $T'$ to obtain an accurate reward estimate.
We use MSRVTT-test ($KB=8$) and Gemini reward for experiments.
\autoref{fig:scaling_ablation} (\textbf{Left})  shows that as the number of LA steps increases, the performance improves more. Even $T'=2, 3$ significantly outperforms the posterior mean, which is often used in prior works~\citep{kim2024free2guide,huang2024scg,singhal2025general}.
This is because the sub-optimal performance of inference-time search comes from the approximation errors, and LA estimator can notably reduce them.
As shown in \autoref{fig:main} (\textbf{Left}), DLBS-LA $(KB=8, T'=6)$ achieves comparable or even outperforming results with DLBS $(KB=32)$.
It is quite beneficial to spend a computation to estimate the reward accurately.
See Appendix~\ref{sec:lookahead_ablation} for further discussions.
In addition, an ablation study on diffusion steps is shown in Appendix~\ref{sec:denoising_steps_ablation}.
\textbf{A Large and Complex Prompt Set}~~To confirm the robustness of our approach on a large and highly diverse prompt set, we compare GS (\(KB{=}1\)) with DLBS-LA (\(KB{=}8,\;T'{=}6\)) with all the prompts in Movie Gen Video Bench~\citep{meta2024moviegen}, which comprises 1,003 prompts (\autoref{fig:scaling_ablation}; \textbf{Right}).
We observe that DLBS-LA consistently delivered substantially higher alignment rewards, demonstrating that DLBS generalizes effectively to complex prompt distributions.  
Additionally, we also assess reward transferability by applying weights calibrated on DEVIL-high and DEVIL-medium to MSRVTT-test prompts, consistently improving scores (see Appendix~\ref{sec:generalization}).
% We also quantitatively confirm that DLBS-LA preserves generative diversity in Appendix~\ref{sec:diversity_eval}.

\section{Related Works}
% \textbf{Aligning Diffusion Models via Sampling}~~
Classifier guidance~\citep{dhariwal2021diffusion,ho2022cfg} has been the most popular to enhance text-content alignment.
On top of that, recent works~\citep{huang2024scg,li2024derivative} leverage reward or external feedback at inference time by selecting better latents~\citep{wallace2023endtoend}, which probably achieve higher rewards during the reverse process.
\citet{kim2025das} propose twisted SMC~\citep{wu2024twistedsmc} with reward gradient, which is not suitable for non-differentiable feedback and for domains such as video, where reward gradient needs a huge memory cost.
Gradient-free methods~\citep{zheng2024ensemblekalman,ma2025inferencetime} such as SMC~\citep{singhal2025general} or greedy search~\citep{kim2024free2guide} often exhibit sub-optimal results affected by inaccurate reward estimates from noisy latents.
\citet{yeh2024sampling} uses ODE to estimate the reward, but it highly depends on Karras sampler~\citep{karras2022elucidating} to avoid numerical instability.
In contrast, we address the error propagation from inaccurate reward estimates with beam search and lookahead estimator via deterministic DDIM, which is more popular and stable. Our methods work more scalably when allocating more computation budget at inference time. See \autoref{sec:extended_related_work} for further related works.
% While there are many possible implementations to realize them, their scalability or feasibility are often limited depending on the modality of diffusion models.

% \autoref{sec:extended_related_work} discusses alignment via finetuning and evaluation of video generation.
% Please refer to \autoref{sec:extended_related_work} where we discuss alignment via finetuning and evaluation of text-to-video generation.

\section{Discussion and Limitation}
\label{sec:limitation}
% Our reward calibration assumes that VLMs work as a proxy of human evaluation, and we demonstrate both qualitative and quantitative improvement of the video quality. It would be important to extend it to human feedback, not limited to VLMs.
Our reward calibration assumes that VLMs serve as a proxy for human evaluation, and we demonstrate both qualitative and quantitative improvements in video quality through evaluations by VLMs and human raters. 
In future work, incorporating more specialized and accurate evaluators (e.g., reward models that focus on physical laws~\citep{bansal2024videophy}) could enable a more fine-grained analysis.
In practice, we often do implicit or explicit best-of-N sampling for video generation. In contrast, DLBS-LA exhibits much better computational efficiency. Spending more computation at inference time significantly improves perceptual quality, but it is orthogonal compared to speeding up the sampling process via distillation~\citep{meng2022distill,song2023consistency}, architecture changes~\citep{oshima2024ssmvdm}, or parallel sampling~\citep{shih2023parallel}. We believe both high-quality and speedy sampling have practical needs and should be balanced.

%Our reward calibration assumes that VLMs work as a proxy of human evaluation, and we demonstrate both qualitative and quantitative improvement of the video quality. It would be important to extend it to human feedback not limited to VLMs.
%
%While spending more computation at inference time significantly improves perceptual quality, the generation definitely requires more time. It is orthogonal compared to speeding up the sampling process via distillation~\citep{meng2022distill,song2023consistency}, architecture changes~\citep{oshima2024ssmvdm}, or parallel sampling~\citep{shih2023parallel}. We believe both high-quality and speedy sampling have practical needs and should be balanced.
%
%As discussed in Section~\ref{sec:diversity_eval}, inference-time alignment reduces sample diversity in diffusion process, although we do not change or degrade the model itself in contrast to finetuning.
% It would be an interesting direction to measure the effect of diversity in terms of model performance, and to consider a method to maintain the diversity yet to improve the performance.

\section{Conclusion}
This paper studies which metrics we should optimize and how to optimize them for better text-to-video generation.
We point out that feedback from humans or capable VLMs reflects multiple dimensions of video quality, so optimizing an existing metric alone is insufficient; rather, we should calibrate the reward by combining.
Our DLBS with LA estimator reduces the error propagation from the inaccurate reward estimate.
We demonstrate that DLBS is the most scalable, efficient, and robust inference-time search that significantly improves video quality under the same computational costs.
We hope our work encourages more uses of inference-time computation for text-to-video models.

\clearpage

\section*{Acknowledgements}
We thank Po-Hung Yeh, Shohei Taniguchi, Kuang-Huei Lee, Arnaud Doucet, Heiga Zen, Robin Scheibler, and Yusuke Iwasawa for their support and helpful discussion on the initial idea of this work.
HF was supported by JSPS KAKENHI Grant Number JP22J21582 (by March 2025), and MS was supported by JSPS KAKENHI Grant Number JP23H04974.

\bibliography{neurips_2025}
\bibliographystyle{unsrtnat}

%%%%%%%%%%%%%%%%%%%%%%%%%%%%%%%%%%%%%%%%%%%%%%%%%%%%%%%%%%%%%%%%%%%%%%%%%%%%%%%
%%%%%%%%%%%%%%%%%%%%%%%%%%%%%%%%%%%%%%%%%%%%%%%%%%%%%%%%%%%%%%%%%%%%%%%%%%%%%%%
% APPENDIX
%%%%%%%%%%%%%%%%%%%%%%%%%%%%%%%%%%%%%%%%%%%%%%%%%%%%%%%%%%%%%%%%%%%%%%%%%%%%%%%
%%%%%%%%%%%%%%%%%%%%%%%%%%%%%%%%%%%%%%%%%%%%%%%%%%%%%%%%%%%%%%%%%%%%%%%%%%%%%%%
\clearpage
\appendix

\section{Broader Impacts}
\label{sec:broader_impacts}
Our work contributes to the progress of text-to-video by focusing on improving the perceptual quality and fidelity of generated videos, specifically addressing issues like unnatural movement, deformation, and temporal inconsistencies, through the inference-time alignment algorithm. Such advancements hold immense potential for revolutionizing creative fields and enabling new applications in gaming, filmmaking, and robotics.

On the other hand, the ability to generate highly realistic videos raises concerns about the potential for misuse in creating deceptive content, including deepfakes and misinformation. 
Like other generative models, such as large language models, text-to-video models, and their inference-time search, may inherit and amplify biases present in the training data due to the misalignment.
This might lead to the generation of videos that perpetuate harmful stereotypes or underrepresent certain groups.

Lastly, by focusing on inference-time alignment, our method promotes more use of computational resources at test time. On one side, this may increase the environmental footprint for running large generative models and on the other side, our detailed recipe can contribute to designing efficient use of resources and reducing the footprint associated with training. We believe that discussing this aspect is crucial as the scale of these models continues to grow.

\section{Implementation Details}
\label{sec:implementation_details}

\textbf{Code}~~
Our implementation for the experiments are available at \url{https://github.com/shim0114/T2V-Diffusion-Search}.

\textbf{Models}~~
Our experiments cover three text-to-video diffusion models.

\begin{itemize}[leftmargin=0.5cm,topsep=0pt,itemsep=0pt]
  \item \textbf{Latte}~\citep{ma2024latte}: a T5-conditioned latent diffusion transformer with 1.1\,B parameters, built on PixArt-$\alpha$~\citep{raffel2020t5, chen2023pixartalpha}.
  \item \textbf{CogVideoX}~\citep{yang2024cogvideox}: a larger DiT-based model with 2B or 5B parameters.
  We mainly used CogVideoX-5B.
  \item \textbf{Wan~2.1}~\citep{wan2025}: a DiT-based flow model with 1.3B or 14B parameters.
  We use Wan~2.1-1.3B for reward calibration and Wan~2.1-14B for inference-time search experiments.
\end{itemize}

\textbf{Hyperparameters}~~
\begin{itemize}[leftmargin=0.5cm,topsep=0pt,itemsep=0pt]
  \item \textbf{Latte}: DDIM scheduler with a linear noise schedule
        $(\beta_{\text{start}}=1.0\times10^{-4},\ \beta_{\text{end}}=2.0\times10^{-2})$  
        and classifier-free guidance scale $w_{\text{cfg}}=7.5$.
  \item \textbf{CogVideoX}: DDIM scheduler with the original settings and $w_{\text{cfg}}=6.0$. Owing to computational resource constraints, we limit the frame length to 17 per sample.\footnotemark[3]
  \item \textbf{Wan~2.1}: DPMSolver++ with guidance scale $w_{\text{cfg}}=5.0$. For the same computational reasons, the spatial resolution is limited to $832\times480$ and the frame length to 33 per sample.\footnotemark[3]
\end{itemize}

\footnotetext[3]{The experiments reported in Appendix~\ref{sec:long_scalability} were conducted with a different number of frames.}

\textbf{Hardware configuration}~~
\begin{itemize}[leftmargin=0.5cm,topsep=0pt,itemsep=0pt]
  \item \textbf{Latte}: FP16 inference on a single NVIDIA A100 (40\,GB), batch size~1.
  \item \textbf{CogVideoX}: BF16 inference on a single NVIDIA A100 (40\,GB), batch size~1.
  \item \textbf{Wan~2.1}: FP16 inference on four NVIDIA H100s (80\,GB each), batch size~1.
\end{itemize}

\textbf{AI-feedback endpoints}~~
We use API endpoints: \texttt{gemini-1.5-pro-002} and \texttt{gpt-4o-2024-11-20}.

\clearpage
\section{Details of Metric Rewards}
\label{sec:metric_reward}

\textbf{Subject Consistency}~~We adopt the subject consistency metric proposed in VBench~\citep{huang2023vbench} to quantify how consistently a subject is depicted across consecutive video frames. 
Concretely, for each frame \(i\) in a video, we extract a feature representation \(\mathbf{d}_i\) using DINO~\citep{caron2021dino} with a ViT-B/16~\citep{dosovitskiy2020vit} backbone. 
Let \(\langle \mathbf{d}_i, \mathbf{d}_j \rangle\) denote the cosine similarity between the features \(\mathbf{d}_i\) and \(\mathbf{d}_j\). Then, VBench defines the subject consistency metric as follows:
\begin{equation}
R_{\text{subject}} = \frac{1}{T - 1} \sum_{t=2}^{T} \frac{1}{2}\bigl(\langle \mathbf{d}_1,\, \mathbf{d}_t\rangle + \langle \mathbf{d}_{t-1},\, \mathbf{d}_t\rangle\bigr).
\end{equation}
DINO, which is trained in a self-supervised manner using unlabeled images and image augmentations, does not explicitly suppress intra-class variations. 
As a result, it remains particularly sensitive to identity shifts within the same subject, making it well-suited for evaluating subject consistency across frames.
% This metric quantifies how consistently the subject appears across video frames. 
% Using DINO~\citep{caron2021dino} with a ViT-B/16~\citep{dosovitskiy2020vit} backbone, we extract frame-wise features and compute their similarities. 
% By comparing each frame’s features to those of the first and preceding frames, and averaging these similarities.

\textbf{Motion Smoothness}~~We adopt the frame-interpolation-based metric originally proposed in VBench~\citep{huang2023vbench} to assess whether a generated video’s motion is smooth and physically plausible. 
In particular, this metric leverages the motion prior from AMT~\citep{li2023amt}, employing its AMT-S variant for frame reconstruction. 
Concretely, let \(\bigl[\mathbf{f}_0, \mathbf{f}_1, \mathbf{f}_2, \ldots, \mathbf{f}_{2n}\bigr]\) denote the frames of a generated video. We remove each odd-numbered frame to obtain a lower-frame-rate sequence \(\bigl[\mathbf{f}_0, \mathbf{f}_2, \mathbf{f}_4, \ldots, \mathbf{f}_{2n}\bigr]\), and rely on AMT-S to reconstruct the missing frames \(\bigl[\hat{\mathbf{f}}_1, \hat{\mathbf{f}}_3, \ldots, \hat{\mathbf{f}}_{2n-1}\bigr]\).
We then compute the Mean Absolute Error (MAE) between these reconstructed frames and the original odd-numbered frames, denoting this measure by \(R_{\text{smoothness}}\). 
Finally, following the normalization scheme introduced in VBench, we define:
\begin{equation}
R_{\text{smoothness-norm}} = \frac{255 - R_{\text{smoothness}}}{255},  
\end{equation}
which ensures that the final score lies in the range \([0, 1]\), with higher values indicating smoother motion.
This measure leverages the motion prior in AMT to evaluate whether the generated video’s motion is smooth and physically plausible. 
We remove each odd-numbered frame, then use AMT-S to reconstruct those frames based on short-term motion assumptions. 

\textbf{Dynamic Degree}~~This measure quantifies the overall magnitude of dynamic object movement. 
Let \(T\) be the total number of frames in the generated video. 
For each pair of consecutive frames \(t\) and \(t+1\), we estimate the optical flow \(\mathbf{v}_t\) using RAFT~\citep{teed2020raft}, compute its norm \(\|\mathbf{v}_t\|\), and sum these values across all frames:
\begin{equation}
R_{\text{dynamics}} = \sum_{t=1}^{T-1}\|\mathbf{v}_t\|.
\end{equation}
We then apply a logarithmic transformation to \(R_{\text{dynamics}}\) and divide by 16:
\begin{equation}
R_{\text{dynamics-rescaled}} = \frac{\log\bigl(R_{\text{dynamics}}\bigr)}{16}.
\end{equation}
This rescaling helps ensure that the value range of \(R_{\text{dynamics-scaled}}\) is roughly comparable to other metrics in our evaluation.

\textbf{Aesthetic Quality}~~This criterion evaluates compositional rules, color harmony, and overall artistic merit on a per-frame basis. Concretely, for each frame \(i\) in a video, we extract a CLIP image embedding \(\mathbf{c}^{image}_i\) using the CLIP ViT-L/14 model~\citep{radford2021clip}. 
We then feed \(\mathbf{c}^{image}_i\) into the LAION aesthetic predictor~\citep{laion2022aesthentic}, which assigns a raw rating \(r_i \in [0,10]\). To normalize these scores to the \([0,1]\) range, we set 
\begin{equation}
r'_i = \frac{r_i}{10}.
\label{eq:aesthetic_rescale}
\end{equation}
Let \(T\) be the total number of frames. The final aesthetic reward is then obtained by taking the average of the normalized ratings across all frames:
\begin{equation}
R_{\text{aesthetic}} = \frac{1}{T} \sum_{i=1}^{T} r'_i.
\label{eq:aesthetic_average}
\end{equation}
Because the LAION aesthetic predictor leverages CLIP embeddings instead of raw images, it captures higher-level features related to composition, color harmony, and artistic appeal.
% This criterion measures compositional rules, color harmony, and the overall artistic merit of each video frame. 
% We use the LAION aesthetic predictor~\citep{laion2022aesthentic} to assign a 0–10 rating to each frame, then linearly normalize those ratings to 0–1. 
% The final aesthetic score is the mean of these normalized scores across all frames.

\textbf{Imaging Quality}~~This indicator assesses low-level distortions (e.g., over-exposure, noise, blur) in each generated frame. 
We adopt the MUSIQ predictor~\citep{ke2021musiq}, trained on the SPAQ dataset~\citep{fang2020spaq}. 
The frame-wise score is normalized to [0, 1] by dividing by 100, and the final video score is the mean of these normalized values across all frames in the same way as Equation~\ref{eq:aesthetic_rescale} and Equation~\ref{eq:aesthetic_average}.

\textbf{Text-Video Consistency}~This measure captures how closely a generated video's content aligns with its text prompt. 
We employ ViCLIP~\citep{xu2021videoclip}, a model pre-trained on a 10M video-text dataset and fine-tuned to handle temporal relationships, to embed both the video frames and the text. 
Since ViCLIP computes embeddings from 8-frame inputs, we sample 8 frames from each video.
Let \(\mathbf{v}^{\text{video}}\) denote the resulting video embedding and \(\mathbf{v}^{\text{text}}\) denote the text embedding. 
We then define the final alignment score as the cosine similarity between these embeddings:
\begin{equation}
R_{\text{tv-consistency}} = \langle\mathbf{v}^{\text{video}}, \mathbf{v}^{\text{text}} \rangle
\label{eq:tv-consistency}
\end{equation}
% This measure captures how closely a generated video's content aligns with its text prompt. 
% We employ ViCLIP~\citep{xu2021videoclip}, a model pre-trained on a 10M video-text dataset and fine-tuned to handle temporal relationships, to embed both video frames and the text. 
% The final alignment score is obtained by evaluating the similarity between these embeddings.

\section{Details of Sample Diversity}
\label{sec:detailed_diversity}
% Alignment for diffusion models can steer desirable outputs, but it is said that the diversity of generated samples or the performance of original models often degrade~\citep{lee2023aligning,wu2024twistedsmc}.
% While inference-time search does not change or degrade the model itself, we here compare the diversity of samples among BoN, GS, and DLBS. 
We measure the sample diversity as the mean pairwise distance of ViCLIP~\citep{xu2021videoclip} embeddings to quantify the diversity in videos, inspired by the approach for evaluating diversity in images~\citep{kim2025das}. 
Specifically, given \( N \) generated video samples, we first extract ViCLIP embeddings \(\mathbf{v}^{\text{video}, (i)}\) for each sample \( i \). The pairwise diversity score is then computed as the mean pairwise distance:

\begin{equation}
D_{\text{video-diversity}} = \frac{1}{N(N-1)} \sum_{i \neq j} \left( 1 - \langle \mathbf{v}^{\text{video} , (i)}, \mathbf{v}^{\text{video}, (j)} \rangle \right).
\end{equation}

Here, \(\langle \mathbf{v}^{\text{video}, (i)}, \mathbf{v}^{\text{video}, (j)} \rangle\) denotes the cosine similarity between the ViCLIP embeddings of two generated videos \( i \) and \( j \). 
This formulation is similar to Equation~\ref{eq:tv-consistency}, but in the case of pairwise distance computation, we take the pairwise mean of \( 1 - \text{(cosine similarity)} \) to obtain a diversity measure.

% \section{Details of Sample Diversity}
% \label{sec:detailed_diversity}
% We compute the pairwise distances of ViCLIP~\citep{xu2021videoclip} embeddings for video generation to quantify the diversity in videos, inspired by the approach for evaluating diversity in images~\citep{kim2025das}. 
% Specifically, given \( N \) generated video samples, we first extract ViCLIP embeddings \(\mathbf{v}^{\text{video}, (i)}\) for each sample \( i \). The pairwise diversity score is then computed as the mean pairwise distance:

% \begin{equation}
% D_{\text{video-diversity}} = \frac{1}{N(N-1)} \sum_{i \neq j} \left( 1 - \langle \mathbf{v}^{\text{video} , (i)}, \mathbf{v}^{\text{video}, (j)} \rangle \right).
% \end{equation}

% Here, \(\langle \mathbf{v}^{\text{video}, (i)}, \mathbf{v}^{\text{video}, (j)} \rangle\) denotes the cosine similarity between the ViCLIP embeddings of two generated videos \( i \) and \( j \). 
% This formulation is similar to Equation~\ref{eq:tv-consistency}, but in the case of pairwise distance computation, we take the pairwise mean of \( 1 - \text{(cosine similarity)} \) to obtain a diversity measure.

\clearpage
\section{Algorithms with Continuous-time Diffusion Process}
\label{sec:dlbs_for_dpmsolver++}
In this section, we present our algorithms for a continuous-time diffusion process.
For Wan 2.1 \citep{wan2025}, we integrated the proposed search algorithm into DPMSolver++ \citep{lu2022dpmsolverpp}, a widely used continuous-time solver.
Algorithms \ref{alg:dlbs_dpm} and \ref{alg:lookahead_dpm} present the pseudocode.
Although we present the first-order variant for clarity, the procedure extends straightforwardly to higher-order formulations.
Throughout this section, we adopt the notation of \citet{lu2022dpmsolverpp}.
\begin{algorithm}[H]
% \begin{algorithm}[tb]
\caption{Diffusion Latent Beam Search (DLBS) with SDE-DPMSolver++}
\renewcommand{\algorithmicrequire}{\textbf{Input:}}
\label{alg:dlbs_dpm}
\begin{algorithmic}[1]
\REQUIRE signal prediction latent diffusion model $z_{\theta}$, reward function $r'$, time steps $\{t_s\}_{s=0}^{M}$, noise scheduling parameter $\{\alpha_{t_s}\}_{s=0}^{M}, \{\sigma_{t_s}\}_{s=0}^{M}$, number of beams $B$, number of candidates $K$
\STATE $\textbf{z}^{1}_{t_0}, \cdots, \textbf{z}^{B}_{t_0} \sim \mathcal{N}(\mathbf{0}, \mathbf{I})$ ~ {\color{richlilac} $\triangleright$ Initial $B$ beams} \alglinelabel{op:init_dpm}
\FOR{$s=1$ {\bfseries to} $M$}
    \FOR{$j=1$ {\bfseries to} $B$}
        \STATE {\color{richlilac} $\triangleright$ Update one step to produce $\mathbf{z}_{t_{s}}^j$}
        \STATE $\textbf{z}^j_{t_{s}} = \frac{\sigma_{t_{s}}}{\sigma_{t_{s-1}}}e^{-h}\textbf{z}_{t_{s-1}}^j + \alpha_{t_{s}}(1-e^{-2h})z_\theta(\textbf{z}_{t_{s-1}}^j) $
    \ENDFOR
    \IF{$s<M$}
        \FOR{$j=1$ {\bfseries to} $B$}
            \STATE {\color{richlilac} $\triangleright$ Sample $K$ next candidate latents}
            \STATE $\mathbf{z}_{t_s}^{ij} = \mathbf{z}^{j}_{t_s} + \sigma_t\sqrt{e^{-2h}-1} \epsilon_{t_{s-1}}^i$ with $\epsilon_{t_{s-1}}^1, ..., \epsilon_{t_{s-1}}^K \sim \mathcal{N}(\mathbf{0}, \mathbf{I})$ \alglinelabel{op:sampling_dpm}
            \STATE {\color{richlilac} $\triangleright$ Estimate the clean sample from noisy latent}
            \STATE $\hat{\mathbf{z}}_{t_M|t_s}^{ij} = \frac{\sigma_{t_M}}{\sigma_{t_s}}\mathbf{z}_{t_s}^{ij} - \alpha_{t_M}(e^{-h}-1)z_\theta(\mathbf{z}^{ij}_{t_s}))$ \alglinelabel{op:posterior_dpm}
        \ENDFOR
        \STATE {\color{richlilac} $\triangleright$ Search $B$ higher-reward beams from $KB$ latents}
        \STATE $\texttt{budget} := \{(\mathbf{z}_{t_s}^{11},\hat{\mathbf{z}}_{t_M|t_s}^{11}),\cdots,(\mathbf{z}_{t_s}^{KB},\hat{\mathbf{z}}_{t_M|t_s}^{KB}) \}$
        \FOR{$j'=1$ {\bfseries to} $B$}
            \STATE $\mathbf{z}_{t_s}^{j'} = \argmax_{\mathbf{z}^{ij}_{t_s}\in\texttt{budget}}~r'(\hat{\mathbf{z}}^{ij}_{t_M|t_s})$ \alglinelabel{op:eval_dpm}
            \STATE $\texttt{budget} = \texttt{budget} \setminus \{(\mathbf{z}_{t_s}^{j'},\hat{\mathbf{z}}^{\argmax}_{t_M|t_s})\} $ \alglinelabel{op:bs_dpm}
        \ENDFOR
        \STATE $j \in \{1,\cdots,B\} \leftarrow j'$ ~ {\color{richlilac} $\triangleright$ Reset selected $B$ indices}
    \ENDIF
\ENDFOR
\STATE {\bfseries return: } $\mathbf{z}_{t_M} = \argmax_{\mathbf{z}^{j}_{t_M}\in\{\textbf{z}^1_{t_M}, \cdots,\textbf{z}^B_{t_M}\}}~r'(\mathbf{z}^{j}_{t_M})$
\end{algorithmic}
\end{algorithm}

\begin{algorithm}[H]
% \begin{algorithm}[tb]
\caption{Lookahead (LA) with DPMSolver++}
\renewcommand{\algorithmicrequire}{\textbf{Input:}}
\label{alg:lookahead_dpm}
\begin{algorithmic}[1]
\REQUIRE signal prediction latent diffusion model $z_{\theta}$, current diffusion latent $\textbf{z}_{t_s}$, number of lookahead steps $M' (<<M)$
\STATE {\color{richlilac} $\triangleright$ Run $M'$-step deterministic DPMSolver++  starting from $\textbf{z}_{t_s}$}
% \STATE Interpolate time steps $\tilde{t}(s) \in \{\lfloor\frac{T'}{T'}(t-1)\rfloor, \lfloor\frac{T'-1}{T'}(t-1)\rfloor, \cdots, \lfloor\frac{s}{T'}(t-1)\rfloor, \cdots , \lfloor\frac{1}{T'}(t-1)\rfloor\}$
\STATE $\tilde{s}(u) \in \{s, \dots,\lfloor\frac{M'-u}{M'}s+\frac{u}{M'}M\rfloor,\dots , \lfloor\frac{1}{M'}s+\frac{M'-1}{M'}M\rfloor,M\}$
% \STATE $\tilde{t}(s) \in \{\lfloor\frac{T'}{T'}(t-1)\rfloor, \cdots, \lfloor\frac{s}{T'}(t-1)\rfloor, \cdots , \lfloor\frac{1}{T'}(t-1)\rfloor\}$
\STATE Select new lookahead noise schedule $\{\tilde{\alpha}_{t_u}\}_{u=0}^{M'}$ for \textbf{$M'$-step interpolation} of the rest of original $\{\alpha_{t_{s'}}\}_{s'=0}^{M}$
\STATE $\textbf{z}_{t_{\tilde{s}(0)}} := \textbf{z}_{t_s}$
\FOR{$u=1$ {\bfseries to} $M'$}
    \STATE $\tilde{\textbf{z}}_{t_{\tilde{s}(u)}|t_{\tilde{s}(u-1)}} = \frac{\sigma_{t_{\tilde{s}(u)}}}{\sigma_{t_{\tilde{s}(u-1)}}}\mathbf{z}_{t_{\tilde{s}(u-1)}}^{ij} - \alpha_{t_{\tilde{s}(u)}}(e^{-h}-1)z_\theta(\mathbf{z}^{ij}_{t_{\tilde{s}(u-1)}}))$
\ENDFOR
\STATE {\bfseries return: } $(\textbf{z}_{t_s}, \tilde{\textbf{z}}_{t_M|t_{\tilde{s}(0)}})$ {\color{richlilac} $\triangleright$ Latent and LA estimator}
\end{algorithmic}
\end{algorithm}

\clearpage
\section{Theoretical Analysis on Lookahead Estimator} 
\label{sec:lookahead_proof}
Consider the Lookahead estimator described in Algorithm~\ref{alg:lookahead}, which obtains the state \(\tilde{\mathbf{z}}_{0 \mid \tilde{t}(0)}\) by performing \(T'\) steps of DDIM (or another diffusion-based sampling) with \(\eta=0.0\). Our goal is to show that, as \(T'\) grows, the reward estimate \(r'(\tilde{\mathbf{z}}_{0 \mid \tilde{t}(0)})\) converges to \(r'(\mathbf{z}_0)\), thereby improving estimation accuracy.

Let \(\mathbf{z}_{t-1}\) be a state in the latent space from which we wish to recover the initial latent \(\mathbf{z}_0\). 
By applying \(T'\) steps of DDIM with \(\eta=0.0\), we obtain an approximation \(\tilde{\mathbf{z}}_{0 \mid \tilde{t}(0)}\). 
From prior work~\citep{li2024accelerating}, the error \(\|\tilde{\mathbf{z}}_{0 \mid \tilde{t}(0)} - \mathbf{z}_0\|\) scales as follows:  
\[
 \|\tilde{\mathbf{z}}_{0 \mid \tilde{t}(0)} - \mathbf{z}_0\| \le 
 \begin{cases}
   \mathcal{O}(1/T') & \text{(DDIM)}, \\
   \mathcal{O}(1/\sqrt{T'}) & \text{(DDPM)}, \\
   \mathcal{O}(1/(T')^n) & \text{(an \(n\)-th order solver)}.
 \end{cases}
\]
Hence, increasing \(T'\) yields a progressively better approximation of \(\mathbf{z}_0\).

Assume \(\mathbf{z}_0\) is the latent representation at time \(t=0\). 
By the Continuous Mapping Theorem, if \(\tilde{\mathbf{z}}_{0 \mid \tilde{t}(0)} \to \mathbf{z}_0\) as \(T' \to \infty\), then for any continuous function \(f\), we have 
\[
 f(\tilde{\mathbf{z}}_{0 \mid \tilde{t}(0)}) \;\to\; f(\mathbf{z}_0).
\]
Setting \(f(\cdot) = r'(\cdot)\), where \(r'\) is our reward model, yields
\[
 r'(\tilde{\mathbf{z}}_{0 \mid \tilde{t}(0)}) \;\to\; r'(\mathbf{z}_0),
\]
as \(T' \to \infty\).

We further assume that the reward model \(r'(\cdot)\) is Lipschitz continuous with Lipschitz constant \(L\). 
Then for any two latent states \(\mathbf{z}_a\) and \(\mathbf{z}_b\), the reward estimates satisfy
\[
 |r'(\mathbf{z}_a) - r'(\mathbf{z}_b)| \;\leq\; L \|\mathbf{z}_a - \mathbf{z}_b\|.
\]
Hence, the order of the error in \(r'(\tilde{\mathbf{z}}_{0 \mid \tilde{t}(0)})\) tracks the order of the error in \(\tilde{\mathbf{z}}_{0 \mid \tilde{t}(0)}\) itself. Explicitly,
\[
 \bigl|r'(\tilde{\mathbf{z}}_{0 \mid \tilde{t}(0)}) - r'(\mathbf{z}_0)\bigr| 
 \;\leq\; L \,\|\tilde{\mathbf{z}}_{0 \mid \tilde{t}(0)} - \mathbf{z}_0\|,
\]
implying that an \(\mathcal{O}(1/T')\) (or better) approximation in latent space implies an \(\mathcal{O}(1/T')\) (or correspondingly better) approximation in the reward space.

As \(T'\) increases, \(\tilde{\mathbf{z}}_{0 \mid \tilde{t}(0)}\) converges to \(\mathbf{z}_0\), and consequently \(r'(\tilde{\mathbf{z}}_{0 \mid \tilde{t}(0)})\) converges to \(r'(\mathbf{z}_0\). 
Because the reward model is Lipschitz continuous, this convergence ensures that the error in reward estimation decreases at the same order as the error of the latent approximation. 
Therefore, employing the LA estimator with a larger \(T'\) yields a more accurate reward estimate.

\clearpage
\section{List of Prompts}
\label{sec:prompts}

\scriptsize
\paragraph{MSRVTT-test}
\begin{enumerate}[leftmargin=0.5cm,topsep=0pt,itemsep=0.05pt]
 \item a woman is singing on stage about that one person being the one she wants
 \item someone is filming a parked car in the parking lot
 \item a cat is feed it s babies and a rabbit
 \item mario game with bombs
 \item someone is browsing a set of games on their console
 \item a game is being played
 \item a man holds a very large stick
 \item a yellow-haired girl is explaining about a game
 \item a ship is sailing around on the water
 \item a woman with blonde hair and a black shirt is talking
 \item a buffalo is attacking a man
 \item a band is playing music and people are dancing
 \item a child is playing a video game
 \item a person is showing how to fold paper
 \item a woman is sitting down on a couch in a room 
 \item a man inside of a car is using his finger to point
 \item a man waters his plants
 \item the symmetrical cone is japan s most famous symbol
 \item an indoor soccer game
 \item a japanese monkey bathing in a hot spring with pleasant music
 \item some images of motorcycles are being shown on tv
 \item someone is serving food in the restaurand
 \item this is a competition type show
 \item a woman on the news is talking about a story
 \item this is a phone review video
 \item some fake horses are standing around in a game
 \item a person is filming a white car interior seat
 \item video of clips from a movie
 \item a man with a blue and white shirt is walking around
 \item person making something in the kitchen
\end{enumerate}

\paragraph{DEVIL-high}
\begin{enumerate}[leftmargin=0.5cm,topsep=0pt,itemsep=0.05pt]
 \item  A bookshelf collapses loudly, books flying everywhere, creating chaos in the once quiet room.
 \item Swift scenes of a sandstorm engulfing a desert oasis, with dunes shifting and palm trees bending in the relentless wind.
 \item A chaotic scene of cowboys rounding up cattle during a stampede.
 \item Suddenly, a storm hits the city, rain pouring down like a torrent, making rivers on the streets.
 \item WWI biplanes in a dogfight with canvas wings ripping, dramatic cloud backdrop, ultra-detailed.
 \item In the mountains, a bear erupts from the snow, creating a large cloud of powder.
 \item Amidst a thunderstorm, a lightning bolt strikes a bicycle, setting it ablaze with crackling energy and lighting up the dark, rainy street.
 \item A single eagle dives extremely fast, snatching a fish from the water.
 \item A boat hits a big wave and flips, landing upside down.
 \item A car drives through a wall of fire in a daring escape.
 \item The cat tore across the living room, jumping over toys and furniture to catch the mouse.
 \item A cow jumps over a fence, landing in a pond with a big splash.
 \item Two dogs chase each other, suddenly skidding around a sharp corner.
 \item A storm sweeps an elephant into a raging river, carrying it away swiftly.
 \item Racing the sunset, a giraffe charges across the horizon, shadows stretching long.
 \item Against the wind, a lone horse gallops, mane streaming behind.
 \item Jumping over a gorge, the motorcycle lands just in time on the other side.
 \item A thief sprints away from the scene, with the police in hot pursuit.
 \item The ice cracks beneath their feet, making the sheep skid and slide, rushing to solid ground.
 \item Lightning strikes as a train blasts its horn, cutting through a stormy night.
 \item A truck speeds across the desert, dust clouds swirling behind it.
 \item Under a rainbow, a zebra kicks up a spray of water as it crosses a fast-flowing river.
\end{enumerate}

\paragraph{DEVIL-medium}
\begin{enumerate}[leftmargin=0.5cm,topsep=0pt,itemsep=0.05pt]
 \item  London heathrow, united kingdom - 05 12 2019: 4k super-telephoto plane accelerates down hot runway through heat shimmer
 \item A cool dj teddy bear with sunglasses on top of turntable with video static
 \item Aerial view.  cute girl in the coat drive on country road on the bicycle
 \item Brown pelican flying flight in fall bay harbor in ecuador
 \item Small fishing boat, anchored on a silver ocean, in thailand.
 \item a filled yellow school bus with over-sized black wheels drives through a flooded area with red lights on and gets splattered with mud
 \item St. petersburg, russia - circa march, 2015: vehicles drive on city ringroad at evening time. st. petersburg ring road is a main route encircling the city
 \item cat manages to hang on to dangling object
 \item Taking cow milk cheese with fork 4k footage
 \item dog passes in and out of view
 \item 1930s: elephant roars, man shoots at elephant. elephants walk through jungle. man tries to fire gun, throws gun on ground, runs away.
 \item the baby giraffe is zoomed in on and then camera shakes
 \item Cowboys drive group of horses at farming enterprise.
 \item 4k couple watching film or tv at home \& jumping with shock at the action
 \item contestants are reading themselves to start a mini-motorbike race
 \item Macao beach with stone mountains aerial view from drone. travel destination. summer vacation. dominican republic
 \item Male boxer resting and sweating after boxing training
 \item Wild tulips in a meadow on background sky. sunrise. bonfire. a quiet spring morning in the steppe.
 \item Sheep eating grass in punata and potosi, bolivia.
 \item Bodo arctic town norway - ca july 2018: train station building and rails tilt up
 \item a woman is describing different sets of tubes and hoses in the back of a white pick up truck which is parked on the side of a street with cars going by in the background
 \item Istanbul, turkey - october 2018: commuters inside istanbul metro wagon travelling towards taksim station
\end{enumerate}

\paragraph{DEVIL-static}
\begin{enumerate}[leftmargin=0.5cm,topsep=0pt,itemsep=0.05pt]
 \item airplane with red body is shown for first time.
 \item a man holds up a stuffed bear.
 \item when you can see the first view of the full bike
 \item second bird lands on feedersecond bird lands on feeder
 \item a red boat is first seen.
 \item Tourist bus station 3d realistic footage. public transport front view animation. vehicles on modern urban highway bridge background. passengers transportation parking. city bus stop video
 \item black car is under the blue sign.
 \item cat looks at the camera
 \item dog puts paws together
 \item a white horse standing beside red colored wearing girl dress standing with stick bending down knee  displaying on screen
 \item Blurred conference room with audience - 4k video
 \item first time we see orange branch to the right
 \item A woman and a man. holding a gift.
 \item A tranquil tableau of the old red barn stood weathered and iconic against the backdrop of the countryside
 \item black numbers 1758 at bottomof train
 \item a large white box truck travels through water is followed by two other trucks and ascends a gray road through mountains
 \item view of big city from balconyview of big city from balcony
\end{enumerate}

\clearpage
\paragraph{DEVIL-very-high}
\begin{enumerate}[leftmargin=0.5cm,topsep=0pt,itemsep=0.05pt]
 \item Classical style of a horse partaking in an ancient chariot race, scenes switching quickly from cheering crowds to close-ups of intense wheel clashes.
\item High-speed shots of a volcanic eruption engulfing a tropical island, with lava fountains spewing molten rock and the environment transforming from idyllic paradise to hellish landscape of ash and fire.
\item A thrilling scene of rural mountain biking extreme sports, starting from the early morning cycling adventure, transitioning to the intense chase through fields and forests, and ending with cheers and celebrations under the sunset.
\item Neon-lit streets pulse with energy as vehicles engage in a high-octane pursuit, transitioning seamlessly from chaos to calculated evasion. Against the backdrop of a setting sun, the chase intensifies, each turn a heartbeat away from capture.
\item A fighter jet dodging rapid anti-air gunfire, quick maneuvers, tracer rounds visible.
\item Bear escaping a collapsing cave, rocks tumbling, dust rising, ((masterpiece)), ((best quality)), 8k, high detailed, ultra-detailed, bear, ((dark rocky textures)), sprinting, (echoing rumble), sudden movement
\item A courier on a bike weaves through traffic at breakneck speed, narrowly avoiding cars and pedestrians in a rush to make deliveries on time.
\item A hummingbird rapidly darting between vibrant flowers in a lush garden, with quick cuts to various close-up shots showcasing its rapid wing movement and agility.
\item Venetian gondola chase scene, narrow canals, historic buildings, urgent escape, ((masterpiece)), ((best quality)), 8k, high detailed, ultra-detailed, gondola, ((twisting canals)), (ancient architecture), (urgent paddling), cinematic chase.
\item Futuristic sports cars racing on a vertical loop track against a sci-fi cityscape, cars defying gravity, ((speed trails)), (dizzying heights), (spectacular crashes), the thrill of cutting-edge technology.
\item Cat rapidly zigzagging across a rooftop, avoiding swooping birds under a stormy sky.
\item A sequence of a cow performing acrobatic stunts over a series of colorful, abstract platforms that morph shapes.
\item The dog bursts through a thicket, darting from a foggy forest to a steep hillside, rocks crumbling under its paws as it charges towards a roaring river below.
\item Thundering across a vast desert plain, the elephants race over dunes and dodge sandstorms, before swiftly traversing through a rocky canyon, bounding over boulders and leaping across narrow ravines.
\item A giraffe navigating a city during a robot uprising, with quick cuts showing chaotic battles, explosions, and futuristic technology in a high-stakes escape scenario.
\item A horse leading a wild stampede across a stormy beach with waves crashing, depicted with swift, sweeping camera moves, cinematic composition.
\item Intense motorcycle escape from a volcanic eruption, with transitions from lava-filled landscapes to ash-clouded skies.
\item A futuristic robot uprising, ((lasers firing)), metallic drones, explosions, debris, ((screaming civilians)), dystopian cityscape.
\item Sheeps engaging in a high-speed pursuit through a cyberpunk city, the scene rapidly transitioning between neon-lit streets, bustling marketplaces, and towering skyscrapers.
\item A pulse-pounding sequence of a train barreling through a treacherous storm, the scene transitioning between lightning-lit skies and torrents of rain to flooded tracks and collapsing bridges.
\item A truck rushing away from a treacherous mountain pass during a blizzard, with sudden avalanches and rockslides adding to the danger.
\item A zebra sprinting across the busy lanes of Times Square in New York City, with scene transitions occurring quickly as it moves from iconic billboards to bustling sidewalks filled with tourists.
\end{enumerate}

\paragraph{MovieGen}
\begin{enumerate}[leftmargin=0.5cm,topsep=0pt,itemsep=0.05pt]
 \item A green monster made of plants walks through an airport.
\item A marble goes through a glass cup, breaking it into pieces.
\item A droplet of water falling onto a hot surface, instantly evaporating into a wisp of steam that swirls gracefully into the air.
\item An old man wearing a green dress and a sun hat taking a pleasant stroll in Johannesburg South Africa during a beautiful sunset.
\item A person on a hoverboard colliding with a wall, the board stopping abruptly.
\item A toy robot wearing blue jeans and a white t-shirt taking a pleasant stroll in Johannesburg South Africa during a winter storm.
\item In a marathon race, a female athlete gradually sprints ahead of the male athletes.
\item A teenager eating a slice of pizza, cheese stretching as they pull it away.
\item A man in a suit fights monsters.
\item A dog made of ice melts completely in a hot summer day.
\item A truck right alongside a flowing river, capturing the movement of the water and the surrounding forest.
\item A group of skateboarders perform tricks on ramps and rails at a skate park, showcasing their skills.
\item A hot air balloon descending back to the ground.
\item Chef chopping onions in the kitchen for the preparation of the dish.
\item Zoom in shot to the face of a young woman sitting on a bench in the middle of an empty school gym.
\item The couple runs hand in hand to release a sky lantern, then watches it drift upward into the night sky, carried by the wind with the stars shining above.
\item Aerial view shot of a cloaked figure elevating in the sky between skyscrapers.
\item A softball player sliding safely into second base.
\item A giraffe in a lifeguard outfit, sitting atop a high chair and watching over a crowded pool.
\item A speed skater accelerating during a short track race.
\end{enumerate}
\normalsize

\clearpage
\section{Detailed Analysis on Reward Function for Perceptual Video Quality}
\label{sec:detailed_analysis_reward_metrics}
\begin{figure}[ht]
  \centering
  \includegraphics[width=\linewidth]{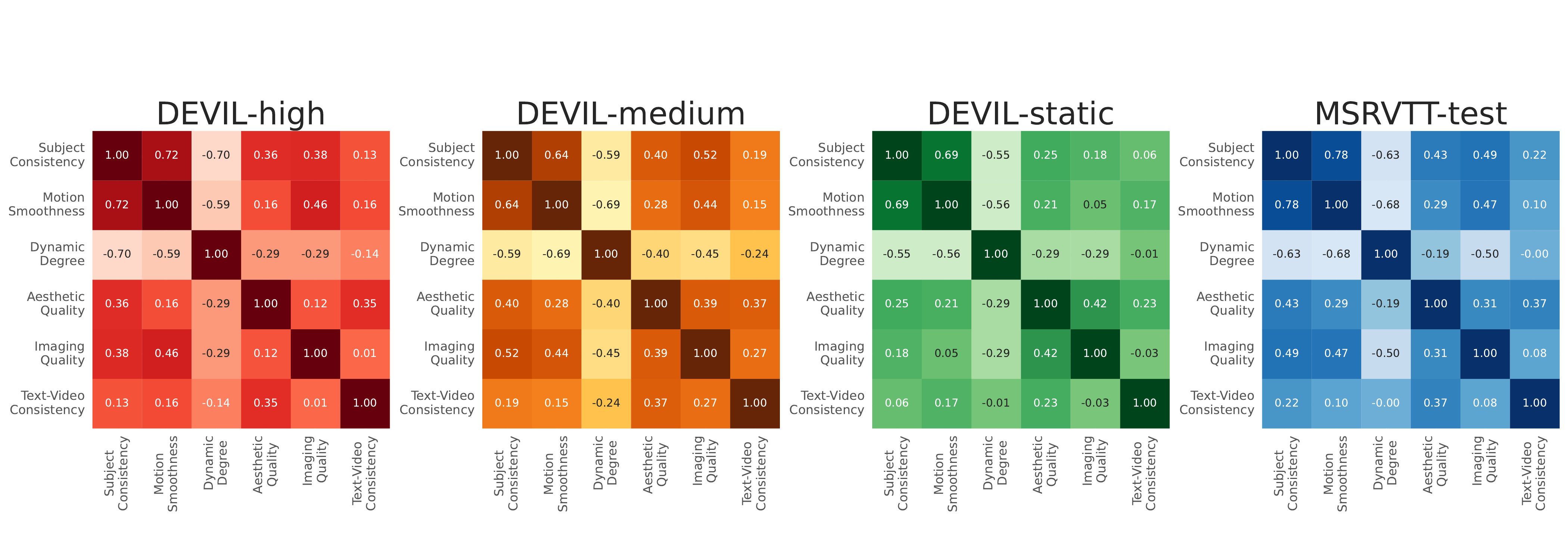}
  % \vskip -0.15in
  \caption{Correlation between reward functions for perceptual video quality.}
  \label{fig:correlation}
\end{figure}

\autoref{fig:correlation} shows that different metrics in reward functions for perceptual video quality often exhibit negative or weak correlations. 
For example, dynamic degree tends to be negatively correlated with many other metrics, indicating that optimizing exclusively for one metric can either reduce motion dynamics or undermine temporal consistency and aesthetic quality. 
These findings underscore the need to balance potentially conflicting reward functions, rather than prioritizing any single one in isolation, and emphasize the importance of a carefully calibrated approach to evaluating generated videos.

% \clearpage
\section{Prompt of AI Feedback}
\label{sec:aif_prompt}
\begin{tcolorbox}[title=Prompt for AI Feedback from VLMs]
\small
You are a helpful assistant that evaluates the quality of a generated video from a textual prompt. \\

\vspace{1.5mm}
Compare the text prompt and generated video and evaluate the quality (visual quality, proper dynamics, etc...) of the video. \\

First explain the reasoning, then present the final assessment. Start the reasoning with 'Reasoning: '.\\

After explaining the reasoning, present the final assessment with 'Assessment: '. \\

Your final 'Assessment' should be a single-number score from 1 to 10, not as a fraction. \\

When evaluating, consider the following points: \\

\vspace{1.5mm}
\qquad - Visual Quality: Evaluate the clearness, resolution, brightness, aesthetic appeal of the video. \\

\qquad - Dynamics: Evaluate whether the video demonstrates appropriate dynamics, ensuring it avoids excessive movement in situations meant to be static or insufficient movement in situations intended to be dynamic. \\

\qquad - Smoothness, Consistency, and Naturalness: Assess the smoothness, consistency, and naturalness of shape and motion for objects, animals, and humans.\\

\qquad - Contents: Evaluate whether the video content aligns with the given text prompt. \\

\vspace{1mm}
Textual Prompt: \{instruction\} \\

\vspace{1mm}
Video: \{video\_file\}\\

\end{tcolorbox}

\clearpage
\section{Further Results for Calibrating Reward to Preference Feedback} % from GPT-4o}

\subsection{Basic Prompts}
\label{sec:reward_gpt_4}

\begin{figure}[ht]
% \begin{wrapfigure}{r}{0.5\linewidth}
  \centering
  \includegraphics[width=\linewidth]{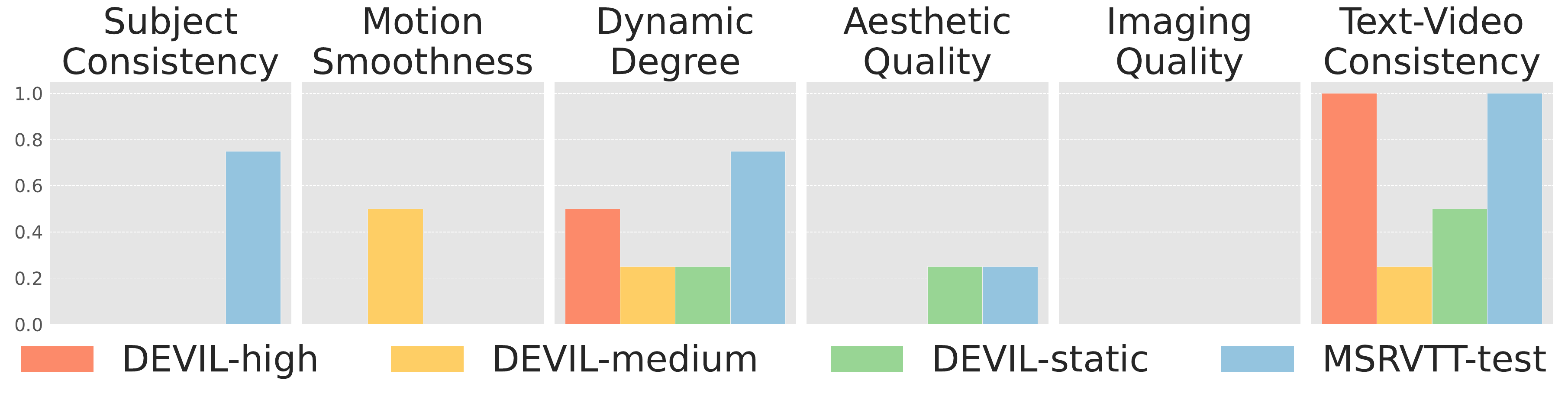}
  \caption{Coefficients of calibrated reward with GPT-4o.}
  \label{fig:weight_selection_gpt}
% \end{wrapfigure}
\end{figure}

\begin{figure*}[ht]
  \centering
  \includegraphics[width=\linewidth]{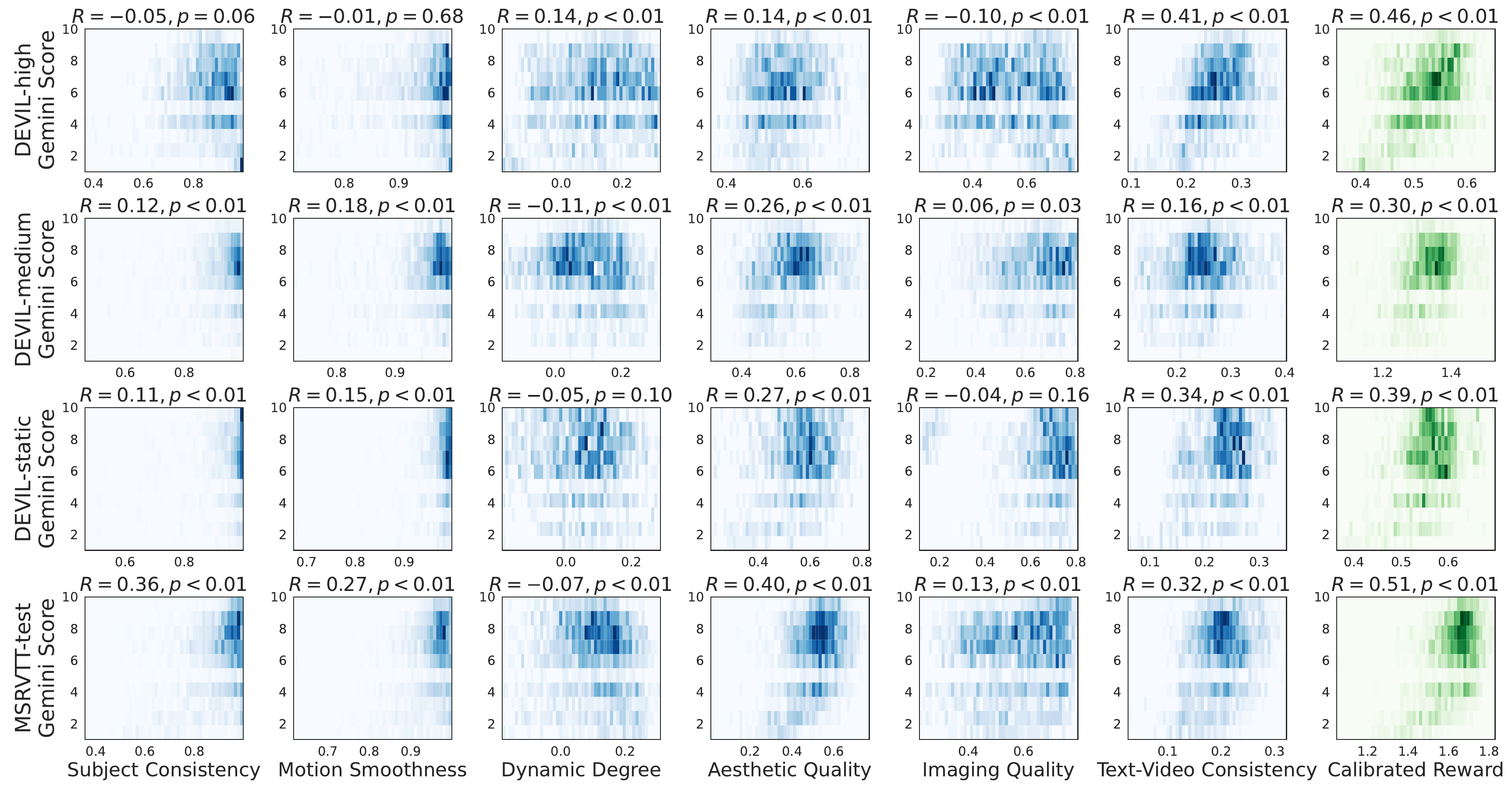}
  \caption{%
  2D-histogram and correlation between reward function and AI feedback from Gemini.
  % 2D-histogram and correlation between reward functions for perceptual video quality~\citep{huang2023vbench} and AI feedback from Gemini~\citep{geminiteam2023gemini}.
  % % We generate 64 videos per prompt from pre-trained Latte~\citep{ma2024latte}.
  % A single reward (e.g., subject consistency; \textcolor{cbblue}{blue}) is often not aligned well with a preference from Gemini, which happens for all the prompt sets with different degree of dynamics.
  % The calibrated reward, a linear combination of perceptual metrics via brute-force search (\textcolor{cbgreen}{green}), achieves the best Pearson correlation coefficient in all settings (statistically significant with $p<0.01$).
  }
  \label{fig:hist_gemini_full}
  % \vskip -0.2in
\end{figure*}

\begin{figure}[ht]
  \centering
  \includegraphics[width=\linewidth]{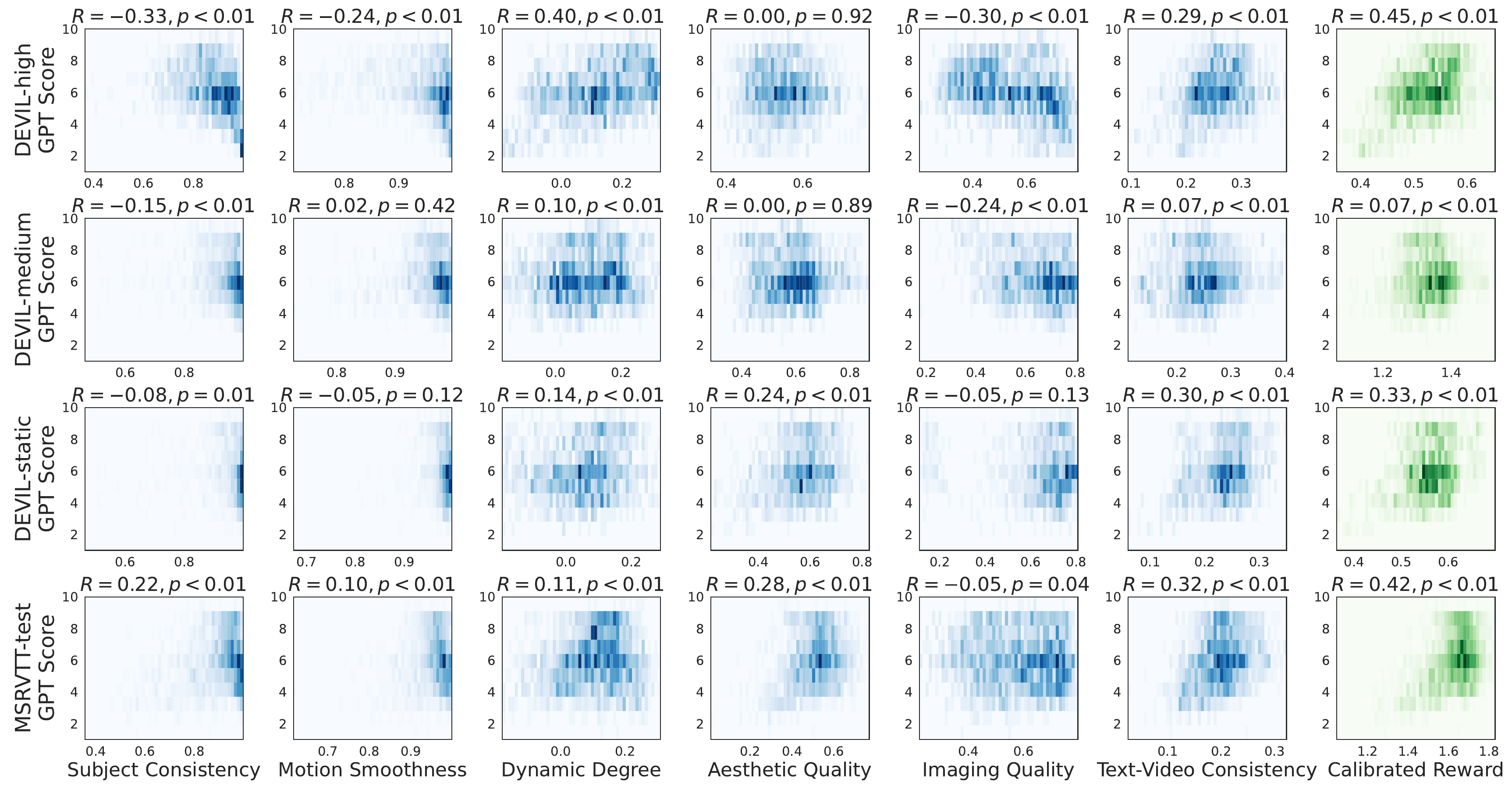}
  \caption{2D-histogram and correlation between reward function and AI feedback from GPT-4o.}
  \label{fig:hist_gpt}
\end{figure}

\autoref{fig:hist_gemini_full} and \autoref{fig:hist_gpt} show the two-dimensional histogram and correlation between reward function and AI feedback from Gemini~\citep{geminiteam2023gemini} and GPT-4o~\cite{openai2023gpt4}, and 
\autoref{fig:weight_selection_gpt} represents the coefficient of calibrated reward designed for GPT-4o.
% The relative weighting of the dynamic degree shifts in accordance with the prompt-specified dynamics, reflecting a pattern similar to that observed in the reward calibration approach with Gemini (DEVIL-high, DEVIL-medium, and DEVIL-static denote prompt sets characterized by high, medium, and low levels of dynamics, respectively). 
The relative weighting assigned to the dynamic degree changes according to the dynamics grade of the prompt. 
Specifically, prompts with a high dynamics grade, i.e., DEVIL-high, place greater weight on the dynamic degree. 
In contrast, prompts that describe slight motion, i.e., DEVIL-medium and DEVIL-static, place a smaller weight on it. 
This behavior matches the pattern observed in reward calibration with Gemini (\autoref{fig:weight_selection_gemini}).
GPT-4o exhibits a stronger inclination toward dynamics than Gemini. 

\subsection{Challenging Prompts}
\label{sec:reward_challenging}

\begin{figure}[ht]
  \centering
  \includegraphics[width=\linewidth]{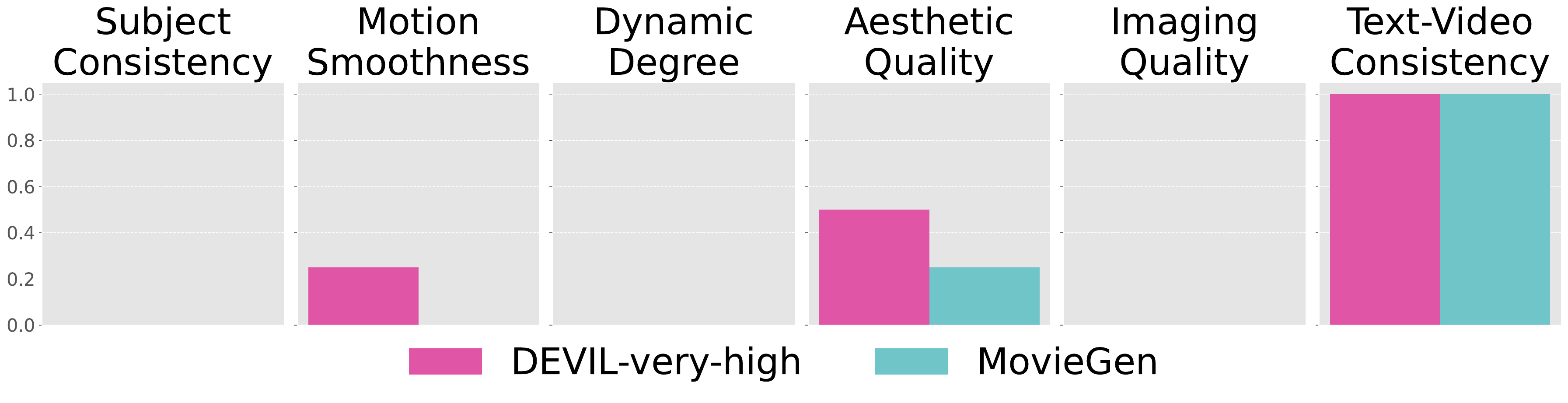}
  \caption{Coefficients of calibrated reward with Gemini.}
  \label{fig:weight_selection_challenging}
\end{figure}

\begin{figure}[ht]
  \centering
  \includegraphics[width=\linewidth]{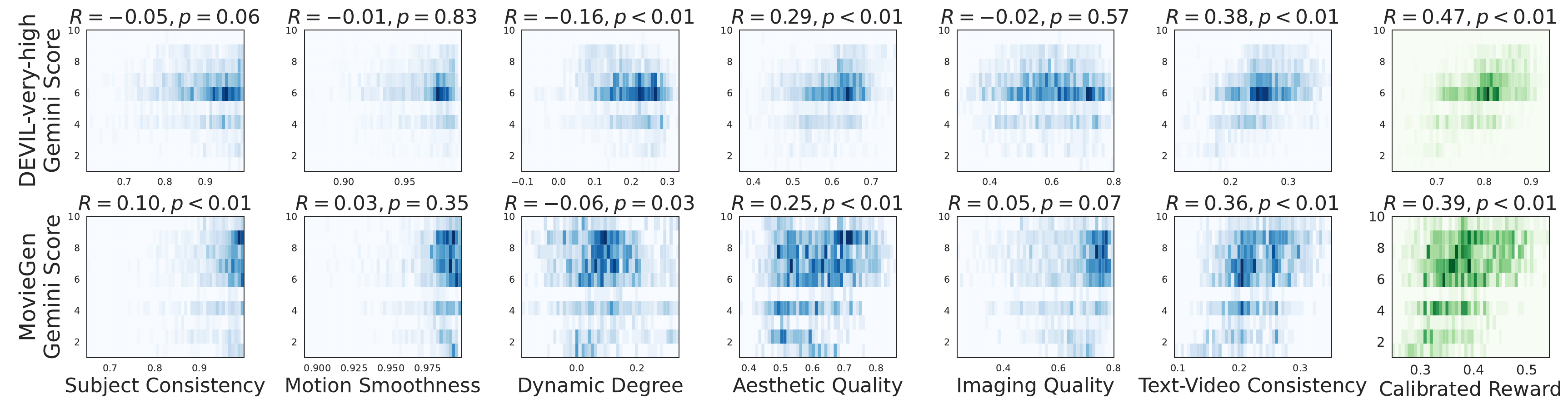}
  \caption{2D-histogram and correlation between reward function and AI feedback from Gemini for challenging prompt sets, DEVIL-very-high and MovieGen.}
  \label{fig:hist_challenging}
\end{figure}

This section describes the reward calibration procedure and results for two challenging prompt sets, DEVIL-very-high and MovieGen, which were introduced to evaluate our method with larger T2V models, such as CogVideoX~\citep{yang2024cogvideox} and Wan 2.1~\citep{wan2025}. 
Following the methodology for reward calibration with Latte (see Section~\ref{sec:reward_calibration}), we generated 64 videos per prompt using Wan 2.1-1.3B~\citep{wan2025}. 
Consistent with observations in Appendix~\ref{sec:reward_gpt_4}, using solely text-video consistency is insufficient to fully capture AI feedback from Gemini (\autoref{fig:hist_challenging}). 
We choose the combination of weights $w_i$ to maximize correlation with Gemini’s evaluations. 
The coefficients of calibrated weights are shown in \autoref{fig:weight_selection_challenging}.

\clearpage
\subsection{Cost of Reward Calibration}
\label{sec:reward_calibration_cost}

As described in Section~\ref{sec:summary_of_video_metric} and Appendix~\ref{sec:reward_challenging}, we generate 64 videos per prompt using pre-trained Latte and Wan 2.1 models. 
Compared to naively querying VLMs at every inference step, our calibration approach is substantially more cost-efficient, since the VLM queries are amortized through a one-time weight estimation. 
\autoref{tab:reward_calibration_cost} summarizes the difference in per-prompt query count and execution time when applying DLBS ($K\!B=32$).
These results demonstrate that reward calibration reduces the number of VLM queries, making large-scale search with DLBS computationally feasible.

\begin{table*}[ht]
  \centering
  \caption{Comparison of query count and execution time between naive VLM queries during search and reward calibration.
  Assuming 15 seconds per VLM query.}
  \label{tab:reward_calibration_cost}
  \begin{tabular}{lcc}
    \toprule
    Method & Query Count & Exec. Time (sec) \\
    \midrule
    Querying VLMs during Search ($K\!B=32$) & $T{=}50 \times K\!B{=}32 = 1600$ & $\approx 102{,}400$ \\
    Reward Calibration & $\bf{64}$ & $\bf{\approx 960}$ \\
    \bottomrule
  \end{tabular}
\end{table*}

\subsection{Generalization of Reward Calibration across prompts}
\label{sec:generalization}

% \textbf{Generalization across prompts}~~
Video generation inherently involves trade-offs between fundamental properties such as dynamics and consistency (\autoref{sec:detailed_analysis_reward_metrics}), which may require category-specific calibration for optimal performance. 
However, despite these domain-specific requirements, we hypothesize that calibrated rewards can generalize to some extent across different datasets, as they are based on shared principles of perceptual quality. 
To test the out-of-domain transferability, we conducted additional experiments applying the reward weights calibrated on DEVIL-high and DEVIL-medium to MSRVTT-test prompts (\autoref{tab:generalization_prompts}). 
We used Latte~\citep{ma2024latte} as a base model and evaluated the results using VideoScore~\citep{he2024videoscore}, a human preference-trained evaluator, measuring five key metrics along with their corresponding average scores.
With the DEVIL-high reward, we can enhance other metrics while maintaining dynamics. 
DEVIL-medium reward, which is a closer domain to MSRVTT-test, shows a different trade-off pattern. 
While it slightly reduces dynamics, it significantly improves other metrics and achieves a higher average score than the MSRVTT-test reward, demonstrating higher transferability.
% Moreover, as confirmed in Section~\ref{sec:ablation_study}, generalization to unseen in-domain prompts using the same reward also improves performance (\autoref{fig:scaling_ablation}; \textbf{Right}).

\begin{table*}[ht]
  \centering
  \caption{Out-of-domain prompt generalization. 
  Rewards calibrated on DEVIL-high/medium applied to MSRVTT-test prompts. 
  All metrics are derived from VideoScore~\citep{he2024videoscore}. VQ = Visual Quality; TC = Temporal Consistency; DD = Dynamic Degree; T2V Align. = Text-to-video Alignment; FC = Factual Consistency.}
  \label{tab:generalization_prompts}
  \begin{tabular}{lcccccc}
    \toprule
    (R: Reward, P: Prompt) & VQ & TC & DD & T2V Align. & FC & Average \\
    \midrule
    Latte & 2.32 & 2.01 & \textbf{2.91} & 2.67 & 2.07 & 2.40 \\
    \midrule
    + DLBS ($K\!B=8$) & & & & & & \\
    with R=MSRVTT, P=MSRVTT & \textbf{2.50} & \textbf{2.27} & 2.88 & \textbf{2.74} & 2.28 & 2.53 \\
    with R=DEVIL-medium, P=MSRVTT & 2.49 & 2.26 & 2.89 & 2.73 & \textbf{2.31} & \textbf{2.54} \\
    with R=DEVIL-high, P=MSRVTT & 2.36 & 2.03 & \underline{2.90} & 2.68 & 2.10 & 2.42 \\
    \bottomrule
  \end{tabular}
\end{table*}

\clearpage
\section{Correlation between VLM and Human Evaluation}
\label{sec:corr_vlm_human_eval}

\begin{figure}[ht]
  \centering
  % \vskip -0.05in
  \includegraphics[width=0.60\linewidth]{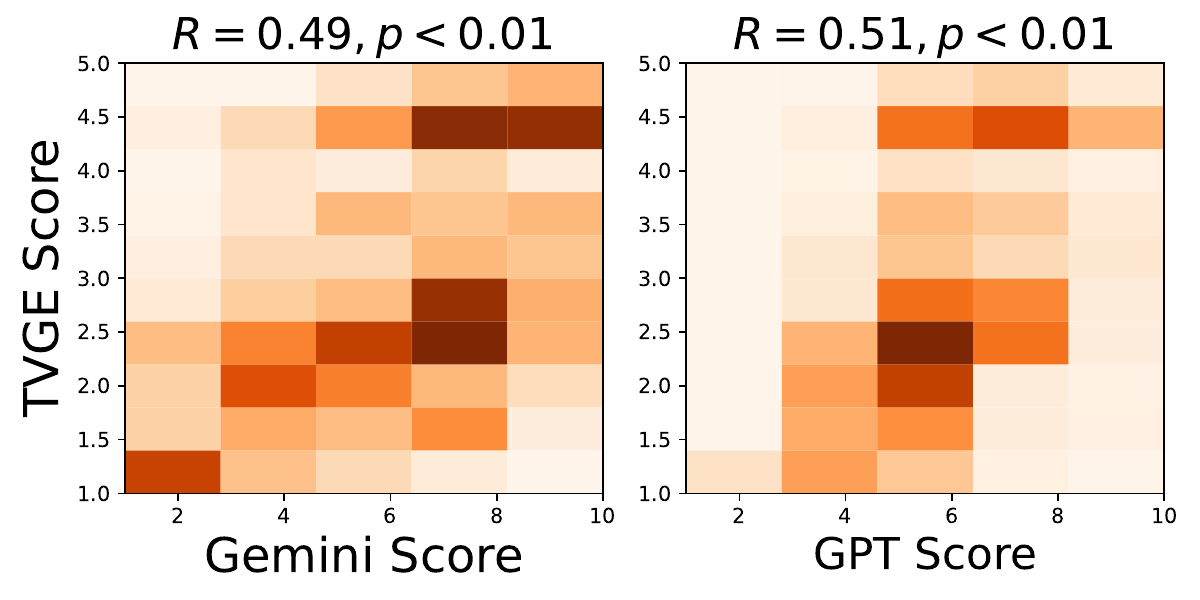}
  % \vskip -0.05in
  \caption{
  Correlation between VLM outputs and human labels in the TVGE dataset.
  }
  % \vskip -0.1in
  \label{fig:corr_hf_aif}
\end{figure}

As mentioned in prior research~\citep{na2024boost, wu2024boosting, furuta2024improving}, evaluation from VLMs such as Gemini and GPT-4o exhibits a high correlation with human assessment compared to other existing metrics. 
As an experiment, we measured the correlation between the AI feedback from these VLMs and human labels in the TVGE dataset~\citep{wu2024better}. 
As shown in~\autoref{fig:corr_hf_aif}, Gemini achieved a correlation of 0.49, and GPT-4o achieved 0.51. 
Consequently, optimizing for these VLM rewards is a valid way to improve human-perceived quality, rather than merely ``gaming'' the metrics.

\begin{figure}[ht]
  \centering
  % \vskip -0.05in
  \includegraphics[width=\linewidth]{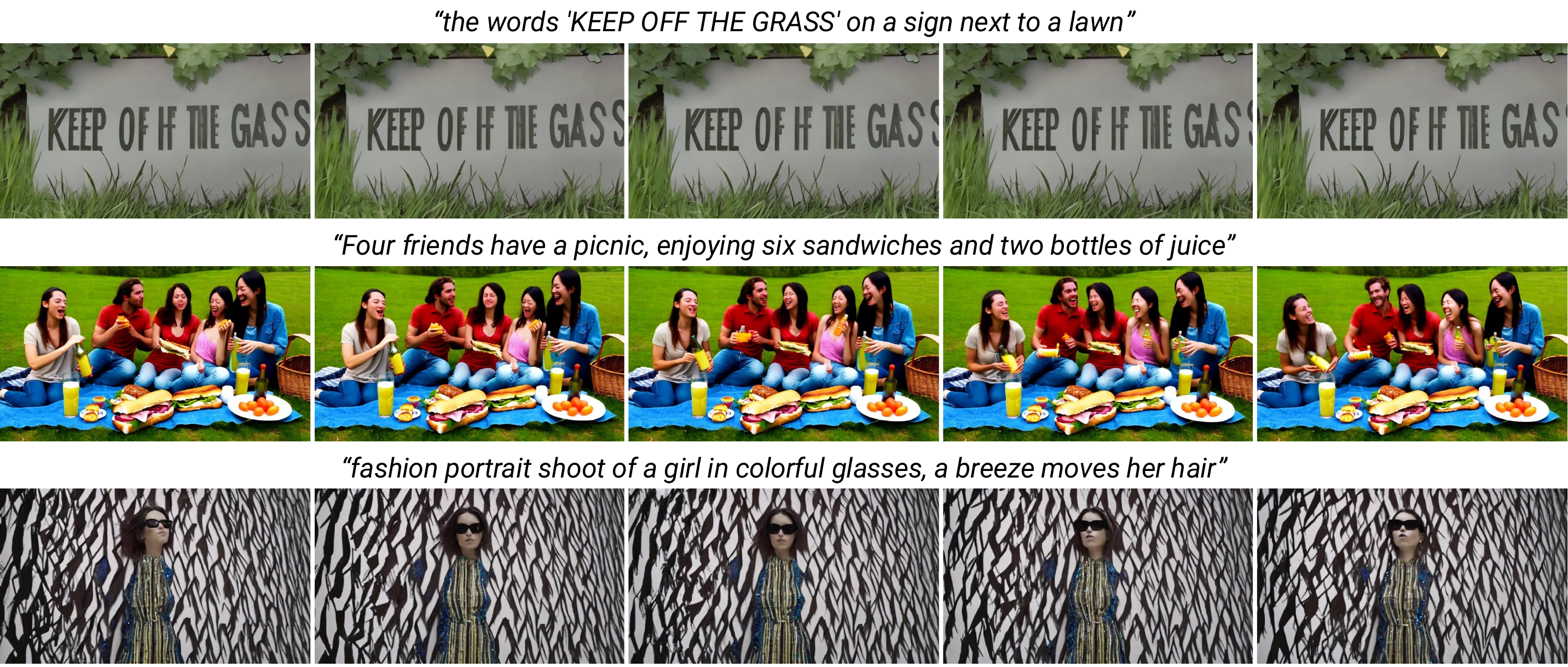}
  % \vskip -0.05in
  \caption{
  Misaligned cases of Gemini-based evaluation with human preferences. 
  (\textbf{Top}) For prompts specifying text, such as \textit{"the words 'KEEP OFF THE GRASS' on a sign next to a lawn,"} the VLM was significantly harsher than human evaluation on text rendering (VLM: 2/10, Human: 4.5/5).
  (\textbf{Middle}) For prompts specifying quantities, such as \textit{"Four friends have a picnic, enjoying six sandwiches and two bottles of juice,"} the VLM was more lenient than human evaluation (VLM: 8/10, Human: 1.5/5).
  (\textbf{Bottom}) For \textit{"fashion portrait shoot of a girl in colorful glasses, a breeze moves her hair,"} despite missing arms in the generated person, the VLM was misled by distracting background patterns, possibly mistaking them for curtain-like elements that obscure the arms behind the background (VLM: 8/10, Human: 1.2/5).
  }
  % \vskip -0.1in
  \label{fig:tvge_vlm_misalign}
\end{figure}

For a deeper analysis of failure cases, we qualitatively examined the top 5\% outliers between human preference labels in the TVGE dataset and VLM (Gemini) evaluation (\autoref{fig:tvge_vlm_misalign}). 
As far as we observed, VLM sometimes makes subtle mistakes, but we did not see any critical failures.

\clearpage
\section{Qualitative Evaluation of Calibrated Reward}
\label{sec:reward_calibration_qualitative}
\begin{figure}[ht]
  \centering
  \vskip -0.05in
  \includegraphics[width=\linewidth]{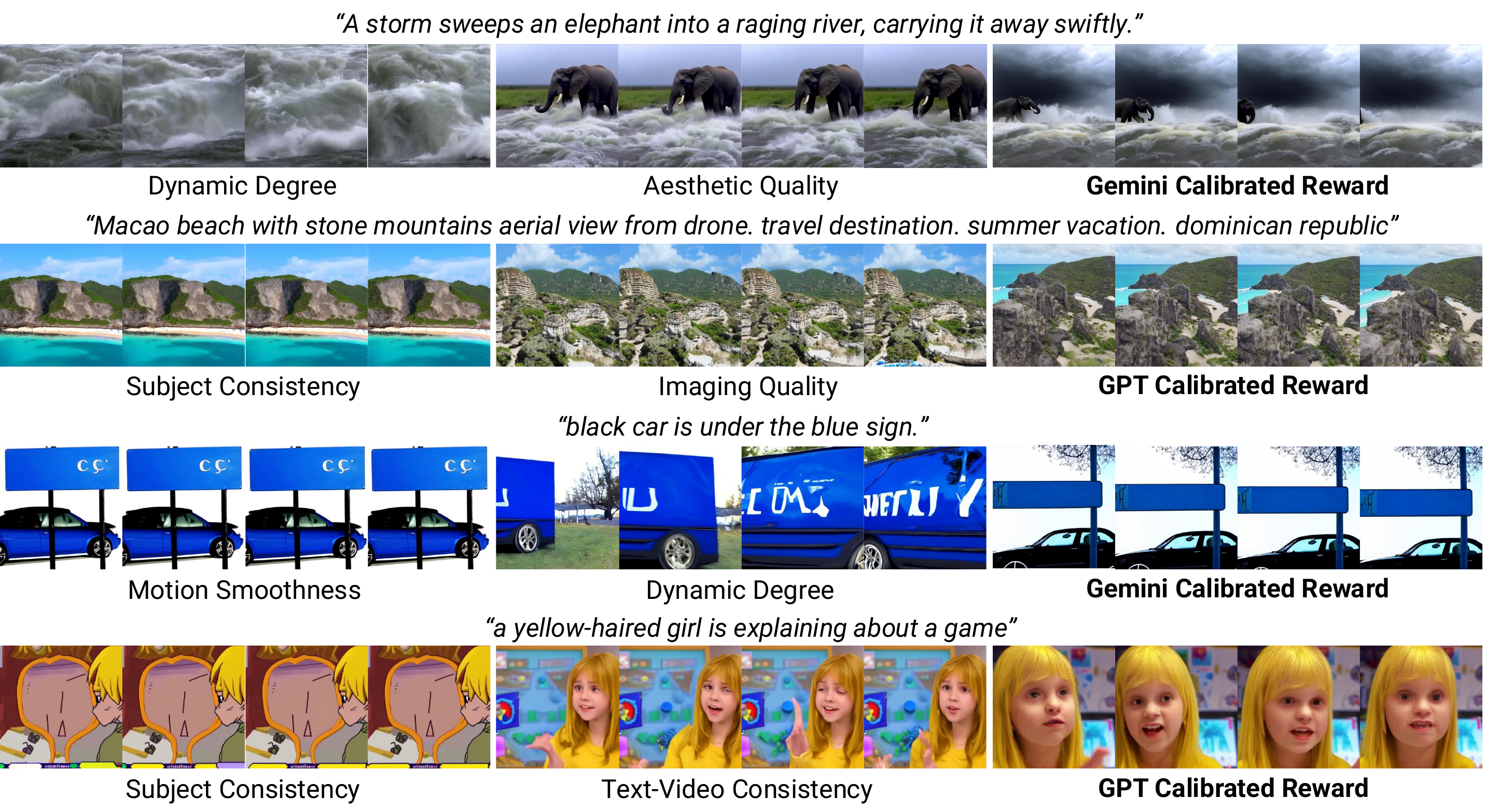}
  % \vskip -0.05in
  \caption{
  We select the video with the highest reward out of 64 randomly generated candidates for each prompt, drawn from DEVIL-high, DEVIL-medium, DEVIL-static, and MSRVTT-test (arranged from top to bottom). 
  Videos chosen using VLM-calibrated rewards achieve a more balanced quality compared to those relying on any single metric. 
  For instance, when subject consistency, motion smoothness, or aesthetic quality serves as the sole selection criterion, the resulting videos often lack dynamic movement, whereas prioritizing dynamic degree can compromise temporal consistency. 
  Moreover, single-metric evaluations may occasionally fail to align with the intended prompt.
  % Prompts: 'Under a rainbow, a zebra kicks up a spray of water as it crosses a fast-flowing river.'(DEVIL-high, above), 'dog puts paws together'(DEVIL-static, bottom).
  }
  % \vskip -0.175in
  \label{fig:reward_qualitative}
\end{figure}

We provide best-of-64 videos by individual rewards and VLM calibrated rewards in \autoref{fig:reward_qualitative}. 
Videos selected solely on a single metric can over-optimize one aspect while neglecting others, whereas those chosen via VLM-calibrated rewards exhibit a more balanced quality.
For instance, videos chosen solely based on temporal consistency (i.e., subject consistency and motion smoothness) or frame-by-frame quality (i.e., aesthetic quality, imaging quality) tend to lack dynamic movement, whereas those selected based on dynamic degree often lose temporal consistency. 
Evaluations relying on a single metric also fail to reflect the given prompt in some cases.
Text-video consistency, which often exhibits a high correlation with VLM-based evaluation among individual metrics (\autoref{fig:hist_gemini}), is relatively effective in capturing the overall quality of a video. 
However, it may overlook certain aspects, such as frame-wise artifacts.
In contrast, videos selected using VLM-calibrated rewards exhibit a more balanced overall quality.

\clearpage
\section{Further Results for Diffusion Latent Beam Search}

\subsection{Scaling Search Budget with GPT-4o Calibrated Reward}
\label{sec:gpt_scaling}

\begin{figure}[ht]
  \centering
  \includegraphics[width=\linewidth]{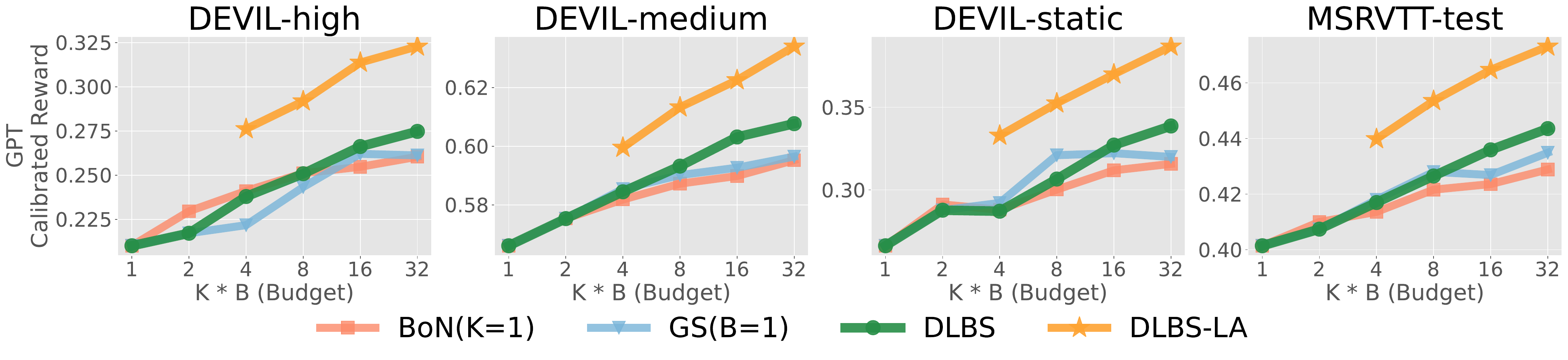}
  % \vskip -0.1in
  \caption{Inference-time search on reward calibrated to GPT-4o.}
  \label{fig:gpt_scaling}
\end{figure}

We measure the performance using a reward calibrated to GPT-4o (\autoref{fig:gpt_scaling}).
DLBS improves all the calibrated rewards the best as the search budget $KB$ increases (especially $KB=16,32$), while BoN and GS, in some cases, eventually slow down or saturate the performance.
Notably, an LA estimator with a small search budget ($KB=8, T'=6$) is comparable to or even outperforms DLBS ($KB=32$).

\subsection{Scaling Search Budget to Larger Regimes}
\label{sec:larger_scaling}
\begin{figure}[ht]
  \centering
  \includegraphics[width=\linewidth]{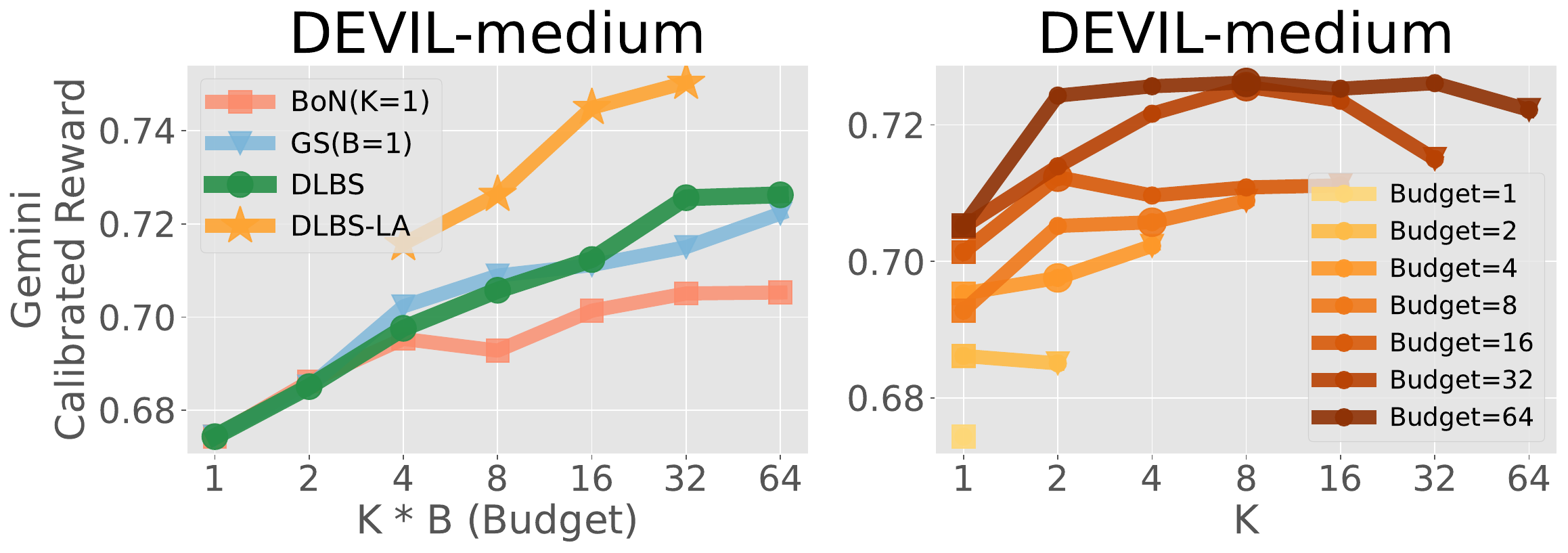}
  % \vskip -0.1in
  \caption{Inference-time search on reward calibrated to Gemini including $KB=64$.}
  \label{fig:large}
\end{figure}

\begin{figure}[ht]
  \centering
  \includegraphics[width=\linewidth]{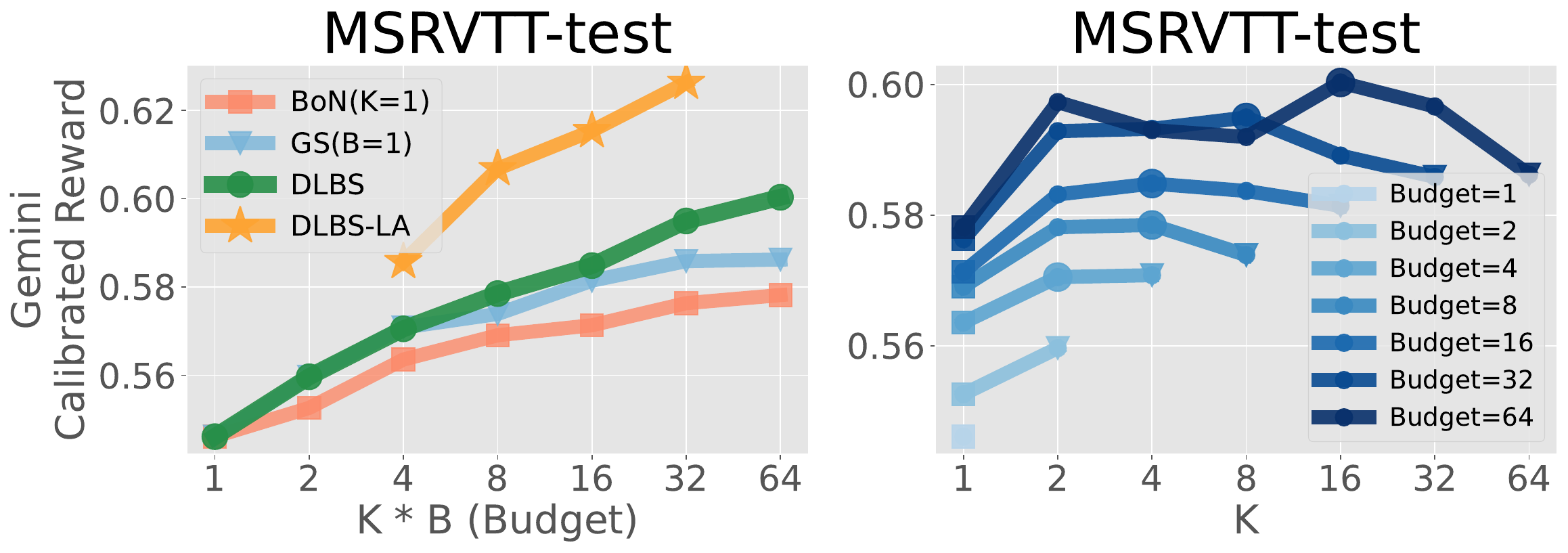}
  % \vskip -0.1in
  \caption{Inference-time search on reward calibrated to Gemini including $KB=64$.}
  \label{fig:large_msrvtt}
\end{figure}

\autoref{fig:large} and \autoref{fig:large_msrvtt} show the performance of inference-time search on DEVIL-medium and MSRVTT-test that includes the results with $KB=64$. We can observe that the increasing trends still continue.

\clearpage
\subsection{Full Results for Scaling Trend of DLBS}
\label{sec:full_optimal_balance}

\begin{figure*}[ht]
  \centering
  \includegraphics[width=\linewidth]{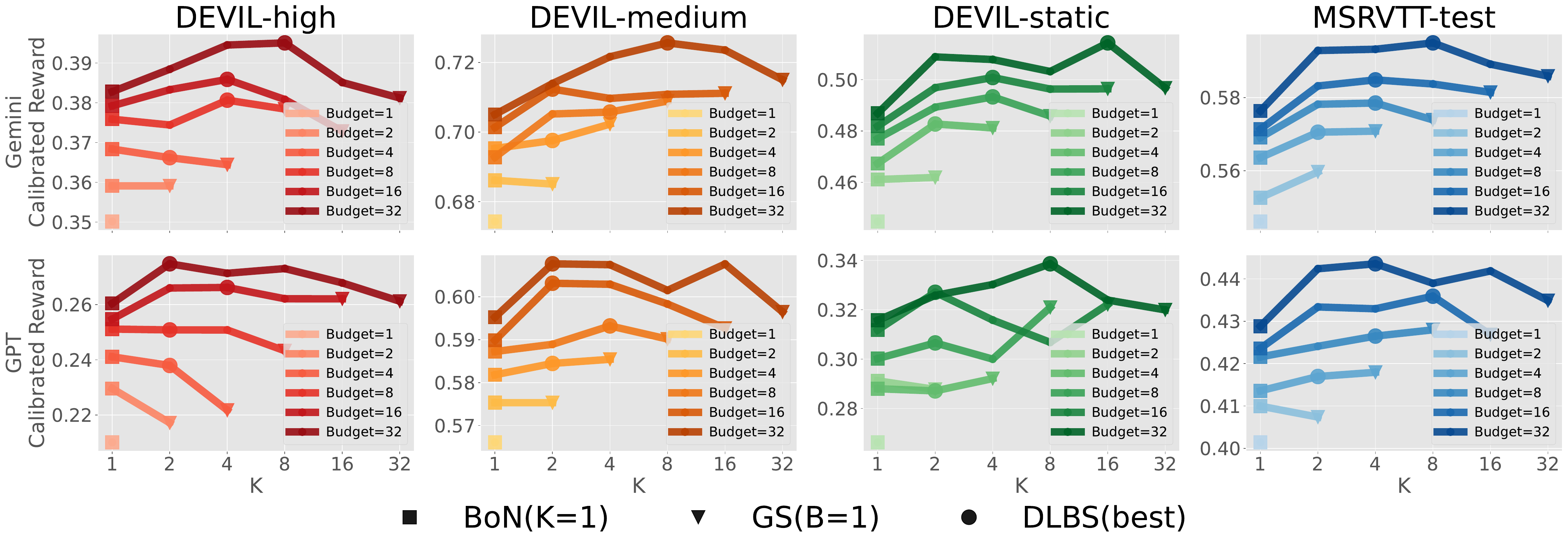}
  % \vskip -0.1in
  \caption{
    DLBS can improve the performance in any prompts or reward, as we increase the search budget $KB \in \{1,2,4,8,16,32\}$.
    In addition, we can see an optimal balance between the number of latent $K$ and the number of beam $B$ under the same budget.
    For instance, as we increase the budget to $KB=16, 32$, we have a peak around $K=4, 8, 16$, which is about 25--50\% of the budget.
  }
  \label{fig:budget_full}
  % \vskip -0.175in
\end{figure*}

\autoref{fig:budget_full} demonstrates the scaling trend of DLBS, proportional to the search budget, under various choices of $K$.
The results show that there is an optimal balance between the number of latent $K$ and the number of beams $B$ under the same budget.
For instance, as we increase the budget to $KB=16,32$, we have a peak around $K=4,8,16$, which is about 25--50\% of the budget.
This implies that balancing possession and exploration of diffusion latents in DLBS helps search for the best outputs robustly.

\subsection{Further Analysis on Lookahead Estimator}
\label{sec:lookahead_ablation}

\begin{figure}[ht]
\centering
% \vskip -0.1in
\subfigure{\includegraphics[width=0.49\textwidth]{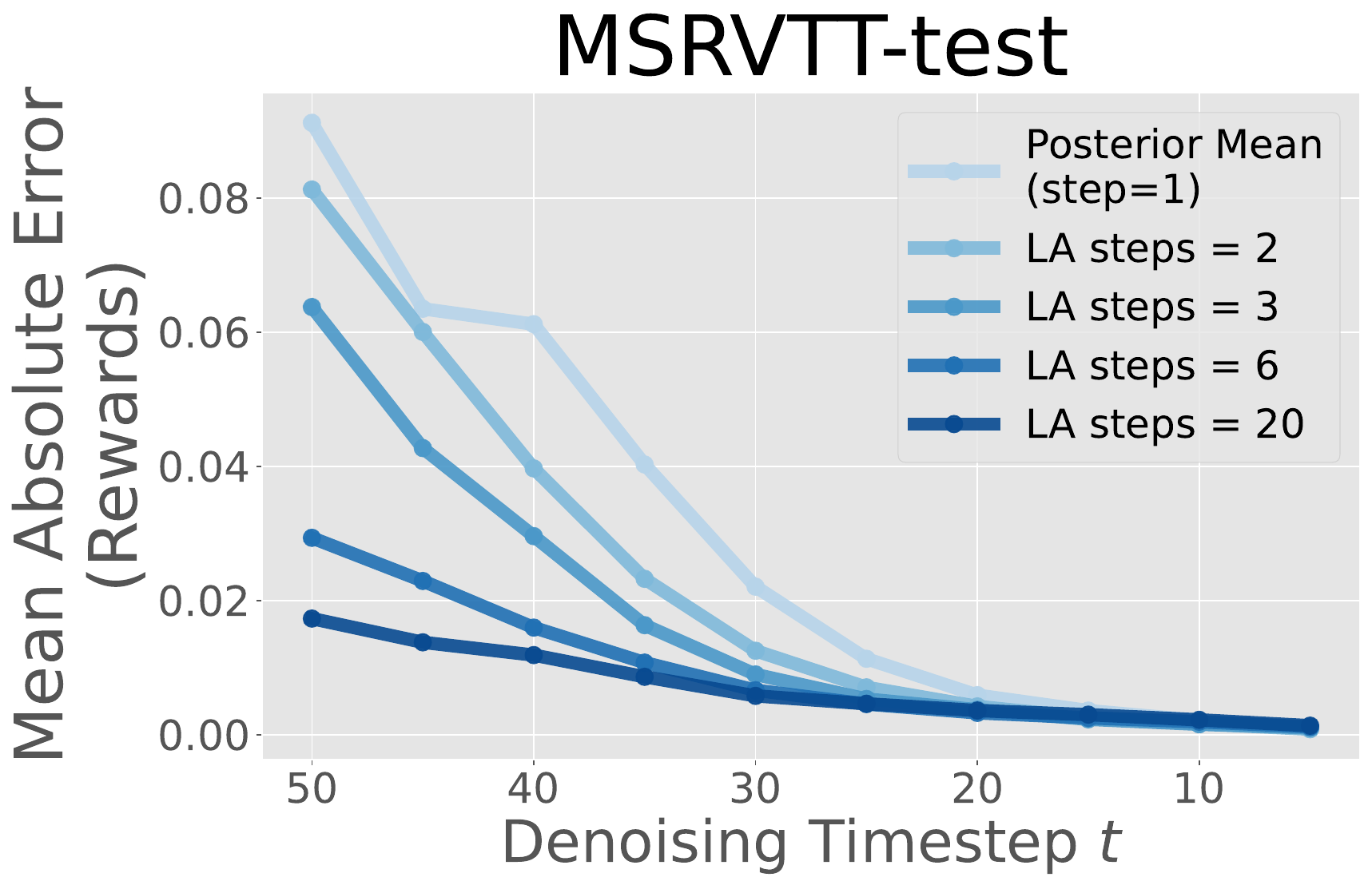}}
\subfigure{\includegraphics[width=0.49\textwidth]{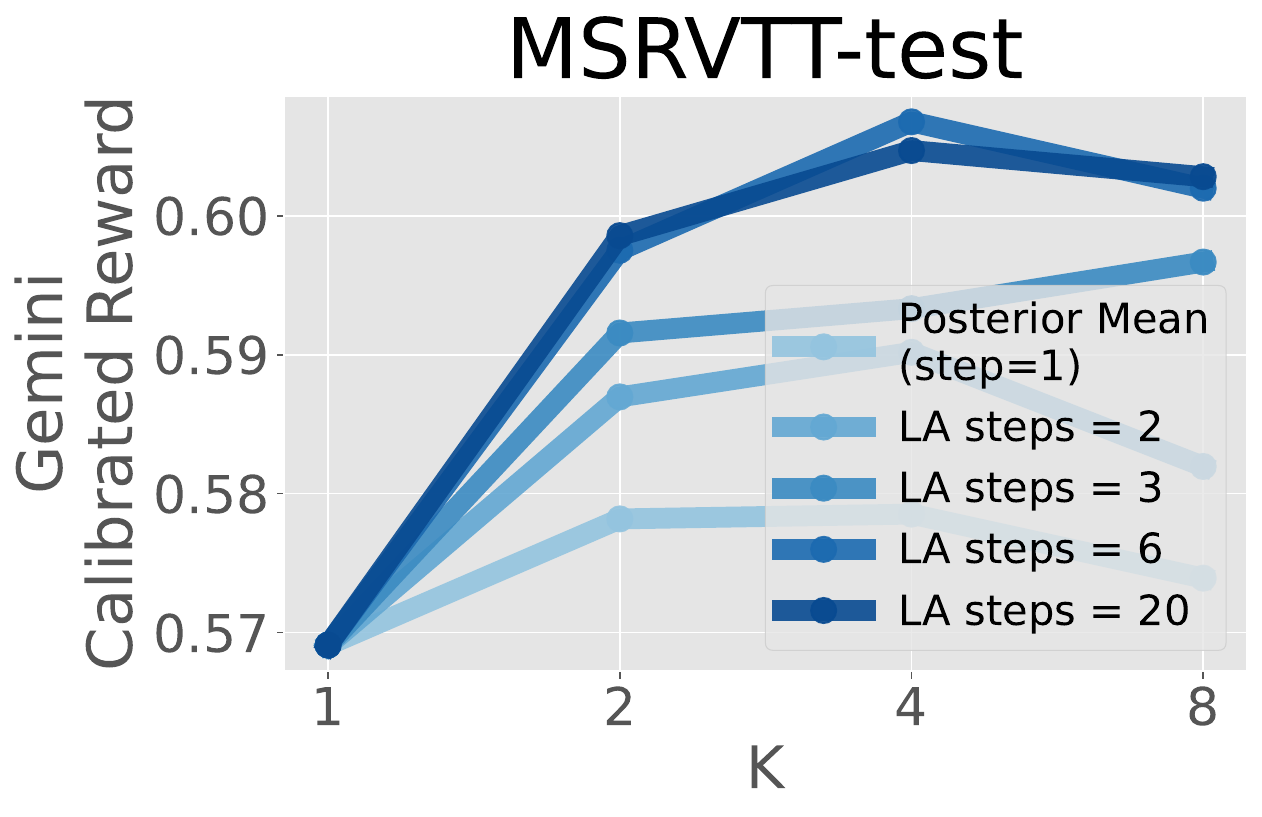}}
% \vskip -0.15in
\caption{(\textbf{Left}) Comparison of the reward estimation error for different LA steps \(T'\).
  We evaluate the reward predicted by the LA estimator, which projects \(\mathbf{z}_t\) to \(\tilde{\mathbf{z}}_{0|\tilde{t}(0)}\) in \(T'\) steps (Algorithm \ref{alg:lookahead}) and computes \(r'(\tilde{\mathbf{z}}_{0|\tilde{t}(0)})\), against the actual reward obtained by projecting \(\mathbf{z}_t\) to \(\mathbf{z}_0\) in \(t\) steps using a DDIM sampler (\(\eta=1.0\)) and evaluating \(r'(\mathbf{z}_0)\).
  (\textbf{Right}) Impact of \(T'\) on search performance. 
  Reward improves rapidly up to \(T'=6\) but saturates thereafter; using \(T'=20\) offers no measurable gain. 
  These results show that a modest \(T'\) is sufficient in practice.}
  \label{fig:reward_estimation_error}
\end{figure}

\autoref{fig:reward_estimation_error} (\textbf{Left}) demonstrates that increasing the number of reward estimation steps \( T' \) in the LA estimator leads to improved reward prediction performance for \( \mathbf{z}_t \) during the denoising process. 
This finding suggests that extending the LA steps enables a more effective search based on accurate reward predictions, particularly in the early stages of the denoising.
As shown in \autoref{fig:reward_estimation_error} (\textbf{Right}), enlarging the look-ahead horizon increases the reward gain to \(T' = 6\); beyond this point, e.g., \(T' = 20\) offers no significant benefit while multiplying the computational cost. 
Accordingly, we fix \(T' = 6\) in the main experiments, as it captures nearly all the attainable gains at minimal cost.
These results were obtained on Latte~\citep{ma2024latte} using a DDIM sampler.

\begin{figure}[ht]
\centering
% \vskip -0.1in
\subfigure{\includegraphics[width=0.49\textwidth]{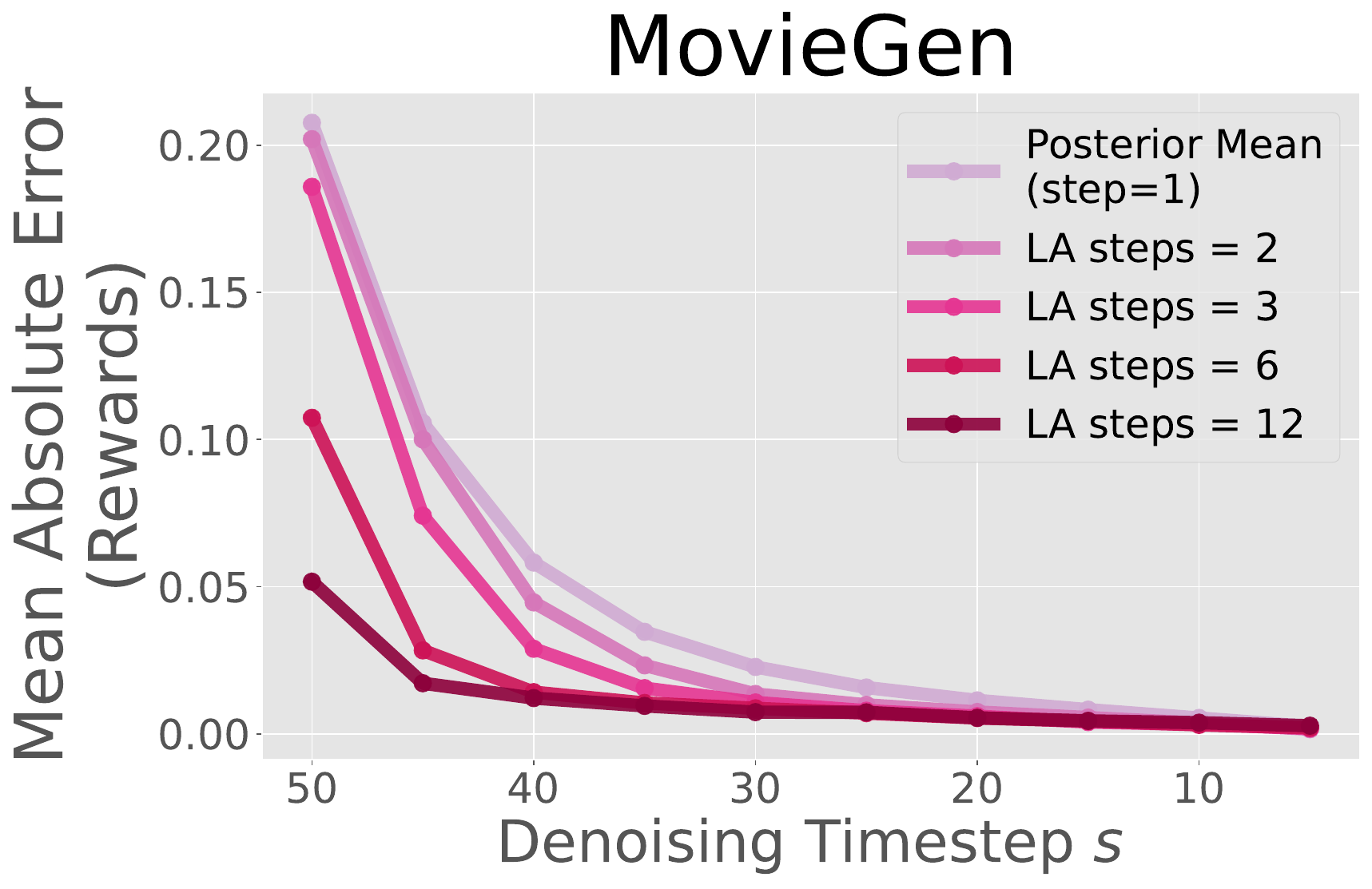}}
\subfigure{\includegraphics[width=0.49\textwidth]{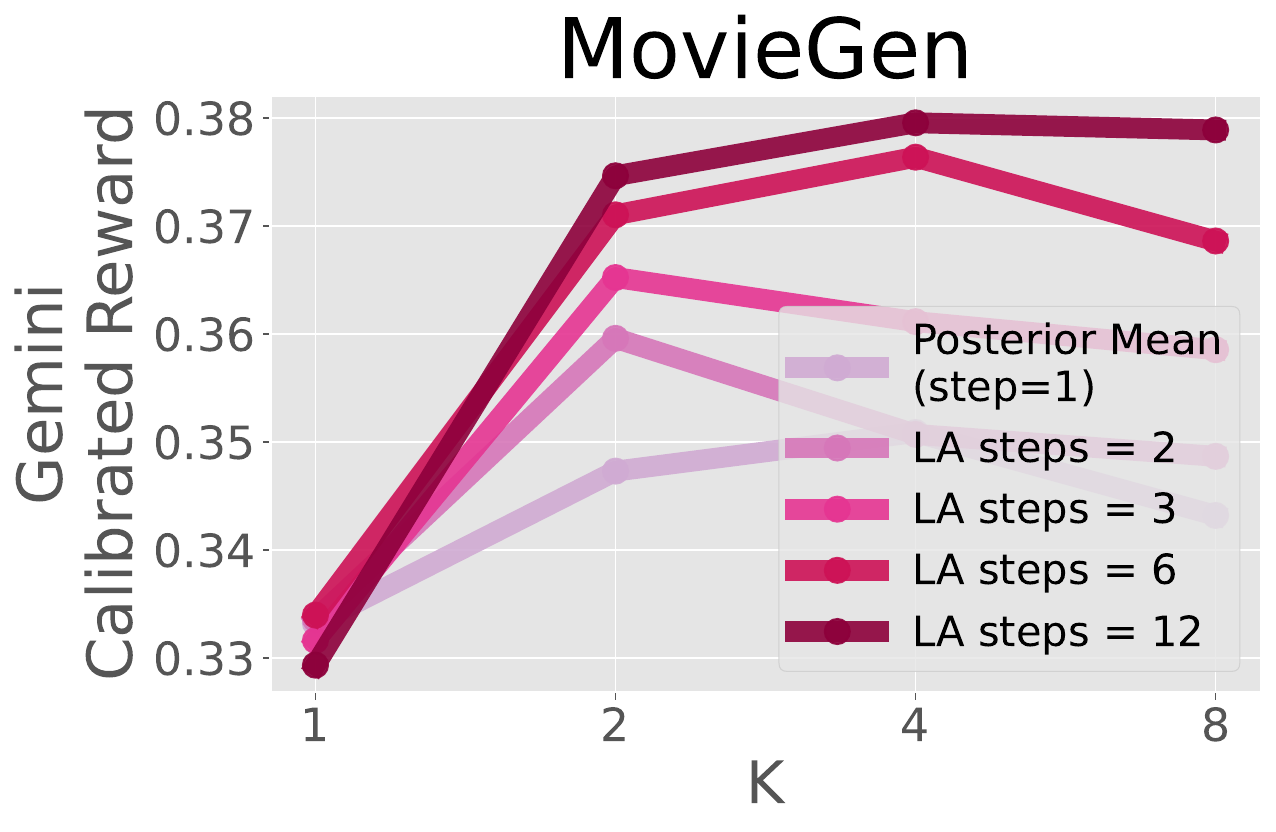}}
% \vskip -0.15in
\caption{(\textbf{Left}) Comparison of the reward estimation error for different LA steps \(M'\).
  We evaluate the reward predicted by the LA estimator, which projects \(\mathbf{z}_{t_s}\) to \(\tilde{\mathbf{z}}_{{t_M}|t_{\tilde{s}(0)}}\) in \(M'\) steps (Algorithm \ref{alg:lookahead_dpm}) and computes \(r'(\tilde{\mathbf{z}}_{{t_M}|t_{\tilde{s}(0)}})\), against the actual reward obtained by projecting \(\mathbf{z}_{t_s}\) to \(\mathbf{z}_{t_M}\) in \((M-s)\) steps using a SDE-DPMSolver++~\citep{lu2022dpmsolverpp} and evaluating \(r'(\mathbf{z}_M)\).
  (\textbf{Right}) Ablation study of \(M'\) on search performance with SDE-DPMSolver++~\citep{lu2022dpmsolverpp} on Wan 2.1-1.3B~\citep{wan2025}. 
  Reward improves rapidly up to \(M'=6\) but saturates thereafter; using \(M'=12\) offers no measurable gain. 
  These results show that a modest \(M'\) is sufficient in practice.}
  \label{fig:ablation_lookahead_dpmsolver}
\end{figure}

To verify that this behavior is not specific to DDIM sampler, we conducted the same ablation with an SDE-DPMSolver++~\citep{lu2022dpmsolverpp} on Wan 2.1-1.3B~\citep{wan2025} (\autoref{fig:ablation_lookahead_dpmsolver}). 
Note that the notation follows Algorithm \ref{alg:lookahead_dpm}.
Specifically, $M'$ in the SDE-DPMSolver++ setting corresponds to the $T'$ used with the DDIM sampler.
We observed the same pattern shown in \autoref{fig:reward_estimation_error}.
Increasing the look-ahead horizon $M'$ monotonically improves the LA estimator’s reward prediction.
Search reward gain up to roughly $M'=6$, after which gains saturate.
For example, $M'=12$ yields no measurable improvement while incurring substantially higher cost. 
This cross-sampler and cross-model consistency provides a practical guideline for choosing $M'$: a modest horizon ($\approx 6$) captures nearly all attainable benefit.

\clearpage
\subsection{Ablation Study for Diffusion Steps}
\label{sec:denoising_steps_ablation} 

\begin{figure}[ht]
  \centering
  % \vskip -0.1in
  \includegraphics[width=0.49\linewidth]{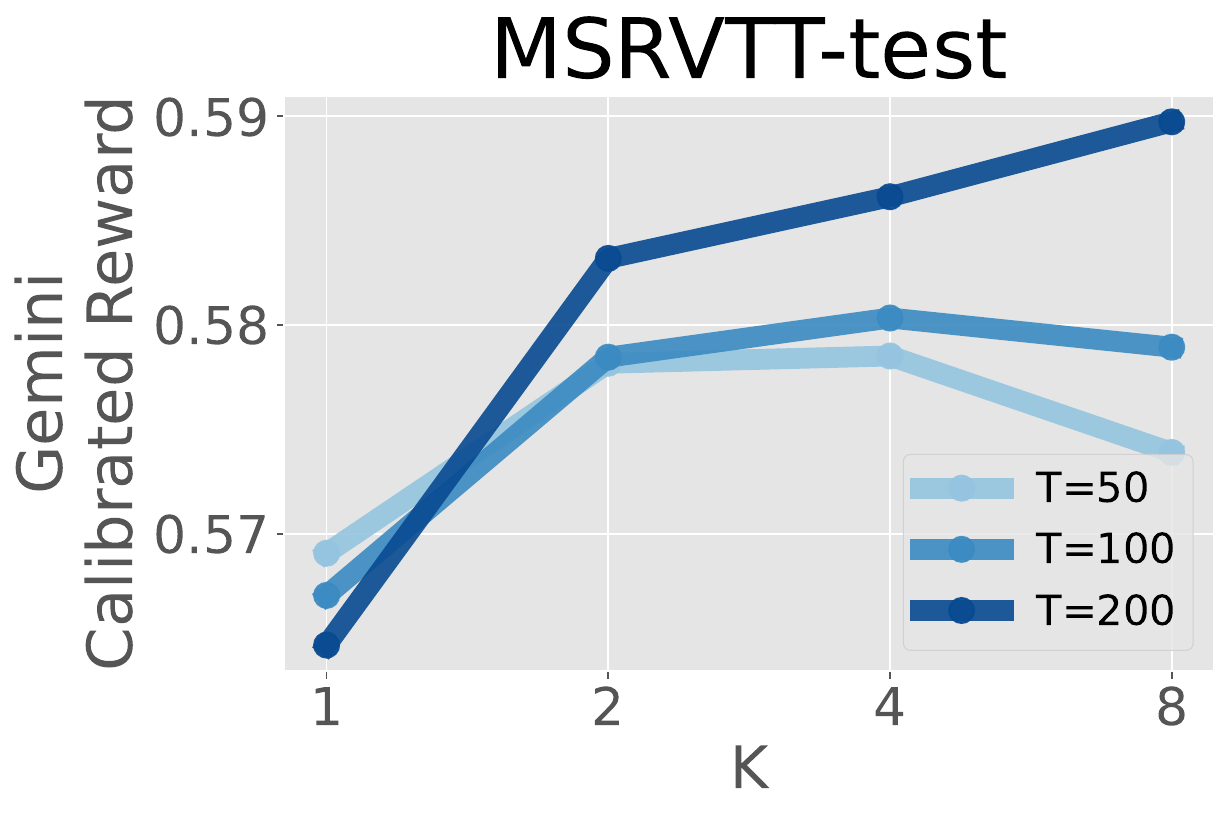}
  \includegraphics[width=0.49\linewidth]{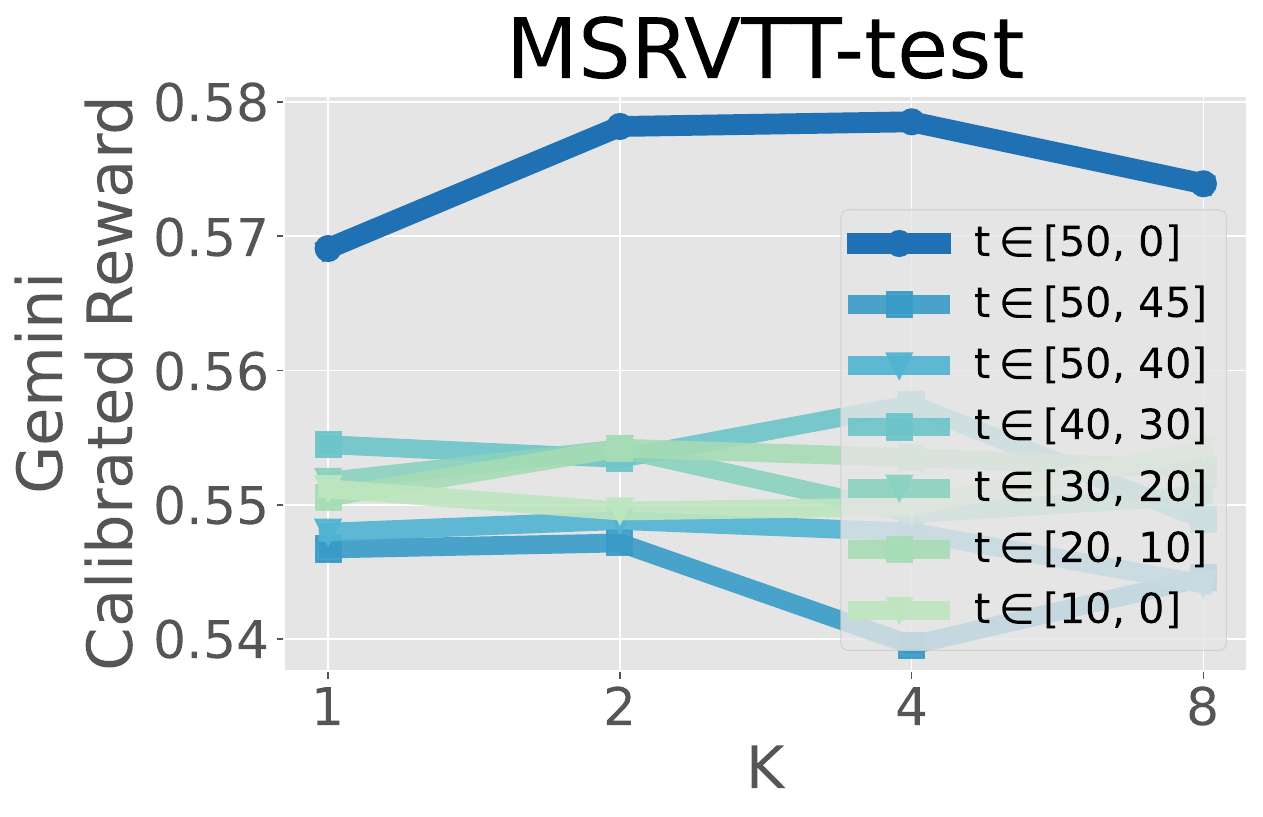}
  % \vskip -0.1in
  \caption{\textbf{(Left)} Scaling the denoising steps $T$. 
    \textbf{(Right)} Range of denoising steps \(t \in [50, 0]\) to apply search methods. While \citet{kim2024free2guide} applies GS in the first 5--10 steps, DLBS over the entire diffusion steps yields the largest improvement.
    }
  \label{fig:ablation_diffusion_steps}
\end{figure}

\textbf{Scaling Diffusion Steps}~~\autoref{fig:ablation_diffusion_steps} (\textbf{{Left}}) shows the performance when increasing the number of denoising steps $T$. 
Since DDIM exhibits fast convergence~\citep{song2021denoising}, BoN with a larger $T$ does not improve the reward much.
DLBS improves performance when scaling denoising steps to $T=200$ more than BoN, which implies that DLBS benefits from larger computational resources in denoising.
However, as \autoref{fig:cost_and_advanced_models} (\textbf{{Left}}) indicates, these gains are smaller than those obtained by widening the beam budget \(KB\) or leveraging the LA estimator.  

% We investigate which range of diffusion steps DLBS should be applied to for effective search. 
% Unlike \citet{kim2024free2guide}, who apply GS only during the initial 5--10 steps, our results in \autoref{fig:diffusion_range} show that applying DLBS throughout all steps leads to substantially better performance.
% This suggests that full-range application of DLBS is more effective than early-stage-only strategies.

\textbf{Range of Diffusion Steps for Search}~~We investigate which range of diffusion steps DLBS should be applied to for effective search. 
Unlike \citet{kim2024free2guide}, which applies GS only during the initial 5--10 steps, our results in \autoref{fig:ablation_diffusion_steps} (\textbf{{Right}}) show that applying DLBS throughout all steps leads to substantially better performance.
This suggests that applying DLBS entirely is more effective than focusing on the early stage.

\subsection{Ablation Study for $\eta$ in DDIM scheduler}
\label{sec:eta_ablation}

\begin{figure}[ht]
  \centering
  % \vskip -0.1in
  \includegraphics[width=0.55\linewidth]{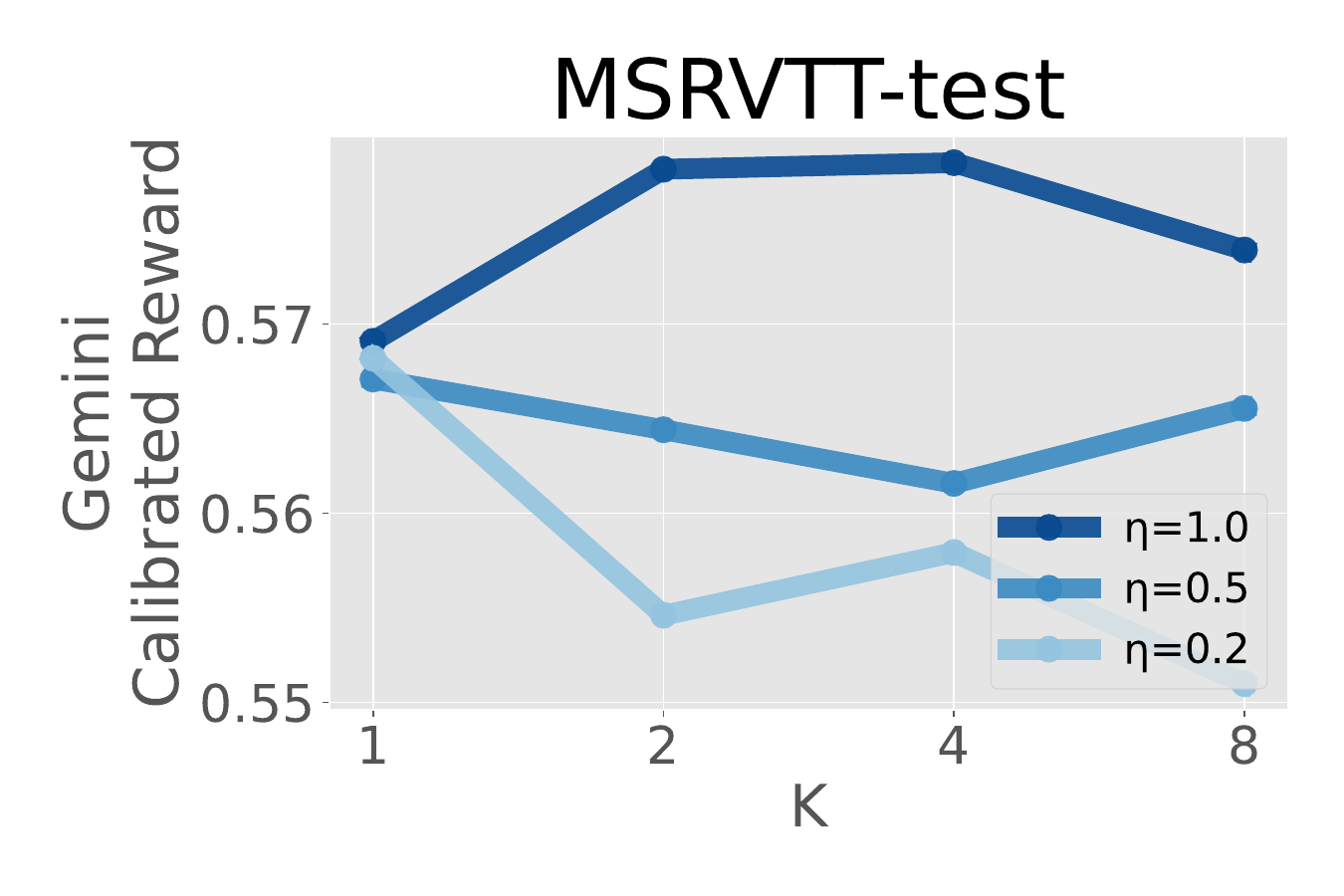}
  % \vskip -0.15in
  \caption{Comparison among different $\eta$ in DDIM sampler.}
  \label{fig:eta}
\end{figure}

\autoref{fig:eta} illustrates how varying the value of $\eta$ in DDIM influences search performance. 
Here, $\eta$ controls the degree of randomness in the DDIM scheduler: $\eta = 0.0$ corresponds to the deterministic version of DDIM, while $\eta = 1.0$ is equivalent to DDPM. As $\eta$ decreases below $1.0$, performance in terms of the final reward diminishes, presumably because lowering the randomness in the sampling process narrows the scope of exploration.

\clearpage
\subsection{Comparison with Gradient-Based Search Methods}
\label{sec:gradient_based_search}

\textbf{Applicability to diverse reward models}~~One advantage of our zero-order search framework is its applicability to reward models for which computing gradients is computationally prohibitive (e.g., large-scale VLMs) or fundamentally impossible (e.g., human evaluators or external API-based models). 
To demonstrate this, we simulated search with human feedback by employing VideoScore~\citep{he2024videoscore}, a VLM-based reward model trained on human evaluations, as the reward function. 
As shown in \autoref{tab:videoscore_dlbs}, DLBS-LA with VideoScore as the evaluator achieved substantial improvements over the vanilla baseline, suggesting that high-performance VLMs or human evaluators can, in principle, be directly incorporated as reward functions in our framework.

\begin{table*}[ht]
  \centering
  \caption{Results of DLBS-LA with VideoScore as the evaluator, illustrating its applicability to reward models for which computing gradients is computationally prohibitive (e.g., large-scale VLMs) or fundamentally impossible (e.g., human evaluators or external API-based models).}
  \label{tab:videoscore_dlbs}
  \begin{tabular}{lc}
    \toprule
    Method & VideoScore \\
    \midrule
    Latte & 2.40 \\
    + DLBS-LA ($K\!B=4$, $T’=6$) with VideoScore & \textbf{2.69} \\
    \bottomrule
  \end{tabular}
\end{table*}

\textbf{Efficiency in time and memory}~~We further compared DLBS with DAS~\citep{kim2025das}, a first-order gradient-based method based on Sequential Monte Carlo. 
Experiments were conducted on Stable Diffusion 1.5~\citep{rombach2022ldm} with LAION Aesthetic V2~\citep{laion2022aesthentic} as the reward model, using an NVIDIA RTX 6000 Ada (48GB). 
\autoref{tab:das_combined} shows that under the assumption of equal execution time, DLBS (a zero-order method) takes the lead because it can have a larger search budget, which refers to the number of particles used for search, i.e., $K\!B$ for DLBS and $N$ for DAS. 
Our observation that a zero-order method achieves better performance than a first-order method under the equal execution time aligns with prior findings on inference-time search for image generation~\citep{singhal2025general}.

Gradient-based search methods also exhibit significant increases in memory usage due to gradient computations required for the reward function and the VAE decoder. 
In other words, gradient-based methods are actually inefficient in terms of memory cost. 
The results shown in \autoref{tab:das_combined} are based on a single evaluator and a single frame (i.e., image generation). 
Note that for video generation, as mentioned in \autoref{fig:hist_gemini}, a single evaluator metric does not correlate well with perceptual quality, necessitating the combination of multiple evaluators, which roughly multiplies the memory requirements for gradient calculation. 
Additionally, video generation models do not decode just one frame from the VAE (maximum frames are 81 for Wan 2.1~\citep{wan2025} and 49 for CogVideoX~\citep{yang2024cogvideox}). 
The gradient increase would be significantly larger in video generation than in image generation, making gradient-based methods almost impossible in practice.
For reference, the memory usage of vanilla Latte, DLBS-LA, and DAS in \autoref{tab:memory_usage}.

\begin{table*}[ht]
  \centering
  \caption{Comparison between DLBS and DAS on Stable Diffusion 1.5 with LAION Aesthetic V2 reward. DLBS (a zero-order method) achieves better performance than DAS (a first-order method) under equal execution time.}
  \label{tab:das_combined}
  \begin{tabular}{lccc}
    \toprule
    Method & Score & Time (sec) & Memory (GB) \\
    \midrule
    SD 1.5 & 5.81 & 2 & 4.3 \\
    \midrule
    + DAS ($N=8$) & 6.59 & 108 & 14.2 \\
    + DLBS ($K=8, B=2$) & \textbf{6.63} & \textbf{103} & \textbf{5.9} \\
    \midrule
    + DAS ($N=16$) & 6.68 & 220 & 14.2 \\
    + DLBS ($K=16, B=2$) & \textbf{6.69} & \textbf{209} & \textbf{5.9} \\
    \bottomrule
  \end{tabular}
\end{table*}

\begin{table*}[ht]
  \centering
  \caption{Memory usage of search methods on Latte. The memory advantage of DLBS-LA (a zero-order method) becomes more critical in video generation.}
  \label{tab:memory_usage}
  \begin{tabular}{lccc}
    \toprule
    Method & Latte & + DLBS-LA & + DAS \\
    \midrule
    Memory Usage (GB) & 16.3 & 38.9 & $>$48.0 (\textbf{OOM}) \\
    \bottomrule
  \end{tabular}
\end{table*}

% We note that our observation, that DLBS (a zero-order method) and DAS (a first-order method) achieve comparable performance, aligns with prior findings on inference-time search for image generation~\citep{singhal2025general}. 
% In~\citep{singhal2025general}, first-order gradient-based guidance and a zero-order Sequential Monte Carlo-based method demonstrated equivalent performance, suggesting that this is not a novel trend. 
% One possible explanation is that computing first-order gradients requires mapping from latent space to image space via Tweedie’s formula, which introduces large gradient noise due to the inherent noisiness of reward estimation. 
% This instability may hinder efficient reward maximization, thereby diminishing the expected advantage of first-order methods.

% Taken together, these results demonstrate that while gradient-based methods can be competitive in image generation under controlled conditions, their excessive memory cost and infeasibility in video generation make zero-order approaches such as DLBS the more practical and scalable solution.

\clearpage
\subsection{Comparison with Other Inference-Time Search Methods}
\label{sec:gradient_free_search}

\begin{figure}[ht]
  \centering
  \includegraphics[width=0.6\linewidth]{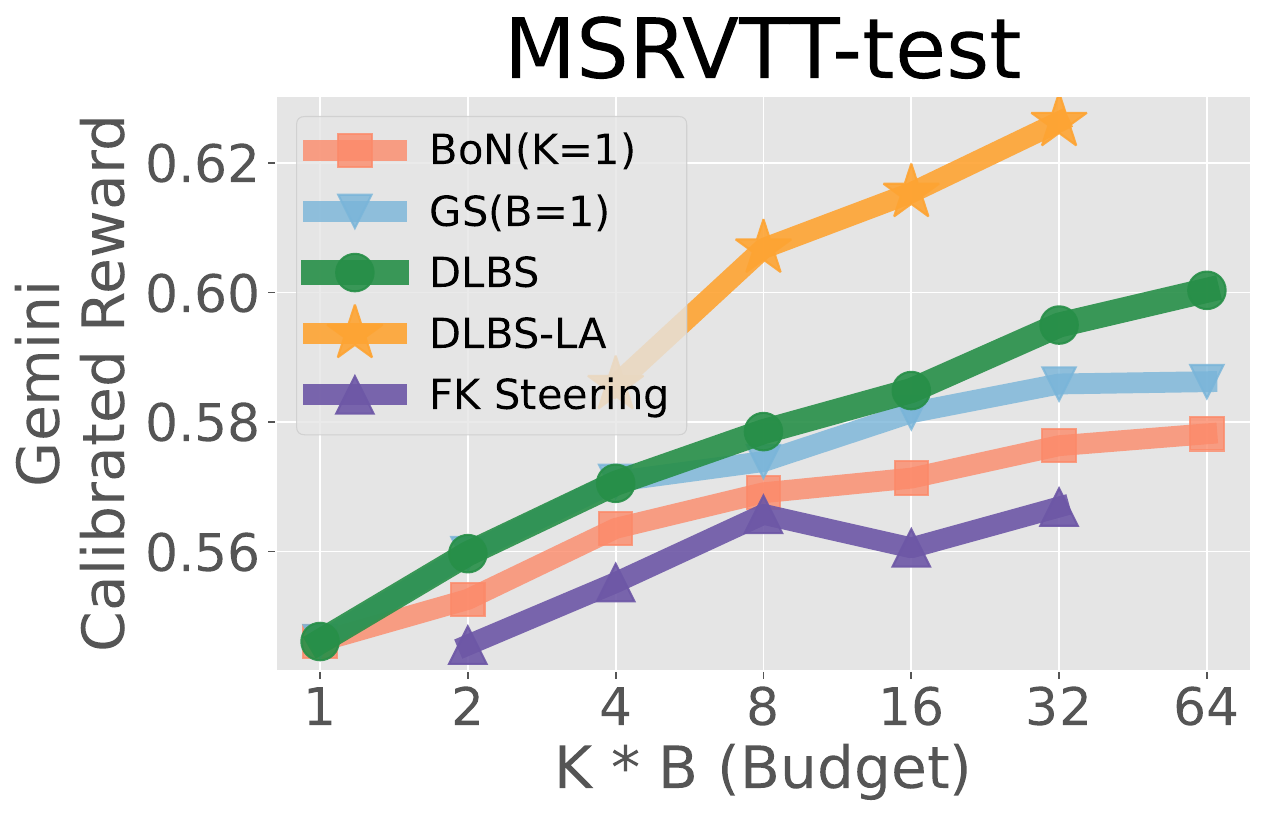}
  % \vskip -0.08in
  \caption{Comparison with FK steering~\citep{singhal2025general}.}
  \label{fig:fkd_steering}
\end{figure}

% \textbf{Sequential Monte Carlo Based Method}~~
Concurrently with our work, inference-time search based on zero-order sequential Monte Carlo (SMC) has been proposed. We include a comparison with FK Steering~\citep{singhal2025general} in \autoref{fig:fkd_steering}, where the resampling mechanism in the SMC-based methods does not occur frequently enough, preventing them from surpassing BoN performance, in our text-to-video experiments.

\subsection{Scalability to Long Videos}
\label{sec:long_scalability}

To demonstrate that our method is scalable even when extending the frame count to the model's maximum, we conducted experiments with the maximum frames for CogVideoX-5B (49 frames, 6 seconds) and Wan 2.1-1.3B (81 frames, 5 seconds). 
The reward, which was calibrated using 33-frame, 2-second videos generated by Wan 2.1-1.3B, was applied as-is. 
As shown in \autoref{tab:long_cogvideo} and \autoref{tab:long_wan}, we confirmed that even with longer frames, the reward values could be improved more efficiently than the BoN baseline.

\begin{table*}[ht]
  \centering
  \caption{Scalability results on CogVideoX-5B with 49 frames, 6 seconds using DEVIL-very-high prompts.}
  \label{tab:long_cogvideo}
  \begin{tabular}{lccc}
    \toprule
    Method & $KB$ & Reward & Inference Compute (NFE) \\
    \midrule
    CogVideoX-5B & 1   & 0.429 & 50 \\
    \midrule
    + BoN     & 64  & 0.474 & 3200 \\
            & 128 & 0.478 & 6400 \\
    \midrule
    + DLBS    & 16  & 0.481 & 1200 \\
            & 32  & 0.490 & 2400 \\
    \midrule
    + DLBS-LA & 8   & \textbf{0.497} & 2500 \\
            & 16  & \textbf{0.517} & 4900 \\
    \bottomrule
  \end{tabular}
\end{table*}

\begin{table*}[ht]
  \centering
  \caption{Scalability results on Wan 2.1-1.3B with 81 frames, 5 seconds using MovieGen prompts.}
  \label{tab:long_wan}
  \begin{tabular}{lccc}
    \toprule
    Method & $KB$ & Reward & Inference Compute (NFE) \\
    \midrule
    Wan 2.1-1.3B & 1   & 0.313 & 50 \\
    \midrule
    + BoN     & 128 & 0.357 & 6400 \\
            & 256 & 0.360 & 12800 \\
    \midrule
    + DLBS    & 16  & 0.357 & 1200 \\
            & 32  & 0.371 & 2400 \\
    \midrule
    + DLBS-LA & 8   & \textbf{0.373} & 2500 \\
            & 16  & \textbf{0.393} & 5000 \\
    \bottomrule
  \end{tabular}
\end{table*}

% \subsection{Compatibility of DLBS with Fine-tuning Methods}

% In image generation, prior studies have shown that allocating additional computation at inference time can be more effective than relying solely on post-training approaches~\citep{ma2025inferencetime}. 
% We find a similar trend in video generation. To examine this, we apply a representative fine-tuning method, VideoDPO~\citep{liu2024videodpo}, to VideoCrafter2~\citep{chen2024videocrafter2}. 
% As shown in \autoref{tab:dpo_dlbs}, VideoDPO alone brings negligible improvement over the baseline. However, when combined with DLBS, the performance increases substantially on both DEVIL-high and MSRVTT-test.
% These results indicate that DLBS is complementary to post-training methods, enabling further performance gains even after fine-tuning.

% \begin{table*}[h]
% \centering
% \caption{Performance of DLBS with DPO finetuned VideoCrafter2 on DEVIL-high and MSRVTT-test datasets. 
% While DPO alone yields marginal improvements, combining it with DLBS leads to notable gains, demonstrating the compatibility of inference-time search with fine-tuning approaches.}
% \begin{tabular}{lcc}
% \toprule
% \textbf{Method} & \textbf{DEVIL-high} & \textbf{MSRVTT-test} \\
% \midrule
% VideoCrafter2        & 0.337 & 0.555 \\
% + DPO                & 0.335 & 0.556 \\
% + DPO \& DLBS        & \textbf{0.359} & \textbf{0.576} \\
% \bottomrule
% \end{tabular}
% \label{tab:dpo_dlbs}
% \end{table*}

\clearpage
\subsection{DLBS with Larger Text-to-Video Models}
\label{sec:large_models_full}
We have tested our method on VideoCrafter2~\citep{chen2024videocrafter2} (1.9B) and CogVideoX-5B, 2B~\citep{yang2024cogvideox} and Wan 2.1-14B, 1.3B~\citep{wan2025}.
Our experiments confirm that our DLBS and DLBS-LA yield significant improvements, indicating their effectiveness can be observed in larger video generation models.

\begin{figure}[h]
\centering
\begin{minipage}[b]{\textwidth}
    \centering
    \includegraphics[width=\textwidth]{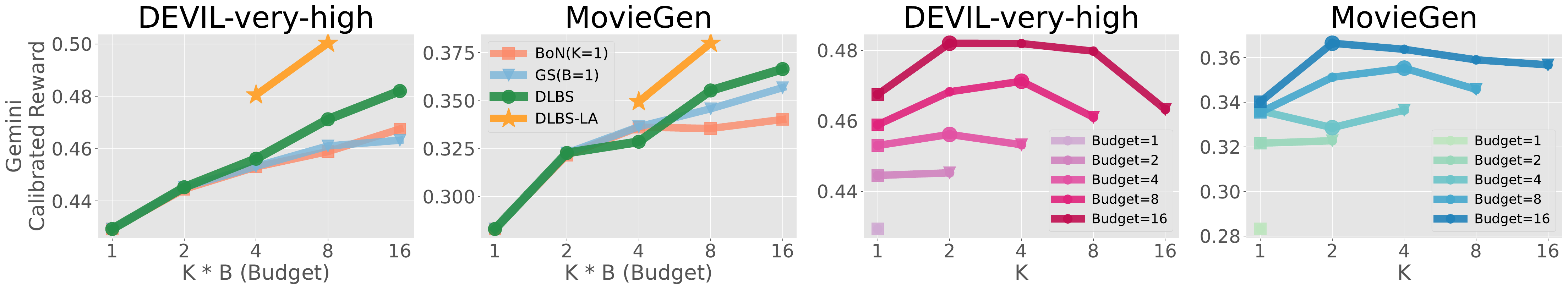}
    % \vskip -0.05in
    \caption{Inference-time search with CogVideoX-5B~\citep{yang2024cogvideox}.}
    \label{fig:advanced_models_cogvideo_full}
\end{minipage}
\vskip 0.2in

\begin{minipage}[b]{\textwidth}
    \centering
    \includegraphics[width=\textwidth]{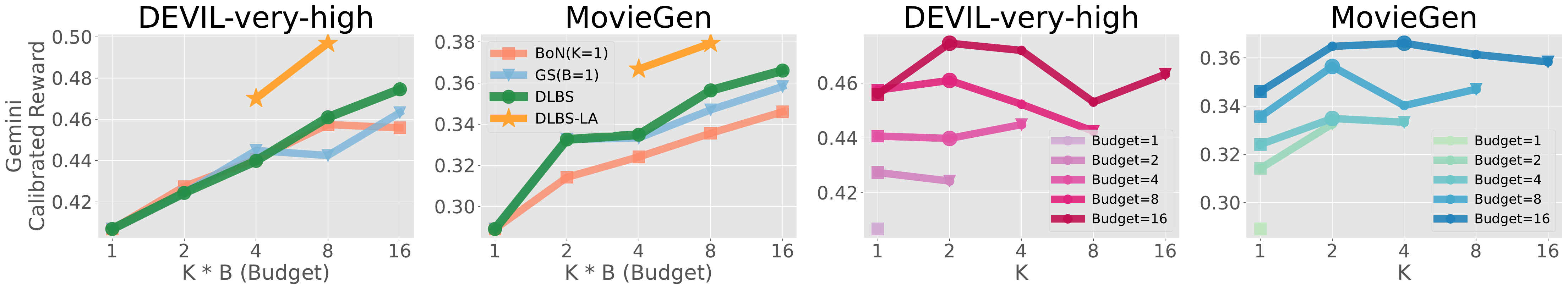}
    % \vskip -0.05in
    \caption{Inference-time search with CogVideoX-2B~\citep{yang2024cogvideox}.}
    \label{fig:advanced_models_cogvideo_2b_full}
\end{minipage}
\vskip 0.2in

\begin{minipage}[b]{\textwidth}
    \centering
    \includegraphics[width=\textwidth]{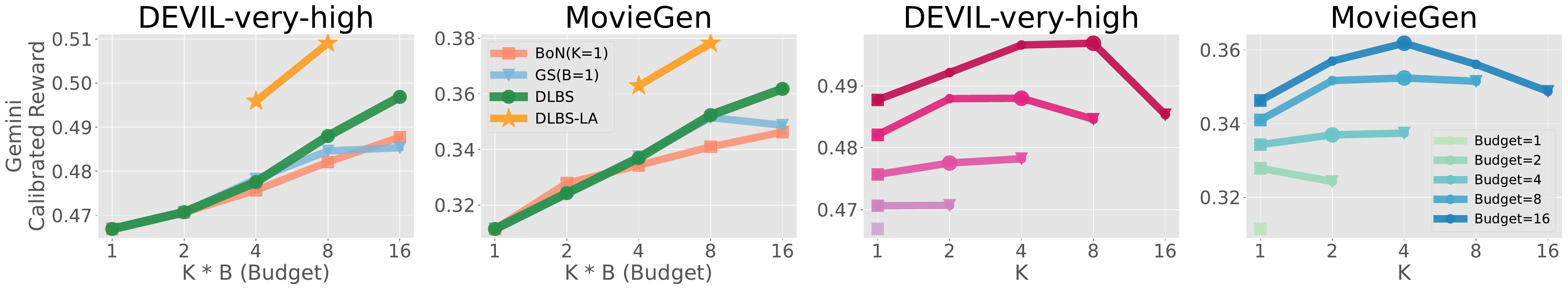}
    % \vskip -0.05in
    \caption{Inference-time search with Wan 2.1-14B~\citep{wan2025}.}
    \label{fig:advanced_models_wan_full}
\end{minipage}
\vskip 0.2in

\begin{minipage}[b]{\textwidth}
    \centering
    \includegraphics[width=\textwidth]{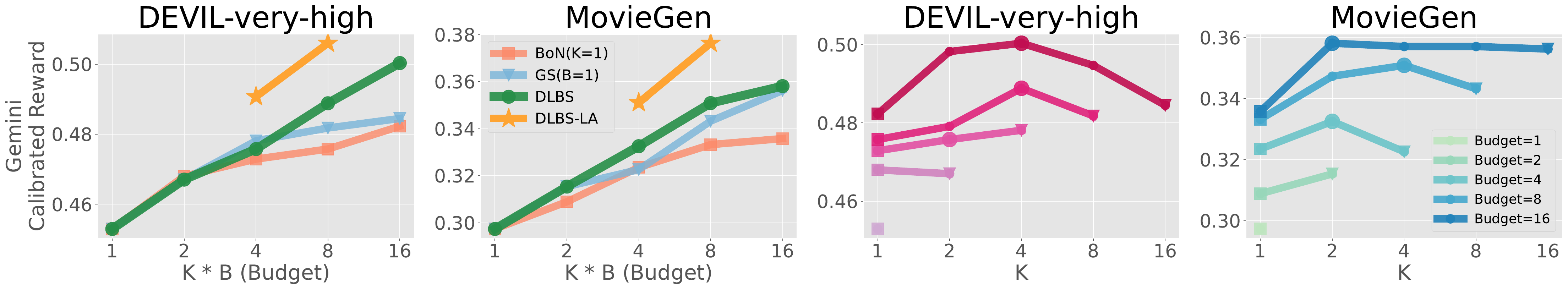}
    % \vskip -0.05in
    \caption{Inference-time search with Wan 2.1-1.3B~\citep{wan2025}.}
    \label{fig:advanced_models_wan_1b_full}
\end{minipage}
\vskip 0.2in

\begin{minipage}[b]{0.49\textwidth}
    \centering
    \includegraphics[width=\textwidth]{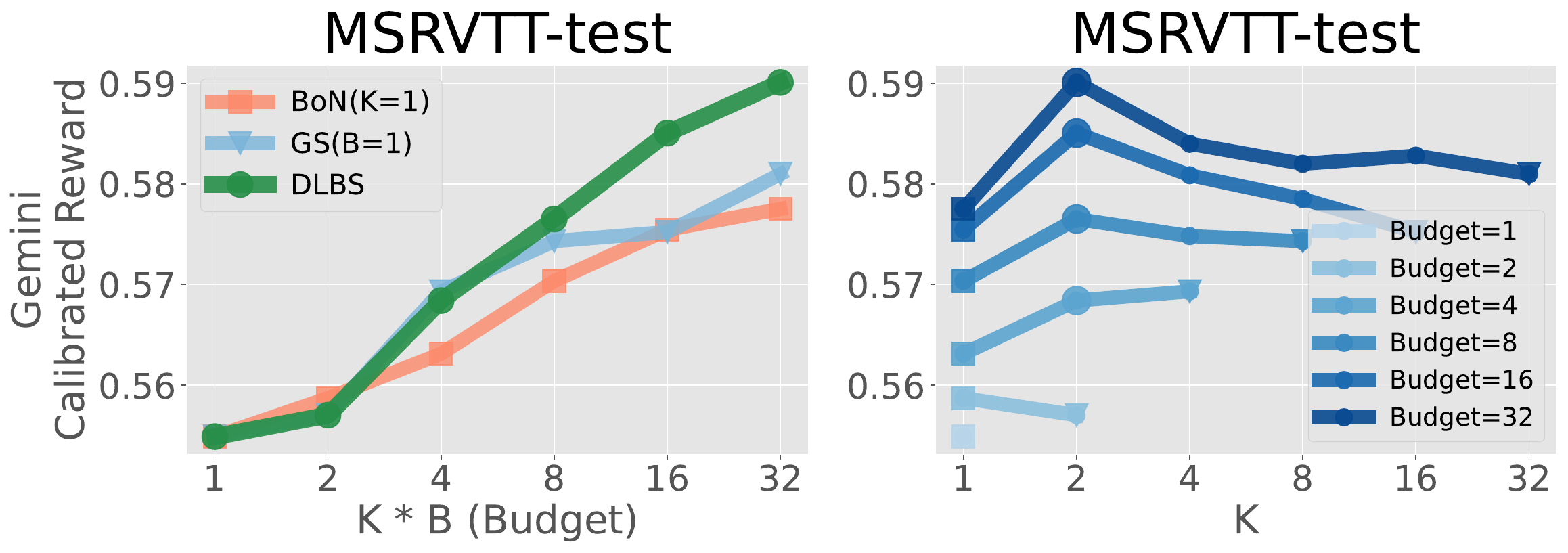}
    % \vskip -0.05in
    \caption{Search with VideoCrafter2~\citep{chen2024videocrafter2}.}
    \label{fig:videocrafter2}
\end{minipage}
\begin{minipage}[b]{0.49\textwidth}
    \centering
    \includegraphics[width=\textwidth]{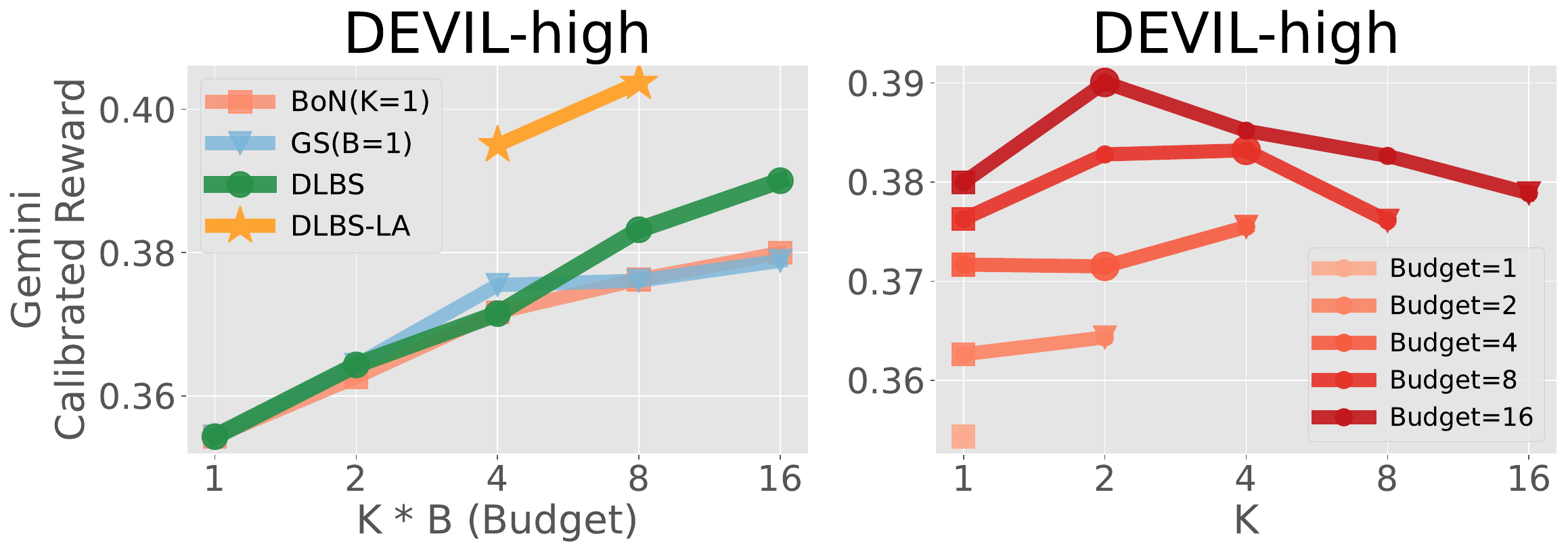}
    % \vskip -0.05in
    \caption{Search with CogVideoX-5B~\citep{yang2024cogvideox}.}
    \label{fig:cogvideox}
\end{minipage}
% \vskip -0.2in
\end{figure}

% \begin{table}[ht]
% \centering
% \caption{Inference-time search with Wan2.1-14B~\citep{wan2025}. We adopt a subset of DEVIL-high prompts for the prompt set and employ a Gemini-calibrated reward as the evaluation metric.}
% \begin{tabular}{lcccc}
% \toprule
% Method & GS~($KB=1$) & BoN~($KB=8$) & DLBS~($KB=8$) & DLBS-LA~($KB=8, T'=6$) \\
% \midrule
% Reward & 0.536 & 0.561 & 0.563 & \bf{0.588} \\
% \bottomrule
% \end{tabular}
% \label{tab:wan}
% \end{table}

% \begin{figure}[ht]
%   \centering
%   \includegraphics[width=0.6\linewidth]{figures/videocrafter.pdf}
%   \vskip -0.1in
%   \caption{Inference-time search on VideoCrafter2~\citep{chen2024videocrafter2}.}
%   \vskip -0.05in
%   \label{fig:videocrafter2}
% \end{figure}

\clearpage
\subsection{Further Results in AI and Human Evaluation}
\label{sec:vlm_eval_appendix}

\begin{figure}[ht]
  \centering
  \includegraphics[width=\linewidth]{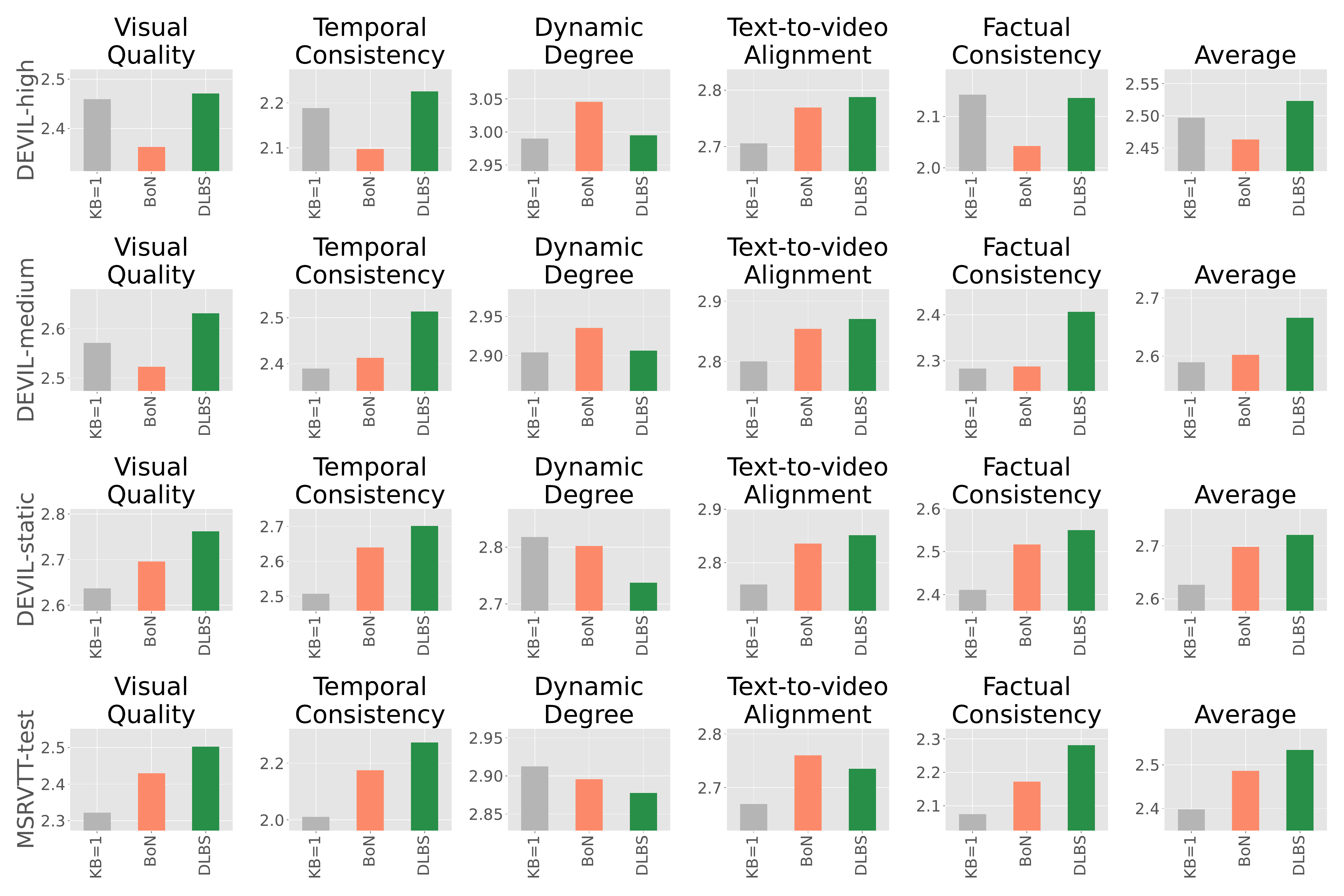}
  \caption{DLBS on calibrated reward also improves another qualitative metric, the most, VideoScore~\citep{he2024videoscore}, which is not involved in a reward calibration.}
  \label{fig:videoscore_full}
\end{figure}

We show full results of evaluations using VideoScore~\citep{he2024videoscore}, a metric trained on human judgments that evaluates videos at 8 fps across five dimensions and outputs scores ranging from 1.0 to 4.0 (see \autoref{fig:videoscore_full}).
Under this metric, videos generated with our DLBS consistently outperformed those generated without search and with BoN.

\begin{figure}[ht]
  \centering
  % \begin{minipage}[t]{0.74\textwidth}
    \centering
    \subfigure{\includegraphics[width=0.49\textwidth]{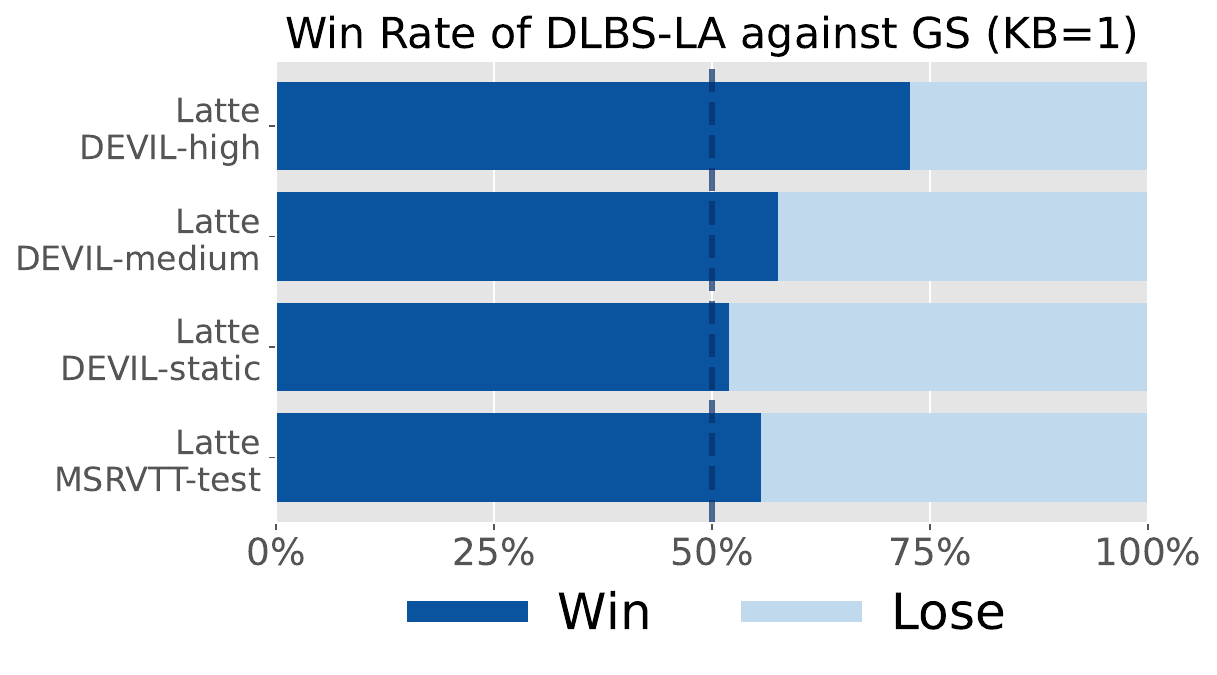}}
    \subfigure{\includegraphics[width=0.49\textwidth]{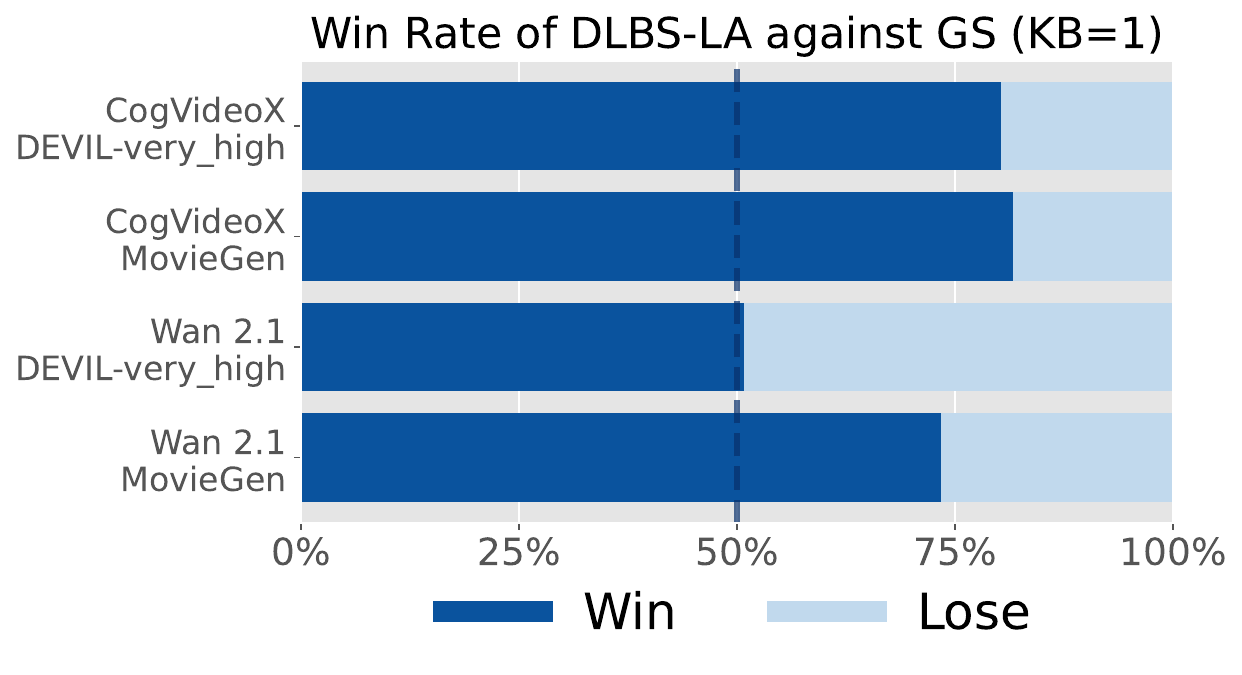}}
    \caption{Human evaluation results. We searched videos using the Gemini calibrated reward and asked three human evaluators to compare outputs from GS ($KB=1$) and DLBS-LA ($KB=8$, $T’=6$). 
    % Each video was rated as win (better quality) or lose (worse quality). 
    "Win" indicates that the video generated by DLBS-LA was preferred.}
    \label{fig:human_eval_full}
\end{figure}

We also show additional results of human judgment by three human evaluators (see \autoref{fig:human_eval_full}). 
These experiments confirmed that, regarding human preference (win rate), content generated with our search strategy consistently outperformed content produced without search.

\subsection{Qualitative Results for DLBS}
Qualitative results are shown in \url{https://sites.google.com/view/t2v-dlbs}.

\clearpage
\subsection{DLBS for Image Generation}
We adopt PixArt-$\alpha$~\citep{chen2023pixartalpha} as our base text-to-image generation model.
For evaluation, we directly reuse the prompt set of 45 common animal categories from prior works~\citep{black2024training, yeh2024sampling}.
As a reward model, we employ the LAION aesthetic predictor~\citep{laion2022aesthentic} to assess image quality.

\begin{figure}[ht]
  \centering
  \includegraphics[width=0.8\linewidth]{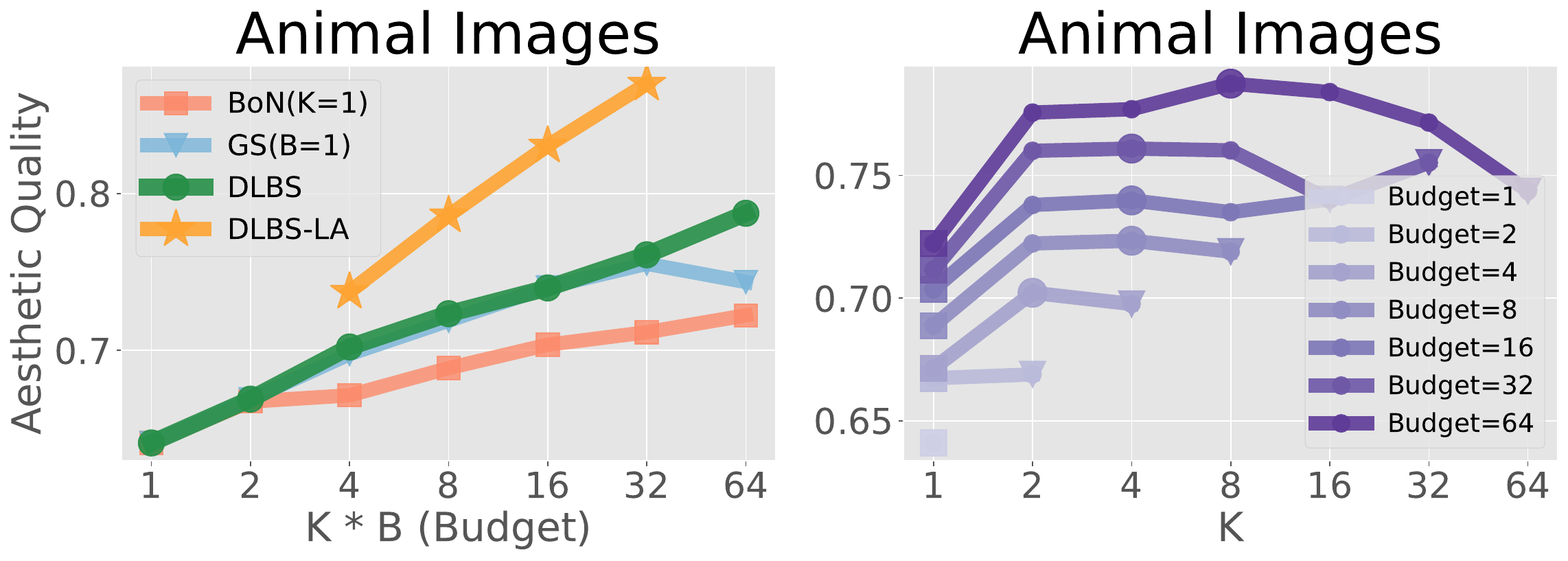}
  % \vskip -0.05in
  \caption{Inference-time search with PixArt-$\alpha$~\citep{chen2023pixartalpha}.}
  \label{fig:pixart-alpha}
\end{figure}

\begin{figure}[ht]
  \centering
  \includegraphics[width=0.8\linewidth]{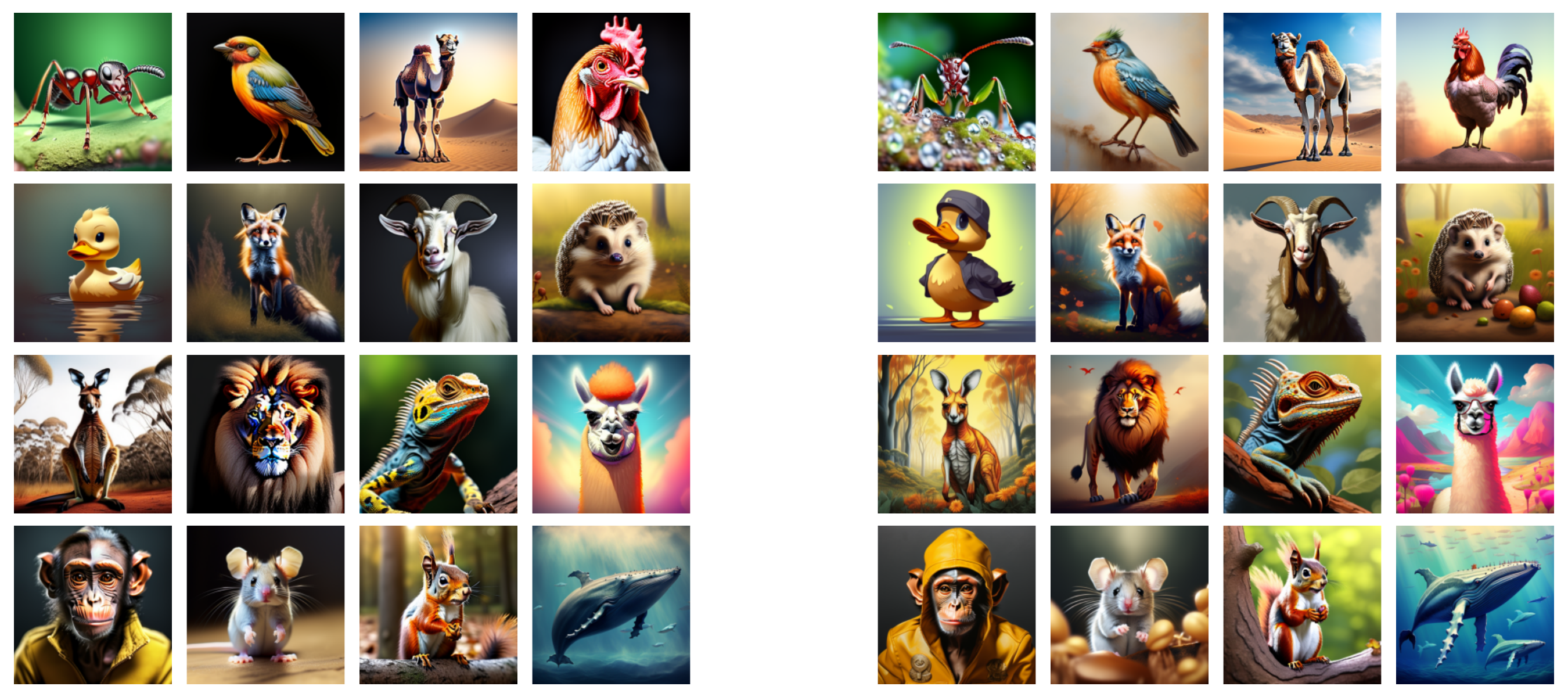}
  % \vskip -0.05in
  \caption{Qualitative Results in inference-time search with PixArt$-\alpha$~\citep{chen2023pixartalpha}. \textbf{(Left)} GS ($KB=1$). \textbf{(Right)} DLBS-LA ($KB=32, T'=6$). }
  \label{fig:pixart-alpha_qualitative}
\end{figure}

\section{Extended Related Works}
\label{sec:extended_related_work}
\textbf{Aligning Diffusion Models via Finetuning}~~
Alignment by finetuning text-conditioned models has been investigated for image~\citep{lee2023aligning} and video~\citep{furuta2024improving,wu2024boosting} generation.
Typically, LoRA~\citep{hu2021lora} in a backbone model~\citep{cicek20163dunet} is finetuned through policy gradient~\citep{black2024training,fan2023reinforcement,zhang2024largescale}, direct preference optimization~\citep{yuan2024selfplay,na2024boost,yang2023using,liang2024stepaware,wallace2023diffusion}, reward-weighted regression~\citep{dong2023raftalign}, or direct reward gradient~\citep{prabhudesai2024video,clark2024directly,wu2024boosting,li2024t2vturbo}.
Some train an extra model for better initial noise space~\citep{qi2024noise,ahn2024noise,zhou2024goldennoise}.
In contrast, we focus on the search over the denoising process at inference time, which does not require any model updates and may not degrade the original performance.

\textbf{Evaluation of Text-to-Video Generation}~~
While there are several conventional metrics for video generation (or the one repurposed from image generation) such as SSIM~\citep{wang2004}, IS~\citep{salimans2016improvedgans}, LPIPS~\citep{zhang2018perceptual}, or FVD~\citep{unterthiner2019fvd}, those are not always suitable to evaluate how the quality of contents in video is, which is much more emphasized in text-to-video generation~\citep{wu2024better}.
It has been a long-standing challenge to comprehensively and semantically evaluate the dynamics of contents or physical commonsense in generated videos~\citep{bansal2024videophy,liao2024evaluation}.
To deal with that, VBench~\citep{huang2023vbench} has recently been proposed as a suite of holistic evaluations for text-to-video generation to reflect the perceptual aspect of the quality, such as consistency, smoothness, aesthetics of contents, or text--video alignment.
Moreover, inspired by the success in LLMs~\citep{bai2022constitutional,lee2023rlaif,furuta2024geometric}, we could leverage VLMs, which become more capable these days, as a proxy of human evaluation of the contents~\citep{wu2024boosting}; by finetuning CLIP-based models~\citep{he2024videoscore,radford2021clip,ma2022xclip}, or prompting GPT-4o~\citep{openai2023gpt4} or Gemini~\citep{geminiteam2023gemini}.
Our paper adopts AI feedback from VLMs as an alternative to human rater, and proposes a recipe to calibrate a reward to other sources of feedback (such as AI or human feedback), by considering a linear combination of fine-grained metrics.

\newpage
\section*{NeurIPS Paper Checklist}

%%% BEGIN INSTRUCTIONS %%%
The checklist is designed to encourage best practices for responsible machine learning research, addressing issues of reproducibility, transparency, research ethics, and societal impact. Do not remove the checklist: {\bf The papers not including the checklist will be desk rejected.} The checklist should follow the references and follow the (optional) supplemental material.  The checklist does NOT count towards the page
limit. 

Please read the checklist guidelines carefully for information on how to answer these questions. For each question in the checklist:
\begin{itemize}
    \item You should answer \answerYes{}, \answerNo{}, or \answerNA{}.
    \item \answerNA{} means either that the question is Not Applicable for that particular paper or the relevant information is Not Available.
    \item Please provide a short (1–2 sentence) justification right after your answer (even for NA). 
   % \item {\bf The papers not including the checklist will be desk rejected.}
\end{itemize}

{\bf The checklist answers are an integral part of your paper submission.} They are visible to the reviewers, area chairs, senior area chairs, and ethics reviewers. You will be asked to also include it (after eventual revisions) with the final version of your paper, and its final version will be published with the paper.

The reviewers of your paper will be asked to use the checklist as one of the factors in their evaluation. While "\answerYes{}" is generally preferable to "\answerNo{}", it is perfectly acceptable to answer "\answerNo{}" provided a proper justification is given (e.g., "error bars are not reported because it would be too computationally expensive" or "we were unable to find the license for the dataset we used"). In general, answering "\answerNo{}" or "\answerNA{}" is not grounds for rejection. While the questions are phrased in a binary way, we acknowledge that the true answer is often more nuanced, so please just use your best judgment and write a justification to elaborate. All supporting evidence can appear either in the main paper or the supplemental material, provided in appendix. If you answer \answerYes{} to a question, in the justification please point to the section(s) where related material for the question can be found.

IMPORTANT, please:
\begin{itemize}
    \item {\bf Delete this instruction block, but keep the section heading ``NeurIPS Paper Checklist"},
    \item  {\bf Keep the checklist subsection headings, questions/answers and guidelines below.}
    \item {\bf Do not modify the questions and only use the provided macros for your answers}.
\end{itemize}

%%% END INSTRUCTIONS %%%

\begin{enumerate}

\item {\bf Claims}
    \item[] Question: Do the main claims made in the abstract and introduction accurately reflect the paper's contributions and scope?
    \item[] Answer: \answerYes{} % Replace by \answerYes{}, \answerNo{}, or \answerNA{}.
    \item[] Justification: Our abstract and introduction appropriately include the claims made in the paper.

\item {\bf Limitations}
    \item[] Question: Does the paper discuss the limitations of the work performed by the authors?
    \item[] Answer: \answerYes{} % Replace by \answerYes{}, \answerNo{}, or \answerNA{}.
    \item[] Justification: See \autoref{sec:limitation}.

\item {\bf Theory assumptions and proofs}
    \item[] Question: For each theoretical result, does the paper provide the full set of assumptions and a complete (and correct) proof?
    \item[] Answer: \answerNA{} % Replace by \answerYes{}, \answerNo{}, or \answerNA{}.
    \item[] Justification: Our paper does not include theoretical results.

    \item {\bf Experimental result reproducibility}
    \item[] Question: Does the paper fully disclose all the information needed to reproduce the main experimental results of the paper to the extent that it affects the main claims and/or conclusions of the paper (regardless of whether the code and data are provided or not)?
    \item[] Answer: \answerYes{} % Replace by \answerYes{}, \answerNo{}, or \answerNA{}.
    \item[] Justification: We describe the details of experiments in Section~\ref{sec:main_results} and other necessary information in \autoref{sec:implementation_details}. 

\item {\bf Open access to data and code}
    \item[] Question: Does the paper provide open access to the data and code, with sufficient instructions to faithfully reproduce the main experimental results, as described in supplemental material?
    \item[] Answer: \answerYes{} % Replace by \answerYes{}, \answerNo{}, or \answerNA{}.
    \item[] Justification: Our experiments are based on the open-source dataset~\citep{liao2024evaluation, xu2016msr, meta2024moviegen}.
    Our experimental code is shown in \url{https://anonymous.4open.science/r/T2V-Diffusion-Search-537B}.

\item {\bf Experimental setting/details}
    \item[] Question: Does the paper specify all the training and test details (e.g., data splits, hyperparameters, how they were chosen, type of optimizer, etc.) necessary to understand the results?
    \item[] Answer: \answerYes{} % Replace by \answerYes{}, \answerNo{}, or \answerNA{}.
    \item[] Justification: We describe the details of experiments in Section~\ref{sec:main_results} and other necessary information in Appendix~\ref{sec:implementation_details}. 

\item {\bf Experiment statistical significance}
    \item[] Question: Does the paper report error bars suitably and correctly defined or other appropriate information about the statistical significance of the experiments?
    \item[] Answer: \answerYes{} % Replace by \answerYes{}, \answerNo{}, or \answerNA{}.
    \item[] Justification: We report the statistical significance of the Pearson correlation coefficient in \autoref{fig:hist_gemini} and Appendix~\ref{sec:reward_gpt_4}. We also report the reward values averaged over multiple prompts to reduce the variance.

\item {\bf Experiments compute resources}
    \item[] Question: For each experiment, does the paper provide sufficient information on the computer resources (type of compute workers, memory, time of execution) needed to reproduce the experiments?
    \item[] Answer: \answerYes{} % Replace by \answerYes{}, \answerNo{}, or \answerNA{}.
    \item[] Justification: See \autoref{sec:implementation_details}.
    
\item {\bf Code of ethics}
    \item[] Question: Does the research conducted in the paper conform, in every respect, with the NeurIPS Code of Ethics \url{https://neurips.cc/public/EthicsGuidelines}?
    \item[] Answer: \answerYes{}{} % Replace by \answerYes{}, \answerNo{}, or \answerNA{}.
    \item[] Justification: We believe our research conforms, in every respect, with the NeurIPS Code of Ethics.

\item {\bf Broader impacts}
    \item[] Question: Does the paper discuss both potential positive societal impacts and negative societal impacts of the work performed?
    \item[] Answer: \answerYes{} % Replace by \answerYes{}, \answerNo{}, or \answerNA{}.
    \item[] Justification: See \autoref{sec:broader_impacts}.
    
\item {\bf Safeguards}
    \item[] Question: Does the paper describe safeguards that have been put in place for responsible release of data or models that have a high risk for misuse (e.g., pretrained language models, image generators, or scraped datasets)?
    \item[] Answer: \answerNA{}{} % Replace by \answerYes{}, \answerNo{}, or \answerNA{}.
    \item[] Justification: This paper does not include new datasets or pre-trained models that pose a risk of misuse.

\item {\bf Licenses for existing assets}
    \item[] Question: Are the creators or original owners of assets (e.g., code, data, models), used in the paper, properly credited and are the license and terms of use explicitly mentioned and properly respected?
    \item[] Answer: \answerYes{}{} % Replace by \answerYes{}, \answerNo{}, or \answerNA{}.
    \item[] Justification: We have appropriately cited the papers of existing assets we used.

\item {\bf New assets}
    \item[] Question: Are new assets introduced in the paper well documented and is the documentation provided alongside the assets?
    \item[] Answer: \answerNA{} % Replace by \answerYes{}, \answerNo{}, or \answerNA{}.
    \item[] Justification: he paper does not release new assets.

\item {\bf Crowdsourcing and research with human subjects}
    \item[] Question: For crowdsourcing experiments and research with human subjects, does the paper include the full text of instructions given to participants and screenshots, if applicable, as well as details about compensation (if any)? 
    \item[] Answer: \answerNo{}{} % Replace by \answerYes{}, \answerNo{}, or \answerNA{}.
    \item[] Justification: We only recruited participants for user experiments to validate the effectiveness of our model, where they were asked to choose from generated videos. No human participants were involved in the dataset construction or model training process.

\item {\bf Institutional review board (IRB) approvals or equivalent for research with human subjects}
    \item[] Question: Does the paper describe potential risks incurred by study participants, whether such risks were disclosed to the subjects, and whether Institutional Review Board (IRB) approvals (or an equivalent approval/review based on the requirements of your country or institution) were obtained?
    \item[] Answer: \answerNA{} % Replace by \answerYes{}, \answerNo{}, or \answerNA{}.
    \item[] Justification: Our experiment solely involves measurement and does not entail behavioral manipulation; therefore, we did not apply for IRB approval.

\item {\bf Declaration of LLM usage}
    \item[] Question: Does the paper describe the usage of LLMs if it is an important, original, or non-standard component of the core methods in this research? Note that if the LLM is used only for writing, editing, or formatting purposes and does not impact the core methodology, scientific rigorousness, or originality of the research, declaration is not required.
    %this research? 
    \item[] Answer: \answerNA{} % Replace by \answerYes{}, \answerNo{}, or \answerNA{}.
    \item[] Justification: We use LLMs only for writing.

\end{enumerate}

\end{document}